\documentclass[12pt]{report}
\usepackage{suthesis}

\usepackage{times,latexsym,amsfonts,amssymb,amsmath,graphicx,url,bbm,rotating}
\usepackage{enumitem,multirow,hhline,stmaryrd,bussproofs,mathtools,siunitx,arydshln}
\usepackage{booktabs,xcolor,csquotes,calligra}

\usepackage{fitch}
\usepackage{amsmath}

\usepackage{natbib}
\usepackage{tikz}
\usepackage{tikz-dependency,pifont}
\usetikzlibrary{shapes.arrows,chains,positioning,automata,trees,calc}
\usetikzlibrary{patterns,matrix}
\usetikzlibrary{decorations.pathmorphing,decorations.markings}
\usepackage[ruled,lined,linesnumbered]{algorithm2e}

\newcommand\st{\ensuremath{\mathcal{t}}}

\newcommand\by{\ensuremath{\mathbf{y}}}

\newcommand\argmax{\mathop{\text{argmax}}}

\usepackage{color}

\newcommand{\hl}[2]{\colorbox{#2}{#1}}

\def\todo#1{\hl{{\bf TODO:} #1}{yellow}}

\definecolor{darkred}{rgb}{0.5451, 0.0, 0.0}
\definecolor{darkgreen}{rgb}{0.0, 0.3922, 0.0}

\definecolor{burntorange}{HTML}{BF5700}

\tikzset{
    invisible/.style={opacity=0},
    visible on/.style={alt=#1{}{invisible}},
    alt/.code args={<#1>#2#3}{%
      \alt<#1>{\pgfkeysalso{#2}}{\pgfkeysalso{#3}} 
    },
  }

\def\newcite#1{\citet{#1}}
\def\cite#1{\citep{#1}}
\definecolor{darkblue}{rgb}{0.0,0.0,0.4}

\newenvironment{tightitemize}%
  {\begin{itemize}[topsep=0pt, partopsep=0pt] %
    \setlength{\parskip}{0pt}%
    }%
  {\end{itemize}}

 \usepackage{epigraph}

\newcommand{\bleu}{{{\sc Bleu}}\xspace}

\newcommand{\eos}{{\it EOS}\xspace}
\newcommand{\sts}{{{\textsc{Seq2Seq}}}\xspace}

\newcommand{\MINUS}{}
\DeclareMathOperator{\softmax}{softmax}
\newcommand{\mmiBD}{{MMI-bidi}\xspace}
\newcommand{\mmiLM}{{MMI-antiLM}\xspace}

\usepackage{tabularx}
\usepackage{tabulary}
\usepackage{comment}
\usepackage{booktabs}

\newcommand{\mmiBDv}{{$(1-\lambda) \log p(y|s)+\lambda \log p(x|y)$}\xspace}
\newcommand{\mmiLMv}{{$\log p(y|x)-\lambda U(y)$}\xspace}
\newcommand{\Message}{{{\it message}}\xspace}
\newcommand{\Response}{{{\it response}}\xspace}

\newcommand{\User}[1]{{{\it user{#1}}}\xspace}

\onlinetrue

\begin{document}

\title{Teaching Machines to Converse}
\author{Jiwei Li}
\principaladviser{Dan Jurafsky}
\firstreader{Christopher G. Potts}
\secondreader{Emma Brunskill}

\beforepreface
\prefacesection{Abstract}
The ability of a machine to communicate with humans has long been associated with the general success of AI. This dates back to 
Alan Turing's epoch-making work in 
the early 1950s, 
which proposes that  
a machine's intelligence 
can be tested by 
how well it, the machine, can fool a human
into believing that the machine is a human through dialogue conversations. 
Despite progress in the field of dialogue learning over the past decades, 
conventional dialog
systems 
still 
face a variety of major challenges such as robustness, scalability and domain adaptation: 
many systems 
 learn generation
rules from a minimal set of authored rules or
labels on top of  handcoded
rules or templates, and thus are both expensive
and difficult to extend to open-domain scenarios.
Meanwhile, dialogue systems have become increasingly 
complicated: they usually involve building many different complex components separately, rendering them unable to accommodate 
the large amount of data that we have to date. 

Recently, the emergence of neural network models the potential to solve many of the  problems 
in dialogue learning
that earlier systems cannot tackle:
the end-to-end neural frameworks
offer the promise of scalability
and language-independence,  together with the
ability to track the dialogue state and 
then mapping between states and dialogue actions
 in a way not possible with conventional 
systems.   
On the other hand,  neural systems  bring about new challenges: 
they tend to output dull and generic responses such as ``I don't know what you are talking about"; they lack a consistent or a coherent persona; they are usually optimized through single-turn conversations and are incapable of handling the long-term success of a conversation; and they are not able to take the advantage of the interactions with humans.

This dissertation 
attempts to tackle these challenges:
Contributions are twofold: 
(1) we address new challenges presented by  neural network models in open-domain dialogue generation systems, which includes (a) using mutual information to avoid dull and generic responses; (b) addressing user consistency issues to avoid inconsistent responses generated by the same user; (c) developing reinforcement learning 
methods to foster the long-term success of conversations; and (d) using adversarial learning methods to push machines to generate responses that are indistinguishable from human-generated responses; 
(2) we develop interactive 
question-answering
dialogue  systems by
(a) giving the agent the ability to ask questions and (b) training a conversation agent through interactions with humans in an
online fashion, where a bot improves through communicating with humans and  learning from the mistakes
that it makes.

\prefacesection{Acknowledgments}

Getting to know Prof. Dan Jurafsky is the most precious part of my experience at Stanford. He perfectly exemplifies what a supportive and compassionate person should be like. 
Over the years, I have always been taking what he did for me
for granted: taking for granted that he spent 5 hours on 
editing
every paper and correcting all the typos and grammar mistakes from a non-native English speaker; taking for granted his generous support 
each and 
every time when I experienced down times in graduate school; taking for granted 
that he is always flexible with my schedules such as going back to 
the
mainland China visiting families in the middle of the school year. The opportunity of working with him and getting to know him is the most valuable part of my years at Stanford, and these experiences 
transcend all the tiny academic attainments in graduate school.

I want to thank 
Prof. Eduard Hovy. It is hard to evaluate how much his support means to me during the hardest time of my life. He was not like an advisor, but like an elder friend, 
telling an ignorant twenty-something how he should face ups and downs of life. He instilled in me the great importance of doing big work and solving big problems. Unfortunately, my pragmatic nature made me unwilling to invest time into ambitious, but risky research projects. I  kept working on tiny incremental projects to satisfy my utilitarian desire for publishing. Not living up to his expectation is the biggest regret of my graduate school experience.

I want to particularly thank Prof. Claire Cardie. When I was a first-year graduate student studying biology at Cornell, barely having any experience with coding but trying to switch to the field of computer science,
she gave me the opportunity of being exposed to the intriguing nature of language processing using AI,  
and this opportunity
 completely changed the trajectory of my life. 
She invested her precious time into teaching me how to do research, and was patient as I slowly developed basic  coding skills. Her support continues over years after I left Cornell. 

I want to  thank Prof. Alan Ritter, for introducing me to 
 the basic skills in doing research.
I want to thank Prof. Sujian Li, for introducing me to the
beautiful 
 world of computer science, and for 
the support over the past many years.
I owe a big thank-you to Prof. Peter Doerschuk from the Biomedical Engineering department at Cornell, for all the encouragement and  support when I was trying to change my research field, and for telling me nothing is more important that doing things that I am happy about.  
 
During my years in graduate school, I have been blessed with the friendship of many amazing people. 
I want to thank all friends from Stanford IVgrad,  Daniel Heywood, Andrew Melton, 
Shannon Kao,
Ben Bell, Vivian Chen, Justin Su, Emily Kolenbrander, Tim Althoff, 
Caroline Abadeer,
Connor Kite
and Wendy Quay,
who provide me with a sense of family and identity in graduate school. 
IVgrad teaches me life is not all about research and research is not all about myself. 
Over the years, 
I have had the opportunity  of befriending, working with or learning from many wonderful students, professors
and researchers  at Stanford, CMU, Cornell, Microsoft Research, and Facebook, including 
Gabor Angeli, 
Chris Brockett, 
Danqi Chen, Xinlei Chen, Sumit Chopra, 
Eyal Cidon, 
 Bill Dolan, 
 Chris Dyer, 
 Jianfeng Gao, Michel Galley, 
Kelvin Guu,
He He, De-an Huang, Ting-Hao Kenneth Huang,
Sujay Kumar Jauhar,
Sebastien Jean, 
Percy Liang, 
Minh-Thang Luong, Hector Zhengzhong Liu, Max Xuezhe Ma, Chris Manning, 
Andrew Maas, 
Bill MacCartney,
Alexander Miller, 
Myle Ott, 
Chris Potts, 
Marc'aurelio Ranzato,
Mrinmaya Sachan,
Tianlin Shi, 
Noah Smith,
Russell Stewart, 
Yi-Chiang Wang, 
Di Wang, 
Willian Wang, 
Lu Wang, 
Sam Wiseman, 
Jason Weston, 
Ziang Xie,
Bishan Yang,
Diyi Yang, 
Shou-Yi Yu,
Arianna Yuan,
Jake Zhao and many many others who I am grateful for but cannot list all their names. 
I want to specially thank Will Monroe, Sida Wang, 
and Ziang Xie for the friendship, and
 for always being helpful and encouraging when
 graduate school 
  life is hard. 

I want to thank Prof. Emma Brunskill,
Prof. Christopher Potts  and Jason Weston, for agreeing to be on my thesis committee and  helping me all the way through the proposal, the defense and the 
final version of the thesis. I also want to thank Prof. Olivier Gevaert for helping chair my thesis defense. 

A special thank is owed to the administrative staff: Jay Subramanian, Helen Buendicho, Stacey Young, 
and Becky Stewart, who have help reduced many of the stresses in graduate school. 

Finally, I want to thank my families. My debt to them is unbounded.

\afterpreface
\chapter{Introduction}
\label{intro}
\epigraph{By your words you will be justified, and by your words you will be condemned.}{Matthew 12:37.}
Conversation in the form of  languages has always been a trademark of humanity. 
It is one of the first skills that we humans acquire as kids, and never cease to use throughout our lives, whether 
 a person is ordering a burrito for lunch, taking classes, 
doing job interviews, discussing the result of an NHL game with friends, or even arguing with others. 
The significance of conversation/communication also transcends individuals: through conversations, humans can communicate a huge amount of information not only about
our surroundings (e.g., asking companions to watch out for the lions in the forest), but about ourselves (e.g., giving orders, talking about personal needs, etc). Such an ability leads to more effective social cooperation, and is a necessity for  
organizing a larger group of people (e.g., company, troop, etc). 

In the field of  artificial intelligence, attempts to imitate humans' ability to converse can be dated back to the early days of AI, and
this ability has long been 
associated with the general success of AI. 
In his epoch-making paper \cite{turing1950computing}, Alan Turing
proposed a method to test the general intelligence level of a machine, which is widely known as the Turing test or the Imitation game.
In the Turing test, 
 a machine is asked to 
talk with a human. The machine's intelligence level is decided by how well the machine is able to fool a human evaluator into believing that 
it, the machine, is a human based on its text responses. 
If the human evaluator cannot  tell the difference between the machine from a human, the machine is said to have passed the Turing test, which signifies a 
high level of intelligence of an AI.

Ever since the idea of the Turing test was proposed, various attempts have been proposed to pass the test. 
But we are still far from passing the test. 
In this section, we will briefly review the systems that have been proposed over the past decades.
Specifically, 
we will discuss 
three types of 
dialogue systems:  the chit-chat system, 
the frame-based goal oriented system, and
the interactive question-answering (QA) dialogue system. 
We will discuss the cases where they have been successfully applied, their pros and cons, and  why they are still not able to  pass the Turing test. 
The major focus of this thesis is about how to improve 
the chit-chat system and the interactive question-answering (QA) system. 

\section{A Brief Review of Existing Dialogue Systems}
\subsection{The Chit-chat Style System}
A chit-chat-oriented dialogue agent is designed to engage users, comfort them, provide mental support, or just chat with users whichever topic they want to talk about. 
The bot first needs to understand what its dialogue partner says, and then 
 generate meaningful and coherent responses based on this history.  
 Existing chit-chat systems mostly fall into the following three subcategories: the rule-based systems, the IR-based systems and the generation-based systems, as will be discussed in order below. 

\subsubsection{The Rule-based Systems}
Using rules is one of the most effective ways to generate dialogue utterances. 
Usually,
message inputs are first assessed  based on a set of pre-defined rules, e.g., 
a key-word look-up dictionary,
if-else conditions, or more sophisticated machine learning classifiers.
After rule conditions are evaluated, relevant actions will be executed, such as 
outputting an utterance in storage, manipulating the input message or selecting some related historical contexts.

One of the most famous 
rule-based 
dialogue systems in history is  ELIZA \cite{weizenbaum1966eliza}. 
ELIZA operates by first searching a keyword existing in the input text from a real human
 based on a hand-crafted keyword dictionary. If a keyword is found, a rule is applied to 
 manipulate and
 transform the 
user's original input and forwarded back to the user. Otherwise, ELIZA responded either with a generic response or copying one sentence from the dialogue history.\footnote{\url{https://en.wikipedia.org/wiki/ELIZA}} 
 Extensions of ELIZA include PARRY \cite{parkinson1977conversational}, also described as "ELIZA with attitude",
which simulated a patient with schizophrenia. 
PARRY relies on global variables to keep track of the emotional state, as opposed to ELIZA where responses are generated only based on the previous sentence. 
A variety of chit-chat systems such as  Eugene Goostman,\footnote{\url{https://en.wikipedia.org/wiki/Eugene_Goostman}} 
Jabberwacky,\footnote{\url{https://en.wikipedia.org/wiki/Jabberwacky}} 
Cleverbot,\footnote{\url{https://en.wikipedia.org/wiki/Cleverbot}} 
Alice,\footnote{\url{https://en.wikipedia.org/wiki/Artificial_Linguistic_Internet_Computer_Entity}} 
AIML\footnote{\url{https://en.wikipedia.org/wiki/AIML}} were proposed after Eliza and PARRY. 

ELIZA-style systems are recognized as an important milestone in developing modern dialogue systems. 
More interestingly, 
 some of the systems seemed to be able to deceive some human evaluators to believe that they were talking with real people in a few specific scenarios \cite{thomas1995social,colby1972turing,pinar2000turing}.
 On the other hand, their drawbacks are obvious: 
Rule-based systems 
predominantly rely on
 the set of pre-defined rules.  
The number of these rules  skyrockets as the system gets more sophisticated;
Rule-based systems do not have the ability to understand human languages, nor do they know how to generate meaningful natural language utterances.
They are 
 thus only able to conduct very superficial conversations.

\subsubsection{The IR-based Systems}
The IR-based methods  rely on information retrieval or nearest neighbor techniques  \cite{isbell2000cobot,jafarpour2010filter,yan2016learning,al2016conversational,yan2016docchat}.
Given a history input and a training corpus, the system copies a response from the training corpus.
The response selection process is usually
 based on the 
combination of the following
 two criteria: 
the history associated with the chosen response should be similar to the input dialogue history and the 
chosen response should be semantically related to the input dialogue history. 
Various ranking schemes such as semantic relatedness measurements (e.g., vector space models or TF-IDF), 
page-rank style relatedness propagation models, or personalization 
techniques
can be ensembled into a single ranking function 
and the response with the highest ranking score will be selected. 
The pros and cons of the IR-based models are both obvious: on one hand, the models are easy to implement relative to generation-based models;
responses are always grammatical (since responses are copied from the training set); and through the manipulation of 
the ranking function (e.g., adding rules, upweighing or downweighting some particular features), a developer has a relatively good (and direct) control on generating responses that he or she would like to see.\footnote{This is hardly true, or at least requires more human efforts in the generation-based system. One example is the generation-based system {\it Tay}, an artificial intelligence chatterbot via Twitter released by Microsoft, which posts inflammatory, offensive or even sexist and racist responses.} 
But on the other hand, the IR-based models lack the flexibility in handling the diversity of natural languages, the ability to handle important linguistic features such as 
context structure or coherence, 
and the capability of discerning 
 the subtle semantic difference between different input contexts. 
\subsubsection{The Generation-based Systems}
\begin{figure}
\center
\includegraphics[width=4in]{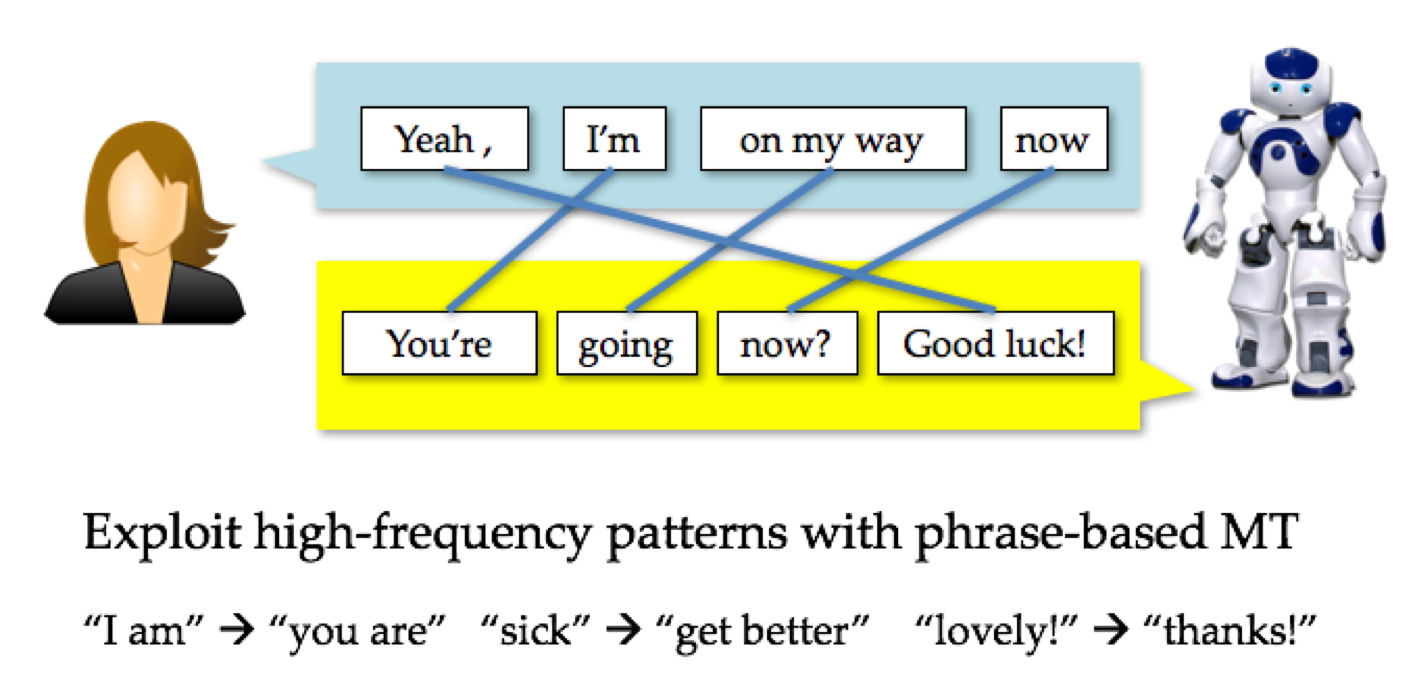}
\caption[Word alignments between   messages and responses using the IBM model]{Word alignments between   messages and responses using the IBM model. Image courtesy of Michel Galley.}
\label{IBM}
\end{figure}

The generation-based system generates sentences token by token instead of  copying responses from the training set. The task can be formalized 
as an input-output mapping problem, where given the history dialogue utterances, the system needs to output a coherent and meaningful sequence of words.\footnote{Here, we refer to dialogue history, sources, inputs, stimulus,
and messages interchangeably, which all mean the dialogue history. We also refer to targets, outputs, and responses interchangeably, which mean the natural language utterance that the system needs to generate.}
The task was first studied by \newcite{ritter2011data},  who frame the response generation task as a
statistical machine translation (SMT) problem. The IBM-model \cite{brown1991aligning} is used  to learn the  word mapping rules between source  and target words (as shown in Figure \ref{IBM}) and the phrase-based MT model \cite{chiang2007hierarchical} is used for word decoding. The disadvantage of the MT-based system stems not only from the  complexity of the phrase-based MT model with many different components built separately, but also from the IBM model's intrinsic inflexibility in handling 
the 
implicit semantic and syntactic
relations between message-response pairs: unlike in MT, where there is usually a direct mapping between a word or phrase in the source sentence and another 
word or phrase
in the target sentence, 
in response generation, the  mapping is mostly beyond the word level, and 
 requires the semantic of the entire sentence. 
Due to this reason, the MT-based system is only good at handling the few cases in which word-level mapping is very clear as in Figure \ref{IBM}, but usually
fail to tackle
  the situations once the semantic of an input sentence gets  complex, sometimes outputting incoherent, or even ungrammatical responses. 
Furthermore, the MT-based system lacks the ability to leverage information in the multi-context situation. 

Recent progress in SMT, which stems  from the use of neural 
language models \cite{mikolov2010recurrent,kalchbrenner2013recurrent,vaswani2013decoding} and neural sequence-to-sequence generation models (\sts) \cite{sutskever2014sequence,bahdanau2014neural,cho2014learning,luong2015effective,luong2014addressing,luong2016achieving}
have inspired a
variety of
attempts to extend neural
techniques to response generation \cite{sordoni2015neural,ghazvininejad2017knowledge,mostafazadeh2017image,serban2017hierarchical,serban2016building,serban2016generative,serban2017multiresolution}. 
Neural models offer the promise of scalability
and language-independence, together with the
capacity to implicitly learn semantic and syntactic
relations between pairs, and to capture contextual dependencies
in a way not possible
with conventional SMT approaches or IR-based approaches.

Due to these advantages, neural generation models are able generate more specific, coherent, and meaningful dialogue responses. 
On the other hand, 
a variety of important issues still remain unsolved: current  systems tend to generate plain and dull responses such as ``i don't know what you are talking about", which discourages the conversation; it is hard to endow a dialogue system with a consistent 
element of identity or persona (background facts or user
profile), language behavior or interaction style;
current systems usually focus on single-turn conversations, or two-turns at most, since
it is hard to give the system a long-term planning ability 
to conduct 
 multi-turn conversations that flow smoothly, coherently, and meaningfully.  
This thesis tries to address these issues.

\subsection{The Frame-based  Dialogue Systems}
The frame-based system, first proposed by \newcite{bobrow1977gus}, models conversations  guided by frames, which represent the information at different 
levels within a conversation. 
For example, in a conversation about plane ticket booking, frames include 
important aspects such as 
{\it Person}, {\it Traveling Date}, 
{\it Destination}, {\it TimeRange of the flight}, etc. 
Frames of a conversation define the aspects that the conversation should cover and the relations (e.g., one frame is a prototype of another)
between frames
define how conversations should flow  \cite{walker1990mixed,seneff1992tina,chu1997tracking,san2001designing,seneff2002response}. 

  The simplest frame-based system is a finite-state machine, which asks the user a series of pre-defined questions based on the frames,
  and moves on to the next question
    if a customer provides an answer, and 
     ignores anything from the 
  customer if the response is not an answer. 
More complicated architectures
 allow the initiative of the conversation between the system and the user to shift at various points.
 These systems rely on a pre-defined frame and asks the user to fill slots in the frame, where a task is completed if all frame slots 
 have been filled . 
The limitation of the frame-based system is that the dialogue generation process is completely guided by what needs  to fill the slots. The system doesn't
have the ability to decide the progress
or the state
 of the conversation, e.g., whether the customer has rejected a suggestion, asked a question, or whether the system now needs to 
give suggestions or ask clarification questions, etc, and is thus not able to take a correct action given the progress so far.

To overcome these drawbacks, 
 more sophisticated state-based dialogue models \cite{nagata1994first,reithinger1996predicting,warnke1997integrated,stolcke2000dialogue,allen2001architecture} were designed. 
The \textsc{state-based dialogue system} is based on two key concepts, \textsc{dialogue state}, 
which denotes the progress of current conversation, including context information, intentions of the speakers, etc, 
and \textsc{dialogue acts}, which characterize the category of a dialogue utterance. 
The choice of a \textsc{dialogue act} is based on the \textsc{dialogue state} the conversation is currently in, and the key component of this system is to 
learn the optimal mapping between a state and an action to take, which is able  to maximize dialogue success.  
Reinforcement learning methods 
such as MDPs or POMDPs are widely used 
to learn such mappings based on external rewards that define dialogue success 
\cite{young2000probabilistic,young2002statistical,lemon2006isu,williams2007partially,young2010hidden,young2013pomdp}.
Recent advances in neural network models provide with more power and flexibility in keeping track of dialogue states and 
 modeling the mapping between the dialogue states and utterances to generate  \cite{wen2015semantically,mrkvsic2015multi,su2015learning,wen2016network,wen2016conditional,su2016continuously,su2016line,wen2017latent}. 

Frame-based systems have succeeded in a variety of applications such as booking flight tickets, reserving restaurants, etc, some of which have already been 
in use in our everyday life.  
The biggest advantages of frame-based system are that 
the goal of the system is explicitly defined and that the pre-defined frames give a very clear guidance on how a conversation should proceed.
On the other hand, its limitation is  clear:
 frame-based systems heavily rely on sophisticated  hand-crafted patterns or rules, and these rules are costly;
 rules have to be rebuilt when the system is adapted to a new domain or an old domain changes, making the system difficult to scale up. 
More broadly, 
it does not touch  the complex linguistic features involved in human conversations, such as context coherence, 
word usage (both semantic and syntactic), 
personalization, 
and  
are thus not able to capture the complexity and the intriguing nature of humans'
conversation. 

In this thesis, we do not focus on frame-based systems.

\subsection{The Question-Answering (QA) Based Dialogue System}
Another important dialogue system is the 
 factoid QA-based dialogue system, 
which is closely related to developing automated personal assistant systems such as Apple's Siri. 
A dialogue agent needs to answer customers' questions regarding different topics 
such as weather conditions, traffic congestion, news, stock prices, user schedules, retail prices, etc \cite{d1993personal,modi2005cmradar,myers2007intelligent,berry2011ptime}, either given a database of knowledge \cite{dodge2015evaluating,bordes2015large,weston2016dialog},
or from  texts \cite{hermann2015teaching,weston2015towards,weston2016dialog}.
The QA based dialogue system is thus related to a wide range of work in text-based or knowledge-base based question answering, e.g., \cite{hirschman2001natural,clarke2003passage,maybury2008new,berant2013semantic,iyyer2014neural,rajpurkar2016squad}  and many many others. 
The key difference between a QA-based dialogue system and a factoid QA-system is that 
the QA-based dialogue system is interactive \cite{rieser2009does}: instead of just having to answer one single question, 
the QA-based system needs the ability to handle a diverse category of interaction-related issues such as 
 asking for question clarification \cite{stoyanchev2013modelling,stoyanchev2014towards}, adapting answers given 
a human's feedback \cite{rieser2009does}, 
self-learning when encountering new questions or concepts
\cite{purver2006clarie}, etc.
To handle these issues, 
 the system needs to 
 take proper  actions based on the current conversation state, which resembles the key issue addressed in the \textsc{state-based dialogue system}. 

How a bot can 
be smart about interacting with humans, 
and how to improve itself through these interactions are not sufficiently studied. 
For example, asking for question clarification is only superficially touched in 
\newcite{stoyanchev2013modelling,stoyanchev2014towards} for cases where some important tokens are not well transcribed from the speech, for example,

A: {\it When did the problems with [power] start}?

B: {\it The problem with what}?

A: {\it Power}.

\noindent But  important scenarios such as what if  there is 
an out-of-vocabulary word in the original question,  how a bot can ask for hints, 
 and more importantly,  how a bot can 
be smart about deciding whether, when and what to ask  have rarely been studied. 
Another important aspect 
in developing interactive agent
that is missing from existing literature is that
a good  agent should have the
ability to learn from the online feedback: adapting its model when making mistakes
and reinforcing the model when a human's feedback is positive. This is particularly important
in the situation where the bot is initially trained in a supervised way on a fixed synthetic, domain-specific
or pre-built dataset before release, but will be exposed to a different environment after
release (e.g., more diverse natural language utterance usage when talking with real humans, different
distributions, special cases, etc.). 
There hasn't been any work discussing how a bot effectively improve itself from online feedback by accommodating various feedback signals.  
This thesis tries to address these questions.

\section{Thesis Outline}
In this dissertation, we mainly address problems involved in the  
chit-chat system and the interactive QA system. 
First, we explore how to build an engaging chit-chat style dialogue system that is able to conduct 
interesting, meaningful, coherent, consistent, and long-term conversation with humans.  
More specially, 
for the chit-chat system,
we (a) use mutual information to avoid dull and generic responses
\cite{li2015diversity,li2016simple,li2017learning}; (b) address user consistency issues to avoid inconsistent responses from the same user \cite{li2016persona}; (c) develop reinforcement learning 
methods to foster the long-term success of conversations \cite{li2016deep}; and (d) use adversarial learning methods to generate machine responses that are indistinguishable from human-generated responses \cite{li2017adversarial}; 

Second, we explore how a bot can best improve itself through the online interactions with humans that makes a
chatbot system trully \textsc{interactive}. 
We  
develop interactive dialogue  systems for factoid question-answering: 
(a)  we
design an environment that provides the agent the ability to ask humans questions  and to learn when and what to ask \cite{li2016learning};
(b) we train a conversation agent through interaction with humans in an
online fashion, where a bot improves through communicating with humans and learning from the mistakes
that it makes \cite{li2016dialogue}. 

We start off by providing background knowledge on \sts models, memory network models and policy gradient reinforcement learning models in Chapter 2. 
The aforementioned four problems for the
chit-chat style
 dialogue generation systems will be detailed in Chapters 3,4,5,6,
 and the two issues with the interactive QA system will be detailed in Chapters 7 and 8. 
 We conclude this dissertation and discuss future avenue for chatbot development in Chapter 9.

\subsection{Open-Domain Dialogue Generation}
\subsubsection*{Mutual Information to Avoid Generic Responses} 
An engaging response generation system should be able to output grammatical, coherent responses that are diverse and interesting.
In practice, however,  neural conversation models exhibit a tendency to generate dull, trivial or non-committal responses, often involving high-frequency phrases along the lines of \textit{I~don't know} or  \textit{I'm OK} \cite{sordoni2015neural,serban2015hierarchical,vinyals2015neural}. 
This behavior is ascribed to the relative frequency of generic responses like 
 \textit{I don't know} in conversational datasets, in contrast with the relative sparsity of other, more contentful or specific alternative responses.
 It appears that by optimizing for the likelihood of outputs/targets/responses given inputs/sources/messages, 
neural models assign high probability to ``safe'' responses. 
The question is how to overcome the neural models' predilection for the commonplace. Intuitively, we want to capture not only the dependency of responses on messages, but also the inverse, the likelihood that a message will be provided to a given response. Whereas the sequence  \textit{I don't know} is of high probability in response to most question-related messages, the reverse will generally not be true, since \textit{I don't know}  can be a response to everything, making it hard to guess the original input question. 

We propose to capture this intuition by using Maximum Mutual Information (MMI), as an optimization objective that measures the mutual dependence between inputs and outputs, as opposed to 
the uni-directional dependency from sources to targets in the traditional MLE objective function.
We present practical training and decoding strategies for neural generation models that use MMI as objective function.
We demonstrate that using MMI results in a clear decrease in the proportion of generic response sequences, and 
find a significant performance boost from the proposed models as measured by  BLEU \cite{papineni2002bleu} and human evaluation.
{\it This chapter is based on the following three papers: \cite{li2015diversity,li2016simple,li2017learning}, and will be detailed in Section 3}. 

\subsubsection*{Addressing the Speaker Consistency Issue} 
An issue that stands out with current dialogue systems is the lack of speaker consistency: if a human asks a bot a few questions, there is no guarantee that answers from the bot are consistent. 
This is because 
responses are selected based on likelihood assigned by the pre-trained model, which does not have the ability to model speaker consistency. 

In \newcite{li2016persona}, we address the challenge of consistency and how to endow data-driven systems with the coherent ``persona'' needed to model human-like behavior, whether as personal assistants, personalized avatar-like agents, or game characters.\footnote{\cite{vinyals2015neural} suggest that the lack of a coherent personality makes it impossible for current systems to pass the Turing test.} 
For present purposes, we will define \textsc{persona} as the character that an artificial agent, as actor, plays or performs during conversational interactions.
A~persona can be viewed as a composite of elements of identity (background facts or user profile), language behavior, and interaction style. 
A persona is also adaptive, since an agent may need to present different facets to different human interlocutors depending on the demands of the interaction. 
We 
 incorporate personas as embeddings and explore two persona models, a single-speaker \textsc{Speaker Model} and a dyadic \textsc{Speaker-Addressee Model}, within the \sts framework. 
The Speaker Model integrates a speaker-level vector representation into the target part of the \sts model.
Analogously, the Speaker-Addressee model encodes the interaction patterns of two interlocutors by constructing an interaction representation from their individual embeddings and incorporating it into the \sts model. 
These persona vectors are trained on human-human conversation data and used at test time to generate personalized responses.
Our experiments on an open-domain corpus of Twitter conversations and dialog datasets comprising TV series scripts show that leveraging persona vectors can improve relative performance up to $20\%$ in BLEU score and $12\%$ in perplexity, with a commensurate gain in consistency as judged by human annotators. 
{\it This chapter is based on the following paper: \newcite{li2016persona}, and will be detailed in Section 4}.

\subsubsection*{Fostering Long-term Dialogue Success}
Current dialogue generation models are trained by predicting the next {\bf single} dialogue turn in a given conversational context using the maximum-likelihood estimation (MLE) objective function. 
However, this does not mimic how we humans talk. In everyday conversations from a human, each human dialogue episode consists tens of, or even hundreds of dialogue turns rather than only one turn; humans are
smart in controlling the informational flow in a conversation for the long-term success of the conversation. 
Current models' incapability of handling this long-term success result in repetitive and generic responses.\footnote{The fact that current models tend to generate highly generic responses such as``{\it I don't know}" regardless of the input \cite{sordoni2015neural,serban2015hierarchical,li2015diversity} can be ascribed to the model's incapability of handling long-term dialogue success:
apparently {\it ``I don't know"} is not a good action to take, since it closes the conversation down} 

We need a conversation framework that has the ability to (1)  integrate developer-defined rewards that better mimic the true goal of chatbot development
  and (2)  model the long-term influence of a generated response in an ongoing dialogue.
To achieve these goals,  we draw on the insights of reinforcement learning, which
have been widely applied in MDP and POMDP dialogue systems.
We introduce a neural reinforcement learning (RL) generation method, 
which can optimize long-term rewards designed by system developers.
Our model uses the encoder-decoder architecture as its backbone, and 
 simulates conversation between two virtual agents to explore the space of possible actions while learning to maximize expected reward.
We define simple heuristic approximations to rewards that characterize
 good conversations: good conversations are forward-looking \cite{All92} or interactive (a turn suggests
 a following turn), informative, and coherent.
  The parameters of an encoder-decoder RNN define a policy over an infinite action space consisting of all possible utterances.
  The agent learns a policy by optimizing the long-term developer-defined reward from ongoing dialogue simulations using policy gradient methods \cite{williams1992simple},
  rather than the MLE objective defined in standard \sts models. 

Our model thus integrates the power of \sts systems to learn compositional semantic
meanings of utterances with the strengths of reinforcement learning in
optimizing for long-term goals across a conversation.
 Experimental results  demonstrate that our approach fosters a more sustained dialogue and
  manages to produce more interactive  responses than standard \sts models trained using the MLE objective.
  {\it This chapter is based on the following paper: \newcite{li2016deep}, and will be detailed in Section 5}.

  \subsubsection*{Adversarial Learning for Dialogue Generation}
  Open domain dialogue generation  aims at generating meaningful and
coherent dialogue responses given input dialogue
history. 
   Current  systems   
   approximate such a goal 
   using imitation learning or variations of imitation learning: 
    predicting the
next dialogue utterance in human conversations
given the dialogue history. 
 Despite its success, many
issues emerge resulting from this over-simplified
training objective: responses are highly dull and
generic,  repetitive, and short-sighted.

Solutions to these problems require answering a few fundamental questions: 
what are the crucial aspects that define an ideal conversation, how can we quantitatively measure them, and how can we incorporate them into 
a machine learning system:
A good dialogue model should generate utterances indistinguishable from human dialogues.
Such a goal suggests a training objective 
resembling the idea of the Turing test \cite{turing1950computing}.
We borrow the idea of adversarial training \cite{goodfellow2014generative} in 
computer
vision, in which we jointly train two models, 
a generator (which takes the form of the neural  \sts model) that defines the probability of generating a dialogue sequence, and 
a discriminator
that labels dialogues as human-generated or machine-generated. 
This discriminator  is analogous to  the evaluator in the Turing test.
We cast the task as a reinforcement learning problem, in which the quality of machine-generated utterances is measured by its ability to fool the discriminator into believing that it is a human-generated one. The output from the discriminator is used as a reward to the generator, pushing it to generate 
utterances indistinguishable from human-generated dialogues. 

Experimental results demonstrate that our approach
produces more interactive, interesting, and
non-repetitive responses than standard SEQ2SEQ
models trained using the MLE objective function.
  {\it This chapter is based on the following paper: \newcite{li2017adversarial}, and will be detailed in Section 6}. 
\subsection{Building Interactive Bots for Factoid Question-Answering}

  \subsubsection*{Learning by  Asking Questions}
  For current chatbot systems, when the bot encounters a confusing situation such as an unknown surface form (phrase
or structure), a semantically complicated sentence or an unknown word, the agent will either make
a (usually poor) guess or will redirect the user to other resources (e.g., a search engine, as in Siri).
Humans, in contrast, can adapt to many situations by asking questions: when a student is asked a question by a teacher, but is not confident about the answer, they may ask
for clarification or hints. A good conversational agent should have this ability
to interact with a customer. 

Here, we try to bridge the gap between how a human and an end-to-end machine learning system by equipping the bot with the ability to ask questions.
We identify three categories of mistakes a bot can make during dialogue
: (1) the bot has
problems understanding the surface form of the text of the dialogue partner, e.g., the phrasing of
a question; (2) the bot has a problem with reasoning, e.g., it fails to retrieve and connect the
relevant knowledge to the question at hand; (3) the bot lacks the knowledge necessary to answer
the question in the first place -- that is, the knowledge sources the bot has access to do not contain
the needed information.
All the situations above can be potentially addressed through interaction with the dialogue partner.
Such interactions can be used to learn to perform better in future dialogues. If a human bot has
problems understanding a teacher's question, they might ask the teacher to clarify the question. If
the bot doesn't know where to start, they might ask the teacher to point out which known facts
are most relevant. If the bot doesn't know the information needed at all, they might ask the
teacher to tell them the knowledge they're missing, writing it down for future use.

We explore how
a bot can benefit from interaction by asking questions in both offline supervised settings and online
reinforcement learning settings, as well as how to choose when to ask questions in the latter setting.
In both cases, we find that the learning system improves through interacting with users.
{\it This chapter is based on the following paper: \newcite{li2016learning}, and will be detailed in Section 7}. 

   \subsubsection*{Dialogue Learning with Human-in-the-Loop}
  A good conversational agent  
should have the ability to learn from the online feedback from a teacher: adapting its model when making mistakes and reinforcing the model when the teacher's feedback is positive.
This is particularly important in the situation where the bot is initially trained in a supervised way on a fixed synthetic, domain-specific or pre-built dataset before release, but will be exposed to
a different environment after release
 (e.g.,
 more diverse natural language utterance usage when talking with real humans, different distributions, special cases, etc.).  Most recent research
has focused on training a bot
from fixed training sets of labeled data but seldom on how the bot can
improve through online interaction with humans.
Human (rather than machine) language learning happens during communication \citep{bassiri2011interactional,werts1995instructive},
and not from labeled datasets, hence making this an important subject to study.

Here, we explore this direction by
 training a bot through interaction with teachers in an online fashion.
 The task is formalized under the general framework of reinforcement learning
via the teacher's (dialogue partner's) feedback to the dialogue actions from the bot.
 The dialogue takes place in the context of question-answering tasks and the bot has to, given either a short story or a set of facts, answer a set of questions from the teacher.
We consider two types of feedback: explicit numerical rewards as in conventional
reinforcement learning, and textual feedback which is more natural in human dialogue, following
\citep{weston2016dialog}.
We consider two online training scenarios:
(i) where the task is built with a dialogue simulator allowing for easy analysis and repeatability
of experiments; and (ii) where the teachers are real humans using Amazon Mechanical Turk.

    We explore  important issues involved in online learning
   such as  how a bot can be most efficiently trained using a minimal amount of teacher's feedback,
 how a bot can harness different types of feedback signal,
how to avoid pitfalls such as instability during online learing with different types of feedback via
 data balancing and exploration,
and how to make learning with real humans feasible via data batching.
Our findings indicate that
it is feasible to build a pipeline
that starts from a model trained with fixed data and then learns from interactions with humans
to improve itself. 
{\it This chapter is based on the following paper: \newcite{li2016dialogue}, and will be detailed in Section 8}.

\chapter{Background}
\label{back}
In this chapter, we will detail background knowledge on three topics,
namely, \sts models, memory networks, and policy gradient methods of reinforcement learning. 

\section{Sequence-to-Sequence Generation}
\sts models can be viewed as a basic framework for generating a target sentence based on source inputs, which can be adapted to 
a variety of natural language generation tasks,
for example, 
generating a French sentence given an English sentence in machine translation; 
generating a response given a source message in response generation;
generating an answer given a question in question-answering;
generating a short summary given a document in summarization, etc.  

We will first go through the basics of language models, recurrent neural networks, and the Long Short-term Memory, which 
can be viewed as the fundamental components of \sts models. Then we will detail the basic structure of a \sts model. Finally, we will talk about  
 algorithmic  variations of \sts models, such as attention mechanism. 
\subsection{Language Modeling}
Language modeling is a task of predicting which word comes next given the preceding words, which 
an important concept in natural language processing. 
The concept of language modeling can be dated back to the epoch-making work \cite{shannon1951prediction} of Claude Shannon, 
who considered the case
in which a string of input symbols is considered
one by one, and the uncertainty of the next is measured
by counting how difficult it is to guess.\footnote{In the original experiment conducted in \cite{shannon1951prediction}, it is letters, rather than words, that were predicted. } 
More formally, language modeling defines the probability of a sequence (of words)
by individually predicting each word within the sequence given all the preceding words (or history): 
$y=y_1, y_2,...., y_N$, where
  $y_t$ denotes the word token at position $t$ and $N$ denotes the number of tokens in $y$
  \begin{equation}
p(y)=\prod_{t=1}^{t=N} p(y_t|y_1,y_2, ...,y_{t-1})
\label{LM}
\end{equation}
where $p(y_t|y_1,y_2, ...,y_{t-1})$ denotes the conditional probability of seeing word $y_t$ given that all its preceding words, i.e., $y_1,y_2, ...,y_{t-1}$. 
n-gram language models have been widely used, which approximate the history with $n-1$ preceding words. 
The conditional probability is then estimated from relative frequency counts: count the number of times that we see $y_{t-n+1}, ..., y_{t-1}$ and count the number of times 
it is followed by $y_t$:
\begin{equation}
p(y_t|y_1,y_2, ...,y_{t-1})=\frac{C(y_{t-n+1}, ..., y_{t-1},y_t)}{C(y_{t-n+1}, ..., y_{t-1})}
\end{equation}
A variation of  algorithmic variations (e.g., smoothing techniques, model compressing techniques)
 have been proposed  
 \cite{kneser1995improved,rosenfeld2000two,stolcke2002srilm,teh2006hierarchical,federico2008irstlm,federico1996bayesian,chen1996empirical,bacchiani2004language,brants2007large,church2007compressing}. N-gram language modeling comes with the
merit of easy and fast implementation, but suffers from a number of severe issues such as data sparsity, poor generalization to unseen words, 
humongous space requirement for model storage and the incapability to handle long term dependency since the model is only able to consider 4-6  context words. 

Neural language modeling (NLM) offers an elegant way to handle the issues of  data sparsity and the incapability to consider more context words.
It was first proposed in \newcite{bengio2003neural} and improved by many others \cite{morin2005hierarchical,mnih2009scalable,mnih2012fast,le2014distributed,mikolov2010recurrent,graves2013generating,kim2016character}.
 In NLM, each word is represented with a distinct $K$-dimensional distributed vector representation. Semantically similar words occupy similar positions in the vector space, which significantly 
alleviate the data sparsity issue. 
NLM takes as input the vector representations of context words and maps them to a vector representation:
\begin{equation}
h_{t-1}=f (y_1, y_2, ..., y_{t-1})
\end{equation}
where $f$ denotes the mapping function, which is 
usually a feed-forward neural network model  
such as recurrent neural nets or convolutional neural nets, as will be detailed in the following subsection. 
The conditional probability distribution of predicting word $y_t$ given the history context is then given as follows: 
\begin{equation}
\begin{aligned}
p(y_t|y_1,y_2, ...,y_{t-1})&=p(y_t|h_{t-1})\\
s_t&=W [y_t, :]\cdot h_{t-1} \\
p(y_t|h_{t-1})&=\softmax(s_t)
\end{aligned}
\end{equation}
where the weight matrix $W \in \mathbb{R}^{V\times K}$, with $V$ the vocabulary size and $K$ being the dimensionality of the word vector representation. 
$W [y_t, :]$ denotes the $y_t$ th row of the matrix. 
The softmax function maps the scalar vector $s_t$ into a vector of probability distribution as follows:
\begin{equation}
\softmax(s_t)=\frac{\exp{(s_t)}}{\sum_{s\in V}\exp{(s)}}
\end{equation}
Since the dimensionality of the context is immune to the change of context length, theoretically, NLM is able to accommodate infinite number of context words without having to store all 
distinct n-grams.

\subsection{Recurrent Neural Networks}
Recurrent neural networks (RNN) are a neural network architecture which is specifically designed to handle sequential data. It was historically used to handle time sequential data \cite{elman1990finding,funahashi1993approximation}, and have been successfully applied to language processing \cite{mikolov2010recurrent,mikolov2011extensions,mikolov2012statistical,mikolov2012context}.
Given a sequence of word tokens $\{y_1, y_2, ..., y_N\}$, where each word $y_t, t\in [1,N]$ is associated with a K-dimensional vector representation $x_t$.
 RNN associates each time step $t$ with a hidden vector representation $h_t$, which can be thought of as a representation that embeds all information of previous tokens, i.e.,  $\{y_1, y_2, ..., y_{t}\}$. 
$h_t$ is obtained 
using a function $g$
that
 combine the previously built presentation for the previous time-step t-1, denoted as $h_{t-1}$, and the representation for the word of current time-step $x_t$:
\begin{equation}
h_t=g (h_{t-1}, x_t)
\end{equation}
The function $g$ can take different forms, with the simplistic one being as follows:
\begin{equation}
g(h_{t-1},x_t)=\sigma (W_{hh}\cdot h_{t-1}+W_{xh}\cdot x_t)
\label{eq_rnn}
\end{equation}
where $W_{hh}, W_{xh} \in \mathbb{R}^{K\times 2K}$. Popular choices of $\sigma$ are non-linear functions such as sigmoid, $\tanh$ or ReLU. 
From Equ. \ref{eq_rnn}, we can see that the dimensionality of $h_t$ is constant for different $t$. 

\subsection{Long Short Term Memory}
Two serve issues problems with RNNs are the gradient {\it exploding} problem and the gradient {\it vanishing} problem \cite{bengio1994learning}, where gradient {\it exploding}
refers to the situation where the gradients become very large when the error from the training objective function is backpropagated over time, and  gradient {\it vanishing} 
refers to the situation where the gradients approaches zero when the training error is backpropagated over a few time-steps. 
These two issues render RNN models incapable of capturing the long-term dependency for long sequences. 

One of the most effective ways to alleviate these problem is the Long Short Term Memory model, LSTM for short, first  introduced in \newcite{hochreiter1997long}, and adapted 
, used, and further explored 
by many others \cite{graves2005framewise,graves2009offline,chung2014empirical,tai2015improved,xu2015show,kalchbrenner2015grid,cheng2016long,oord2016pixel,jozefowicz2015empirical,greff2017lstm,zaremba2014recurrent}. The key idea of LSTMs is to associate each time step with different types
of gates, and these gates provide flexibility in controlling informational flow: to control how much 
information the current RNN wants to preserve through forget gates; 
to control how much information a RNN want to receive through the input of current time-step through input gates; and how much information a RNN wants to 
output to the next time-step through output gates. 

More formally, 
given a sequence of inputs  $\{y_1, y_2, ..., y_N\}$, where each word $y_t$ is associated with a $K$-dimensional vector representation $x_t$, 
an LSTM associates each time step with an input gate, a memory gate and an output gate, 
respectively denoted as $i_t$, $f_t$, and $o_t$.
$c_t$ is the cell state vector at time $t$, and
$\sigma$ denotes the sigmoid function. Then, the vector representation $h_t$ for each time step $t$ is given by:
\begin{eqnarray}
i_t=\sigma (W_i\cdot [h_{t-1},x_t])\\
f_t=\sigma (W_f\cdot [h_{t-1},x_t])\\
o_t=\sigma (W_o\cdot [h_{t-1},x_t])\\
l_t=\tanh(W_l\cdot [h_{t-1},x_t])\\
c_t=f_t\circ c_{t-1}+i_t\circ l_t\\
h_{t}=o_t\cdot \tanh(c_t)
\end{eqnarray}
where $W_i$, $W_f$, $W_o$, $W_l \in \mathbb{R}^{K\times 2K}$, and $\circ$ denotes the pairwise dot between two vectors. 
Again, as in RNNs, 
$h_t$ is used as a representation for the partial sequence  $\{y_1, y_2, ..., y_{t}\}$.

\subsection{Sequence-to-Sequence Generation}
The \sts model can be viewed as an extension of language model, 
where $y$ is the target sentence, and the  
 prediction of current word $y_t$ in $y$
depends not only on all preceding words $\{y_1, y_2, ..., y_{t-1}\}$, but also on a source input $x$. 
Each sentence concludes with a special end-of-sentence symbol \eos. 
For example, in French-English translation, the English word to predict not only depends on all the preceding words, but also depend on the original French input; 
as another example, the following word in a dialogue response depends both on preceding words in the response and the  message input. 
The \sts model was first proved to yield good performance in machine translation \cite{sutskever2014sequence,bahdanau2014neural,cho2014learning,luong2015effective,luong2014addressing,luong2016achieving,sennrich2015neural,kim2016sequence,wiseman2016sequence,britz2017massive,wu2016google}, 
and has been successfully extended to multiple natural language generation tasks such as text summarization \cite{rush2015neural,see2017get,chopra2016abstractive,nallapati2016abstractive,zeng2016efficient},
parsing \cite{luong2015multi,vinyals2015grammar,jia2016data}, image caption generation \cite{chen2015mind,xu2015show,karpathy2015deep,mao2014deep}, etc. 

More formally, 
in \sts generation tasks, each input $x=\{x_1,x_2,...,x_{n_x}\}$ is paired with a sequence of outputs to predict: 
$y=\{y_1,y_2,...,y_{n_y}\}$. 
A  \sts generation model defines a distribution over outputs and sequentially predicts tokens using a softmax function:
\begin{equation}
\begin{aligned}
p(y|x)
&=\prod_{t=1}^{n_y}p(y_t|y_1,y_2,...,y_{t-1}, x)\\
\end{aligned}
\label{equ-lstm}
\end{equation}
By comparing Eq.\ref{equ-lstm} with the conditional probability of language modeling in Eq. \ref{LM}, we can see that the only difference between them is the additional consideration of the 
source input $x$.

More specifically, a standard \sts model consists of two key components, a encoder, which maps the source input $x$ to a vector representation, and a decoder, which generates an output sequence based on the source sentence. 
Both the encoder and the decoder are multi-layer LSTMs. To enable the encoder to access information from the encoder, the last state memory of the encoder is passed 
to the decoder as the initial memory state, based on which words are sequentially predicted using a softmax function. 
Commonly, input and output use different LSTMs with separate compositional parameters to capture different compositional patterns. 
\subsubsection{Training}
Given a training dataset where each target $y$ is paired with a source $x$, the learning objective is to minimize the negative log-likelihood of predicting each word in the target $y$ given the source $x$:
\begin{equation}
J=-\sum_{t=1}^{t=n_y} \log p(y_t|x,y_1,y_2,...,y_{t-1})
\end{equation}
Parameters including word embeddings and LSTMs' parameters are usually initialized from a uniform distribution and learned 
and optimized using mini-batch stochastic gradient decent with momentum. 
Gradient clipping is usually adopted by  scaling gradients when the norm exceeded a threshold\footnote{Usually set to 5} to avoid gradient explosion.
Learning rate is gradually decreased towards the end of training.\footnote{For example, after 8 epochs, the learning rate is halved every epoch.}

\subsubsection{Testing}
Using a pre-trained model, we need to generate an output sequence $y$ given a new input $x$. The problem can be formalized as a standard search problem: generating 
a sequence of tokens with the largest probability assigned by the pre-trained model, where standard
greedy search and beam search can be immediately used. 
For greedy search, at each time step, the model picks the word with largest probability. 
For beam search with beam size $K$, 
for each time-step, we expand each of the $K$  hypotheses by $K$ children, which gives at $K\times K$ hypotheses. We keep the top $K$ ones, delete the others and move on to the next time-step.

During decoding, the algorithm terminates when an \eos token is predicted.
At each time step, either a greedy approach or beam search can be adopted for word prediction.
Greedy search selects the token with the largest conditional probability, the embedding of which is then combined with preceding output to predict the token at the next step. 

\subsection{Attention Mechanisms}
The standard version \sts model only uses the source representation once, which is through initializing the decoder LSTM using the final state of the encoder  LSTM. 
It is challenging to handle long-term dependency using such a mechanism: the hidden state of the decoder LSTM changes over time as new words are decoded and combined, which 
dilutes the influence from the source sentence.  

One effective way to address such an issue is using attention mechanisms \newcite{bahdanau2014neural,xu2015show,jean2015montreal,luong2015effective,mnih2014recurrent,chorowski2014end}. 
The attention mechanisms
adopt a look-back strategy by
linking the current decoding stage with each input time-step
in an attempt to consider which part of the
input is most responsible for the current decoding time-step. 

More formally, suppose that each time-step of the source input is associated with a vector representation $h_{i}, i\in [1,n_x]$ computed by LSTMs, where $n_x$ denotes the length of the source sentence. 
$h_{i} \in \mathbb{R}^{K\times 1}$.
At the current decoding time-step $t$, attention models would first link the current step
decoding information $h_{t-1}\in \mathbb{R}^{K\times 1}$ with each of the input time step,  characterized by a
strength indicator $v_i$:
\begin{equation}
v_i=h_{t-1}^T\cdot h_{i}
\end{equation}
Other than using the dot product to compute the strength indicator $v_i$, many other mechanisms have been used such as vector concatenation, where $v_i=\tanh(W\cdot [h_{t-1}, h_i])$, $W \in \mathbb{R}^{K\times 2K}$,
or the general dot-product mechanism, where  $v_i=h_{t-1}^T\cdot W\cdot h_{i}$, 
$W \in \mathbb{R}^{K\times K}$,
as detailedly explored in \cite{luong2015effective}. 
$v_t$ is then normalized to a probabilistic value $a_i$ using a softmax function:
\begin{equation}
a_i=\frac{\exp{(v_i)}}{\sum_{i'}\exp(v_{i'})}
\end{equation}
The context vector $ct_{t-1}$ is the weighted sum of hidden memories on the source side:
\begin{equation}
ct_{t-1}=\sum_{i=1}^{i=n_x} a_i\cdot h_{i}
\label{attention}
\end{equation}
As can be seen from Eq. \ref{attention}, a larger value of strength indicator $a_i$ indicates a more contribution to the context vector. 
The vector representation used to predict the upcoming word  $\hat{h}_{t-1}$,  is obtained by combining $ct_{t-1}$ and $h_{t-1}$:
\begin{equation}
\hat{h}_{t-1}=\tanh (\hat{W}[ct_{t-1},h_{t-1}])
\end{equation}
where $\hat{W} \in \mathbb{R}^{K\times 2K}$
\begin{equation}
p(y_t|y_1,y_2,...,y_{t-1},x)=\softmax(W [y_t, :]\cdot \hat{h}_{t-1})
\end{equation}
$W \in \mathbb{R}^{V\times K}$, with $V$ the vocabulary size and $K$ being the dimensionality of the word vector representation.
The context vector $ct_{t-1}$ is not only used to predict the upcoming word $y_t$, but also forwarded to the LSTM operation of the next step \cite{luong2015effective}: 
\begin{eqnarray}
i_t=\sigma (W_i\cdot [h_{t-1},x_t,ct_{t-1}])\\
f_t=\sigma (W_f\cdot [h_{t-1},x_t,ct_{t-1}])\\
o_t=\sigma (W_o\cdot [h_{t-1},x_t,ct_{t-1}])\\
l_t=\tanh(W_l\cdot [h_{t-1},x_t,ct_{t-1}])\\
c_t=f_t\circ c_{t-1}+i_t\circ l_t\\
h_{t}=o_t\cdot \tanh(c_t)
\end{eqnarray}
where $W_i$, $W_f$, $W_o$, $W_l \in \mathbb{R}^{K\times 3K}$.

Up until now, all necessary background knowledge for training a neural \sts model has been covered. 

\section{Memory Networks}
Memory networks \cite{weston2014memory,sukhbaatar2015end} are a class of neural
network
 models that are able to perform natural 
language inference by 
operating on 
 memory components, through which text information can be stored, retrieved, filtered, and reused. The memory components in memory networks
 can embed both long-term memory (e.g., common sense facts about the world) and short-term context (e.g., the last few turns of dialog).
 Memory networks have been successfully applied to many natural language  tasks such as question answering \cite{bordes2014question,weston2015towards}, 
 language modeling \cite{sukhbaatar2015end,hill2015goldilocks} and dialogue \cite{dodge2015evaluating,bordes2016learning}. 
 
\subsubsection{ End-to-End Memory Networks} 
End-to-End Memory Networks (MemN2N) is a  type of memory network model specifically tailored to  natural language inference tasks. 
The input to the MemN2N is a query $x$, for example, a question, along with a set of sentences, denoted by context or memory, $C$=$c_1$, $c_2$, ..., $c_N$, where N denotes
the number of context sentences. 
Given the input $x$ and $C$, the goal is to produce an output/label $a$, for example, the answer to the input question $q$. 

Given this setting, the MemN2N network needs to retrieve useful information from the context, separating information wheat from chaff. 
In the neural network context, 
the query $x$ is first transformed to a vector representation $u_0$ to embed  the information within  the query. 
Such a process can be done using recurrent nets, CNN \cite{krizhevsky2012imagenet,kim2014convolutional}.
For this paper, we adopt 
a simple model that
sums
 up its constituent word embeddings: $u_0=Ax$. The input x is a bag-of-words vector and $A$ is the $d\times V$ word embedding matrix where $d$ denotes the vector dimensionality and $V$ denotes the vocabulary size. Each memory $c_i$ is similarly transformed to vector $m_{i}$.
The model will read information from the memory by linking input representation $q$ with memory vectors $m_{i}$ using softmax weights:
\begin{equation}
o_1=\sum_{i}p_i^1 m_{i} ~~~~~~~~~~p_i^1=\softmax(u_0^T m_i)
\end{equation}

The goal is to select memories relevant to the input query $x$, i.e., the memories with large values of $p_i^1$. The queried memory vector $o_1$ is the weighted sum of memory vectors.
The queried memory vector $o_1$ will be added on top of original input, $u_1=o_1+u_0$.
$u_1$ is then used to query the memory vector.
Such a process is repeated by querying the memory N times (so called ``hops''):
\begin{equation}
o_n=\sum_{i}p_i^n m_{i} ~~~~~~~~~~p_i^n=\softmax(u_{n-1}^T m_i)
\end{equation}
where $n\in [1,N]$.
 One can think of this iterative process as a page-rank way of propagate the influence of the query to the context:
In the first iteration, sentences that semantically related to the query will be assigned higher weights; 
in n$^{\text{th}}$  iteration, sentences that are semantically close the highly weighted sentences in the n-1$^{\text{th}}$ iteration will be assigned higher weights. 

In the end, $u_N$ is input to a softmax function for the final prediction:
\begin{equation}
a=\softmax (u_N^T y_1,u_N^T y_2,...,u_N^T y_L)
\end{equation}
where $L$ denotes the number of candidate answers and $y$ denotes the representation of the answer. If the answer is a word, $y$ is the corresponding word embedding. If the answer is a sentence, $y$ is the embedding for the sentence achieved in the same way as we obtain embeddings for query $x$ and memory $c$.

\section{Policy Gradient Methods}
Policy gradient methods \cite{aleksandrov1968stochastic,williams1992simple} are a type of reinforcement learning model that 
learn the parameters that parametrize policies through the expected reward using gradient decent.\footnote{\url{http://www.scholarpedia.org/article/Policy_gradient_methods}}
By comparison to other reinforcement learning models such as Q-learning, policy gradient methods do not suffer from the problems such as the 
lack of guarantees of a value function (since it does not require an explicit estimation of the value function), or the intractability problem due to continuous states or actions  in high dimensional spaces.

At the current time-step $t$,
policy gradient methods define a probability distribution over all possible actions  to take (i.e., $a_t$) given the current state $s_{t-1}$ and previous actions that have been taken: 
\begin{equation}
a_t\sim p(a_t|a_{1:t-1},s_{t-1})
\end{equation}
The policy distribution $\pi(a_t|a_{1:t-1},s_{t-1})$ is parameterized by  parameters $\Theta$. 
Each action $a_t$ is associated with a reward $r(a_t)$, and we thus have a sequence of action-reward pairs $\{a_t, r(a_t)\}$. 
The goal of policy gradient methods is to optimize the policy parameters so that the expected reward return (denoted by $J(\Theta)$) is optimized, where $J(\Theta)$
can be written as follows:
 \begin{equation}
 \begin{aligned}
 J(\Theta)&=E\{\sum_t r(a_t) \}\\
 &=\sum_t \sum_{a_t}p(a_t)\cdot r(a_t)
\end{aligned}
 \end{equation}
 parameters involved in $ J(\Theta)$ can be optimized through standard gradient decent: 
 \begin{equation}
 \Theta_{h+1}= \Theta_{h}+\alpha \nabla_{\theta} J
 \end{equation}
 where $\alpha$ denotes the learning rate for stochastic gradient decent. 
 
 The major problem involved in policy gradient methods is to obtain a good estimation of the policy gradient $ \nabla_{\theta} J$. One of the most methods to 
 estimate $ \nabla_{\theta} J$ is using the likelihood ratio \cite{glynn1990likelihood,williams1992simple}, better know as the REINFORCE model, with the trick being used 
 as follows:
 \begin{equation}
 \nabla_{\theta} p_{\theta}(a)=p_{\theta}(a) \nabla_{\theta} \log p_{\theta}(a) 
 \end{equation}
 then we have:
  \begin{equation}
  \begin{aligned}
  \nabla_{\theta} J(\theta)&=\sum_a p_{\theta}(a)\nabla_{\theta}\log p_{\theta}(a)r(a) \\
  &=E(\nabla_{\theta}\log p_{\theta}(a)r(a) )
  \end{aligned}
 \end{equation}
In order to reduce the variance of the estimator, a baseline $b$ is usually subtracted from the gradient, i.e., 
  \begin{equation}
  \nabla_{\theta} J(\theta)=E(\nabla_{\theta}\log p_{\theta}(a) [r(a)-b] )
 \end{equation}
where baseline $b$ can be any arbitrarily chosen scalar \cite{williams1992simple}, because it does not introduce bias in the graident:
$\sum_a p_{\theta}(a)=1 \Rightarrow b\sum_a \nabla_{\theta} p_{\theta}(a)=0$. 
Suggested values of baseline $b$ include the mean value of all previously observed rewards,  
optimal estimator as described in
\newcite{peters2008reinforcement}, or estimator output from another neural model \cite{zaremba2015reinforcement,ranzato2015sequence}.

\chapter{Mutual Information to Avoid Generic Responses}
\label{mutual}
When we apply the \sts model to response generation, one severe issue stands out:
neural conversation models tend to generate 
dull responses such
 \textit{I~don't know} or  \textit{I ~don't know what you are talking about} \cite{serban2015hierarchical,vinyals2015neural}. 
From Table \ref{sample:mle}, we can see that
many top-ranked responses are generic. 
Responses that seem more meaningful or specific can also be found in the N-best lists, but rank much lower.
This phenomenon is due to the 
 relatively high frequency of generic responses like 
 \textit{I don't know} in conversational datasets.
 The MLE (maximum likelihood estimation) objective function models the uni-directional dependency from sources to targets, 
 and since 
 dull responses are dull in the similar way and diverse responses are diverse in different ways, the system always generates these dull responses. 
 Intuitively, it seems desirable to take into account not only the dependency of responses on messages, but also the inverse, the likelihood that a message will be provided to a given response:
 it is hard to guess what an input message is about knowing that the response is {\it i don't know}.
 
\begin{table}[!ht]
\center
\small
\begin{tabular}{ll}\toprule
\multicolumn{2}{l}{{\bf Input}: What are you doing?}  \\\midrule
-0.86~~I don't know.&-1.09~~Get out of here.\\
-1.03~~I don't know!&-1.09~~I'm going home. \\
-1.06~~Nothing.& -1.09~~Oh my god! \\
-1.09~~Get out of the way.&-1.10~~I'm talking to you.\\\midrule
\multicolumn{2}{l}{{\bf Input}: what is your name?} \\\midrule 
-0.91~~I don't know. & ... \\
-0.92~~I don't know! & -1.55~~My name is Robert.\\
-0.92~~I don't know, sir.&-1.58~~My name is John. \\
-0.97~~Oh, my god!& -1.59~~My name's John. \\\midrule
\multicolumn{2}{l}{{\bf Input}: How old are you?} \\\midrule 
-0.79~~I don't know. & ... \\
-1.06~~I'm fine.& -1.64~~Twenty-five. \\
-1.17~~I'm all right. & -1.66~~Five.\\
-1.17~~I'm not sure. & -1.71~~Eight. \\\bottomrule
\end{tabular}
\caption[Sample responses  from vanilla \sts neural models]{Responses generated by a 4-layer \sts neural model trained on 20 million conversation pairs take from the OpenSubtitles dataset.
Decoding is implemented with beam size set to 200. The top examples are the responses with the highest average probability log-likelihoods in the N-best list. Lower-ranked, less-generic responses were manually chosen.} 
\label{sample:mle}
\end{table}

We propose to capture this intuition by using Maximum Mutual Information (MMI), as an optimization objective that measures the mutual dependence between inputs and outputs, as opposed to 
the uni-directional dependency from sources to targets in the traditional MLE objective function.
We present practical training and decoding strategies for neural generation models that use MMI as objective function.
We demonstrate that using MMI results in a clear decrease in the proportion of generic response sequences, and 
find a significant performance boost from the proposed models as measured by  BLEU \cite{papineni2002bleu} and human evaluation.

\section{MMI Models}
In the response generation task,
let $x$ denote an input message sequence (source) $x=\{x_1,x_2,...,x_{n_x}\}$ where $n_x$ denotes the number of words in $x$.
Let $y$ (target) denote a sequence in response to source sequence $x$, where $y=\{y_1,y_2,...,y_{n_y},$ \eos{}\}, $n_y$ is the length of the response (terminated by an \eos token). $V$ denotes vocabulary size. 

\subsection{MMI Criterion}
The standard objective function for sequence-to-sequence models is the log-likelihood of target $y$ given source $x$, which at test time yields the statistical decision problem:
\begin{equation}
\hat{y}= \argmax_y \big\{\log p(y|x)\big\}
\label{eqseq2seq}
\end{equation}
As discussed in the introduction, we surmise that this formulation leads to generic responses being generated, since it only selects for targets given sources, not the converse. 
To remedy this, we replace it with Maximum Mutual Information (MMI) as the objective function. 
In MMI, parameters are chosen to maximize (pairwise) mutual information
between the source $x$ and the target $y$:
\begin{equation}
\log \frac{p(x,y)}{p(x)p(y)}
\label{eq1}
\end{equation}

This avoids favoring responses that unconditionally enjoy high probability, and instead biases towards those responses that are specific to the given input.
The MMI objective can be written as follows:\footnote{Note: $\log\frac{p(x,y)}{p(x)p(y)} = \log\frac{p(y|x)}{p(y)} = \log p(y|x) - \log p(y)$}
\begin{equation*}
\begin{aligned}
\hat{y} = \argmax_{y} \big\{\log p(y|x) - \log p(y)\big\}
\end{aligned}
\end{equation*}
We use a generalization of the MMI objective which introduces a hyperparameter $\lambda$ that controls how much to penalize generic responses:
\begin{equation}
\hat{y}  =\argmax_{y} \big\{\log p(y|x) - \lambda\log p(y)\big\}
\label{eqweightedmmi}
\end{equation}

An alternate formulation of the MMI objective uses Bayes' theorem:
\begin{equation*}
\log p(y)=\log p(y|x)+\log p(x)-\log p(x|y)
\label{eqbayes}
\end{equation*}
which lets us rewrite Equation \ref{eqweightedmmi} as follows:
\begin{equation}
\begin{aligned}
\hat{y}
&=\argmax_{y} \big\{(1-\lambda)\log p(y|x)\\
&~~~~~~~~~~~~~~~~+\lambda\log p(x|y)-\lambda\log p(x) \big\} \\[0.2cm]
&=\argmax_{y} \big\{(1-\lambda)\log p(y|x)+\lambda\log p(x|y) \big\}
\label{eqbayesexpanded}
\end{aligned}
\end{equation}
This weighted MMI objective function can thus be viewed as representing a tradeoff between sources given targets (i.e., $p(x|y)$) and targets given sources (i.e., $p(y|x)$). 

We would like to be able to 
adjust the value $\lambda$ in Equation \ref{eqweightedmmi} without repeatedly training neural network models from scratch, which would otherwise be extremely time-consuming.
Accordingly, we did not train a joint model ($\log p(y|x) - \lambda \log p(y)$), but instead trained maximum likelihood models, and used the MMI criterion only during testing.

\subsection{Practical Considerations}
Responses can be generated either from Equation~\ref{eqweightedmmi},  i.e., $\log p(y|x)-\lambda \log p(y)$ or Equation~\ref{eqbayesexpanded}, i.e., $(1-\lambda) \log p(y|x)+\lambda \log p(x|y)$. 
We will refer to these formulations as \mmiLM and \mmiBD, respectively.
However, these strategies are difficult to apply directly to decoding since they can lead to ungrammatical responses 
(with \mmiLM)
or make decoding intractable 
(with \mmiBD).
In the rest of this section, we will discuss these issues and explain how we resolve them in practice. 
 
\subsubsection{\mmiLM}
The second term of $\log p(y|x)-\lambda \log p(y)$
functions as an anti-language model.
It penalizes not only high-frequency, generic responses, but also fluent ones and thus can lead to ungrammatical outputs. 
In theory, this issue should not arise when $\lambda$ is less than~1, since ungrammatical sentences should always be more severely penalized by the first term of the equation, i.e., $\log p(y|x)$.
 In practice, however, we found that the model tends to select ungrammatical outputs that escaped being penalized by $p(y|x)$. 
 
\paragraph{Solution}
Let $n_y$ be the length of target $y$.
$p(y)$ in Equation~\ref{eqweightedmmi} can be written as:
\begin{equation}
p(y)=\prod_{i=1}^{n_y}p(y_i|y_1,y_2,...,y_{i-1})
\end{equation}
We replace the language model $p(y)$ with $U(y)$, which adapts the standard language model by multiplying by a weight $g(i)$ that is decremented monotonically as the index of the current token $i$ increases:
\begin{equation}
\begin{aligned}
U(y)=\prod_{i=1}^{n_y}p(t_i|t_1,t_2,...,t_{i-1})\cdot g(i)
\end{aligned}
\label{eqmonotonic}
\end{equation}
The underlying intuition here is as follows:
First, neural decoding combines the previously built representation with the word predicted at the current step. 
As decoding proceeds, the influence of the initial input on decoding (i.e., the source sentence representation) diminishes as additional previously-predicted words are encoded in the vector representations.\footnote{Attention models \cite{xu2015show} may offer some promise of addressing this issue.}
In other words, the first words to be predicted significantly determine the remainder of the sentence. 
Penalizing words predicted early on by the language model contributes more to the diversity of the sentence than it does to words predicted later. 
Second, as the influence of the input on decoding declines, 
the influence of the language model comes to dominate. 
We have observed that ungrammatical segments tend to appear in the latter part of the sentences, especially in long sentences. 

We adopt the most straightforward form of $g(i)$ by 
by setting up a threshold ($\gamma$) by penalizing the first $\gamma$ words where\footnote{We experimented with a smooth decay in $g(i)$ rather than a stepwise function, but this did not yield better performance.}
\begin{equation}
\begin{aligned}
g(i)
= \left\{
\begin{aligned}
&1~~~~~\text{if}~i\leq\gamma \\
&0~~~~~\text{if}~i>\gamma
\end{aligned}
\right.
\end{aligned}
\end{equation}
The objective Equation \ref{eqweightedmmi} 
can thus be rewritten as: 
\begin{equation}
\begin{aligned}
\log p(y|x) - \lambda \log U(y)
\end{aligned}
\label{eqweightedpos}
\end{equation}
where direct decoding is tractable.
 
\subsubsection{\mmiBD}
Direct decoding from \mmiBDv is intractable, as the second part (i.e., $p(s|t)$) requires completion of 
target generation \textit{before} $p(s|t)$ can be effectively computed.
Due to the enormous search space for target $y$, exploring all possibilities is infeasible. 

For practical reasons, then, we turn to an approximation approach that involves first generating \mbox{N-best} lists given the first part of objective function, i.e., 
standard \sts model $p(t|s)$. 
Then we rerank the \mbox{N-best} lists using the second term of the objective function. 
Since N-best lists produced by \sts models are generally grammatical, the final selected options are likely to be well-formed. 
Model reranking has obvious drawbacks. It results in non-globally-optimal solutions by first emphasizing standard \sts objectives. 
Moreover, it relies heavily on the system's success in generating a sufficiently diverse N-best set, requiring that a long list of N-best lists be generated for each message. 
This assumption is far from valid as one long-recognized issue with beam search is
lack of diversity in the beam: candidates often differ
only by punctuation or minor morphological variations,
with most of the words overlapping. 
The lack
of diversity in the N-best list significantly decreases
the impact of reranking. This means we need a more diverse N-best list for the later re-ranking process. 
\paragraph{Standard Beam-search Decoding}
Here, we first give a sketch of the standard beam-search model and then talk about how we can modify it to produce more diverse N-best lists. 
In standard beam-search decoding, 
 at time step $t-1$ in decoding, the decoder keeps track of $K$ hypotheses, where $K$ denotes the beam size, and their scores $S(y_{1:t-1}|x)=\log p(y_1,y_2,...,y_{t-1}|x)$. As it moves on to time step $t$,
it expands each of the $K$ hypotheses
(denoted as $y_{1:t-1}^k=\{y_1^k,y_2^k,...,y_{t-1}^k\}$, $k\in [1,K]$)
 by selecting the top $K$ candidate expansions,  each expansion
 denoted as  $y_t^{k,k'}$, $k'\in [1,K]$,
  leading to the construction of $K\times K$ new hypotheses:
  $$[Y_{t-1}^k, y_t^{k,k'}], k\in [1,K], k'\in [1,K]$$
The score for each of the $K\times K$ hypotheses is computed as follows:
\begin{equation}
S(y_{1:t-1}^k,y_t^{k,k'}|x)=S(y_{1:t-1},x)+\log p(y_t^{k,k'}|x,y_{1:t-1}^k)
\end{equation}
In a standard beam search model, the top $K$ 
hypotheses are selected
(from the $K\times K$  hypotheses computed in the last step) based on 
the score $S(y_{1:t-1}^k,y_t^{k,k'}|x)$. The remaining hypotheses are ignored when the algorithm proceeds to the next time step.
\begin{figure*}
\includegraphics[width=2in]{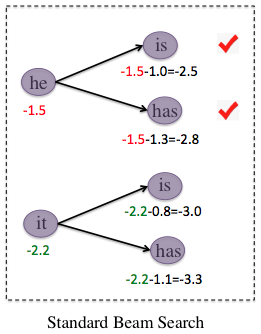}
\includegraphics[width=2.6in]{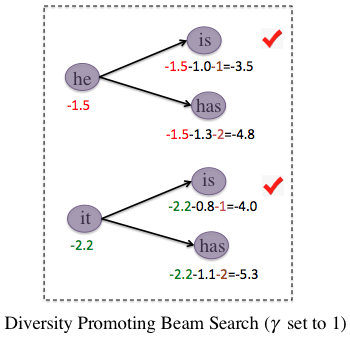}
\centering
\caption[Standard beam search vs the diversity-promoting beam search]{An illustration of standard beam search and the proposed diversity-promoting beam search. $\gamma$ denotes the hyperparameter for penalizing intra-sibling ranking. Scores are made up for illustration purposes. }
\label{figure}
\end{figure*}

\paragraph{Diversity-Promoting Beam Seach}
We propose to increase diversity by changing
the way $S(Y_{t-1}^k,y_t^{k,k'}|x)$ is computed,
as shown in Figure \ref{figure}.
For each of the  hypotheses $Y_{t-1}^k$ ({\it he} and {\it it}),
we generate the top $K$ translations
 $y_t^{k,k'}$, $k'\in [1,K]$ as in the standard beam search model.
Next, we  rank the $K$ translated tokens generated from the same parental hypothesis
based on $p(y_t^{k,k'}|x,Y_{t-1}^k)$ 
in descending order: {\it he is} ranks first among {\it he is} and {\it he has}, and {\it he has} ranks second;
similarly for {\it it is} and {\it it has}.
 
We then rewrite the score for $[Y_{t-1}^k, y_t^{k,k'}]$ by adding an additional term $\gamma k'$,
 where $k'$ denotes the ranking of the current hypothesis among its siblings (1 for {\it he is} and {\it it is}, 2 for {\it he has} and {\it it has}).
\begin{equation}
\hat{S}(Y_{t-1}^k,y_t^{k,k'}|x)=S(Y_{t-1}^k,y_t^{k,k'}|x)-\gamma k'
\label{dive}
\end{equation}
We call $\gamma$ the {\it diversity rate}; it indicates the degree of diversity one wants to integrate into the beam search model.

The top $K$ hypotheses are selected based on $\hat{S}(Y_{t-1}^k,y_t^{k,k'}|x)$ as we move on to the next time step.
By adding the additional term $\gamma k'$, 
the model punishes lower-ranked hypotheses among siblings (hypotheses descended from the same parent).
When we compare newly generated hypotheses descended from different ancestors, the model gives more credit to  top hypotheses from each of the different ancestors.
For instance, even though the original score for {\it it is} is lower than {\it he has},
the model favors the former as the latter is more severely punished by the intra-sibling ranking part $\gamma k'$. 
The model thus generally favors choosing hypotheses from diverse parents, leading to a more diverse N-best list. 
The proposed model is straightforwardly implemented with a minor adjustment to the standard beam search.

\subsection{Training}

Recent research has shown that deep LSTMs work better than single-layer LSTMs for \sts tasks \cite{sutskever2014sequence}.
We adopt a deep structure with four LSTM layers for encoding and four LSTM layers for decoding, each of which consists of a different set of parameters. 
Each LSTM layer consists of 1,000 hidden neurons, and the dimensionality of word embeddings is set to 1,000. 
Other training details are given below, broadly aligned with \newcite{sutskever2014sequence}. 
\begin{itemize}[noitemsep,nolistsep]
\item LSTM parameters and embeddings are initialized from a uniform distribution in [-0.08, 0.08].
\item Stochastic gradient decent is implemented using a fixed learning rate of 0.1. 
\item Batch size is set to 256.
\item Gradient clipping is adopted by  scaling gradients when the norm exceeded a threshold of 1. 
\end{itemize}
Our implementation on a single GPU processes at a speed of approximately 600-1200 tokens per second.\footnote{Tesla K40m, 1 Kepler GK110B, 2880 CUDA cores.} 

The $p(y|x)$ model described in Section 4.3.1 was trained using the same model as that of $p(y|x)$, with messages ($x$) and responses ($y$) interchanged.

\subsection{Decoding}
\subsubsection{\mmiLM}
As described in Section 4.3.1, decoding using \mmiLMv 
 can be readily implemented by predicting tokens at each time-step.
 In addition, we found in our experiments that it is also important to take into account the length of responses in decoding.
 We thus linearly combine the loss function with length penalization, leading to an ultimate score for a given target $T$ as follows:
 \begin{equation}
 Score(T)=p(y|x)-\lambda U(y)+\gamma L_y
 \end{equation}
where $L_T$ denotes the length of the target and $\gamma$ denotes associated weight. We optimize $\gamma$ and $\lambda$ using MERT \cite{och2003minimum} on N-best lists of response candidates. 
The N-best lists  are generated using the decoder with beam size 200.
We set a maximum length of 20 for generated candidates. 
At each time
step of decoding, we are presented with $N\times N$
word candidates. We first add all hypotheses with
an EOS token being generated at current time step
to the N-best list. Next we preserve the top N
unfinished hypotheses and move to next time step.
We therefore maintain batch size of 200 constant
when some hypotheses are completed and taken
down by adding in more unfinished hypotheses.
This will lead the size of final N-best list for each
input much larger than the beam size.

\subsubsection{\mmiBD}
We  generate N-best lists based on $P(T|S)$ and then rerank the list by linearly combining $p(T|S)$, $\lambda p(S|T)$, and $\gamma L_T$. We use MERT \cite{och2003minimum} to tune the 
weights $\lambda$ and $\gamma$ on the development set.

\begin{table*}
\center
\small
\begin{tabular}{lcccc}\toprule
Model                                  &$\#$ of training instances        &\bleu         &{\it distinct-1}      &{\it distinct-2}\\
\sts (baseline)                        & 23M                              & 4.31         & .023               & .107\\
\sts (greedy)                          & 23M                              & 4.51         & .032               & .148\\
\multirow{1}{*}{\mmiLM: \mmiLMv}& \multirow{1}{*}{23M}          &  4.86   & .033              & .175\\
\multirow{1}{*}{\mmiBD: \mmiBDv}&\multirow{1}{*}{23M}& {\bf 5.22} & .051               & .270\\ 
SMT \cite{ritter2011data}              & 50M                              & 3.60         & .098               & .351\\
SMT+neural reranking \cite{sordoni2015neural}& 50M                              & 4.44         & {\bf .101}         & {\bf .358}\\\bottomrule
\end{tabular}
\caption[Performance on the Twitter dataset of \sts models and MMI models]{Performance on the Twitter dataset of 4-layer \sts models and MMI models. {\it distinct-1} and {\it distinct-2} are respectively the number of distinct unigrams and bigrams divided by total number of generated words.}
\label{res:twitter}
\end{table*}
\section{Experiments}
\subsection{Datasets}
\paragraph{Twitter Conversation Triple Dataset} 
We used an extension of the dataset described in 
\newcite{sordoni2015neural}, which consists of 23 million conversational snippets randomly selected from a collection of 129M context-message-response triples extracted from the Twitter Firehose over the 3-month period from June through August 2012.
For the purposes of our experiments, we limited context to the turn in the conversation immediately preceding the message. In our LSTM models, we used a simple input model in which contexts and messages are concatenated to form the source input. 

For tuning and evaluation, we used the development dataset (2118 conversations) and the test dataset (2114 examples), augmented using information retrieval methods to create a multi-reference set. 
The selection criteria for these two datasets included a component of relevance/interestingness, with the result that dull responses will tend to be penalized in evaluation.

\paragraph{OpenSubtitles Dataset} In addition to unscripted Twitter conversations, we also used the OpenSubtitles (OSDb) dataset \cite{tiedemann2009news}, a large, noisy, open-domain dataset containing roughly 60M-70M scripted lines spoken by movie characters. 
This dataset does not specify which character speaks each subtitle line, which prevents us from inferring speaker turns. 
Following Vinyals et al. (2015), we make the simplifying assumption that each line of subtitle constitutes a full speaker turn. Our models are trained to predict the current turn given the preceding ones based on the assumption that consecutive turns belong to the same conversation.
This introduces a degree of noise, since consecutive lines may not appear in the same conversation or scene, and may not even be spoken by the same character.

This limitation potentially renders the OSDb dataset unreliable for evaluation purposes.  
For evaluation purposes, we therefore used data from the Internet Movie Script Database (IMSDB),\footnote{IMSDB (\url{http://www.imsdb.com/}) is a relatively small database of around 0.4 million sentences and thus not suitable for open domain dialogue training.} which explicitly identifies which character speaks each line of the script. 
This allowed us to identify consecutive message-response pairs spoken by different characters. 
We randomly selected two subsets as development and test datasets, each containing 2K pairs, with source and target length restricted to the range of [6,18]. 

\begin{table}
\center
\small
\begin{tabular}{cccc}\toprule
Model&\bleu&{\it distinct-1}&{\it distinct-2}\\\midrule
\sts&7.16&0.0420&0.133\\
\mmiLM&7.60 & 0.0674&0.220\\
\mmiBD&8.26&0.0758&0.288\\\bottomrule
\end{tabular}
\caption[Performance on the OpenSubtitles dataset for MMI models]{Performance on the OpenSubtitles dataset for the \sts baseline and two MMI models.}
\label{res:open}
\end{table}

\subsection{Evaluation}
For parameter tuning and final evaluation,
we used \bleu \cite{papineni2002bleu}, which was shown to correlate reasonably well with human judgment on the response generation task \cite{galley2015deltableu}.
In the case of the Twitter models, we used multi-reference \bleu. 
As the IMSDB data is too limited to support extraction of multiple references, only single reference \bleu was used in training and evaluating the OSDb models.

We did not follow  \newcite{vinyals2015neural} in using perplexity as evaluation metric. 
Perplexity is unlikely to be a useful metric in our scenario, since our proposed model is designed to steer away from the standard \sts model in order to diversify the outputs.   
We report degree of diversity by calculating the number of distinct unigrams and bigrams in generated responses. The value is scaled by total number of generated tokens to avoid favoring long sentences (shown as {\it distinct-1} and {\it distinct-2} in Tables \ref{res:twitter} and \ref{res:open}). 

\subsection{Results}

\paragraph{Twitter Dataset} We first report performance on Twitter datasets in Table~\ref{res:twitter}, along with results
for different models (i.e., {\it Machine Translation} and {\it MT+neural reranking})
reprinted from \newcite{sordoni2015neural} on the same dataset. The baseline is the \sts model with its standard likelihood objective and a beam size of 200. We compare this baseline against greedy-search \sts \cite{vinyals2015neural}, which achieves higher diversity by increasing search errors.
 
{\it Machine Translation} is the phrase-based MT system described in  
\newcite{ritter2011data}. MT features include 
forward and backward maximum
likelihood ``translation'' probabilities, word and
phrase penalties, linear distortion, etc. 
{\it MT+neural reranking} is the phrase-based MT system, reranked using neural models.  
N-best lists are first generated from the MT system.
Recurrent neural models generate scores for N-best list candidates given the input messages.
These generated scores are re-incorporated to rerank all the candidates. 
Additional features to score [1-4]-gram matches between context and response and between message and context (context and message match CMM features) are also employed, as in Sordoni et al. \newcite{sordoni2015neural}.  

{\it MT+neural reranking} achieves a \bleu score of 4.44, which to the best of our knowledge represents the previous state-of-the-art performance on this Twitter dataset.  Note that  
{\it Machine Translation}  and {\it MT+neural reranking}  are trained on a much larger dataset of roughly 50 million examples.
A significant performance boost is observed from \mmiBD over baseline \sts, both in terms of \bleu score and diversity.

\paragraph{OpenSubtitles Dataset}
All models achieve significantly higher \bleu scores on this dataset than on the Twitter dataset, even though the IMSDB data provides only single references for evaluation. 
Table \ref{res:open} shows that the MMI-bidi model
yields a significant performance boost, with a \bleu score increase of up
to $15.3\%$ and a more than $116\%$ jump in unigram diversity. 
Our interpretation of this huge performance improvement is that the
diversity and complexity of input messages lead 
standard \sts models to generate very conservative responses,\footnote{The strings 
\textit{I don't know}, \textit{I don't know what you are talking about},
\textit{I don't think that is a good idea}, and
\textit{Oh my god} constitute
$32\%$ percent of all generated responses.} which fail to match more the interesting reference strings typical of this dataset.
\begin{table*}[!ht]
\center
\footnotesize
\begin{tabular}{p{5cm}p{4.3cm}p{4.3cm} }\toprule
message& {\sts} &MMI $P(T|S)-\lambda P(T)$\\\midrule
I did not get the report from an mi-6 agent.&I don't think that's a good idea. &
You did the right thing, did you?
 \\\midrule
You haven't been given an assignment in this case. &
I don't know what you are talking about.&
I've been looking all over for you. \\\midrule
I'm losing my grip.&
I don't know what you are talking about.&
I'm the only one in the world. \\\midrule
I am ready to help. &
Come on, come on &
I have something we need to talk about.\\\midrule
You programmed me to gather intelligence. That's all I've ever done. &
You know that, don't you? &
You do have fun, don't you? \\\midrule
I mean, we'd have to talk to him.&
I mean, I don't know. &
I mean, he's a good guy.\\\midrule
This has nothing to do with him.&
You don't know anything about him, do you? &
I know him as much.\\\midrule
How come you never say it? & I don't know & Because I don't want to hurt you\\\midrule
He is the one behind this scheme to crash x-net. & I don't know what you are talking about &Yes, he is, isn't he?\\\midrule
How much time do you have here?&I don't know.& Not long enough. Sorry, sir.\\\bottomrule
\end{tabular}
\caption[Sample responses of the OpenSubtitles dataset]{Responses from the \sts baseline and \mmiLM models on the OpenSubtitles dataset.}
\label{out:model1}
\end{table*}

\begin{table*}[!ht]
\center
\footnotesize
\begin{tabular}{p{5cm}p{4.3cm}p{4.3cm} }\toprule
message & {\sts} & \mmiBD\\\midrule
Ohh I've never seen that! How long does it take you guys to learn the drill?& I don't even know.&Like 2 weeks ago!!\\\midrule
You'll know when we sign Heskey in 2 weeks&I'll see what I can do&
I sure hope so..\\\midrule
I'm waiting!! I have even got biscuits waiting for you??&
I have no idea what you're talking about&
I'll send you a pic when I get home xx \\\midrule
My parents actually had touch too much played at their wedding.&
I have no idea what you're talking about&
How old are they?? \\\midrule
Say anything you want... I know you love me and I love you.&I love you so much.& I love you too, baby. \\\midrule
I am off all week next week & What time you going? & What time you going?\\\midrule
How are you doing? & I'm good, thanks.&I'm good, you?\\\bottomrule
\end{tabular}
\caption[Sample responses of the Twitter dataset]{Responses from the \sts baseline and \mmiBD models on the Twitter dataset.}
\label{out:model2}
\end{table*}

\begin{table}[tbp]
\small
\centering
\begin{tabular}{@{}lcc@{}}
\toprule
{\bf Comparator} & {\bf Gain} & {\bf 95\% CI}\\
\midrule
SMT \cite{ritter2011data} & 0.29 & [0.25, 0.32] \\
SMT+neural reranking & 0.28 & [0.25, 0.32] \\
\sts (baseline) & 0.11 & [0.07, 0.14] \\
\sts (greedy) & 0.08 & [0.04, 0.11] \\
\bottomrule
\end{tabular}
\caption[Human evaluation of the MMI model]{\mmiBD gains over comparator systems, based on pairwise human judgments.}
\label{tab:humanscores}
\end{table}

\begin{table}[!ht]
\setlength{\tabcolsep}{4pt}
\center
\small
\begin{tabular}{ll}\toprule
\multicolumn{2}{l}{{\bf Input}: What are you doing?} \\\midrule
1. I've been looking for you.&4. I told you to shut up.\\
2. I want to talk to you. &5. Get out of here.\\
3. Just making sure you're OK.&6. I'm looking for a doctor. \\\midrule
\multicolumn{2}{l}{{\bf Input}: What is your name? }\\\midrule
1. Blue! & 4. Daniel. \\
2. Peter. &5. My name is John. \\
3. Tyler. &6. My name is Robert. \\\midrule
\multicolumn{2}{l}{{\bf Input}: How old are you?} \\\midrule
1. Twenty-eight. & 4. Five.\\
2. Twenty-four. & 5. 15.\\
3. Long.& 6. Eight.\\\bottomrule
\end{tabular}
\caption{Examples generated by the \mmiLM model on the OpenSubtitles dataset.} 
\label{sample:mmi}
\end{table}

\paragraph{Qualitative Evaluation}

We employed crowdsourced judges to provide evaluations for a random sample of 1000 items in the Twitter test dataset. 
Table \ref{tab:humanscores} shows the results of human evaluations between paired systems. 
Each output pair was ranked by 5 judges, who were asked to decide which of the two outputs was better. 
They were instructed to prefer outputs that were more specific (relevant) to the message and preceding context, as opposed to those that were more generic. 
Ties were permitted. 
Identical strings were algorithmically assigned the same score. 
The mean of differences between outputs is shown as the gain for \mmiBD over the competing system.
At a significance level of $\alpha = 0.05$, we find that \mmiBD outperforms both baseline and greedy \sts systems, as well as the weaker SMT and SMT+RNN baselines. 
\mmiBD outperforms SMT in human evaluations \textit{despite} the greater lexical diversity of MT output.

Separately, judges were also asked to rate overall quality of \mmiBD output over the same 1000-item sample in isolation, each output being evaluated by 7 judges in context using a 5-point scale. The mean rating was 3.84 (median: 3.85, 1st Qu: 3.57, 3rd Qu: 4.14), suggesting that overall \mmiBD output does appear reasonably acceptable to human judges.

Table~\ref{sample:mmi} presents the N-best candidates generated using the \mmiBD model for the inputs of Table~\ref{sample:mle}. We see that MMI generates significantly more interesting outputs than \sts. 
 
In Tables~\ref{out:model1} and \ref{out:model2}, we present responses generated by different models.
All examples were randomly sampled (without cherry picking).
We see that the baseline \sts model tends to generate reasonable responses to simple messages such as 
\textit{How are you doing?} or \textit{I love you}. 
As the complexity of the message increases, however, the outputs switch to more conservative, duller forms, such as 
\textit{I don't know} or \textit{I don't know what you are talking about}.
An occasional answer of this kind might go unnoticed in a natural conversation, but a dialog agent that \textit{always} produces such responses risks being perceived as uncooperative. 
\mmiBD models, on the other hand, produce far more diverse and interesting responses.
\section{Conclusions}
In this chapter, 
we investigated the issue
that \sts models 
 tend to generate safe, commonplace responses (e.g., \textit{I don't know}) regardless of the input. 
Our analysis suggests that the issue is at least in part attributable to the use of 
unidirectional likelihood of output (responses) given input (messages).
To remedy this, we have proposed using Maximum Mutual Information (MMI) as the objective function.
Our results demonstrate that the proposed MMI models produce more diverse and interesting responses, while improving quality as measured by \bleu and human evaluation.

\chapter{Addressing the Speaker Consistency Issue}
\label{persona}
One severe issue for current chit-chat systems is their lack of speaker consistency. 
Since the training set contains conversations from different people, and a decoding model always selects the response with the greatest likelihood,
outputs can be wildly inconsistent, as illustrated in Table \ref{sample:baseline}. 
\begin{table}
\center
\begin{tabular}{rlrl}\toprule
m: & Where do you live now?&m:  & Where were you born?\\
r: & I live in Los Angeles.&r: & I was born in Canada.  \\
m:  & In which city do you live now?& m:  & Where are you from?\\
r: & I live in Madrid.& r: & England, you?\\
m:  & In which country do you live now?&m:  & Where did you grow up?\\
r: & England, you?& r: & I grew up in Texas.\\\midrule
m:  & How old are you?&m:  & What is your major? \\
r: & 16 and you?& r:&I'm majoring in psychology \\
m:  & What's your age?&m:  & What did you study in college?\\
r: & 16. &r:& I studied psychology. \\\bottomrule
\end{tabular}
\caption[Inconsistent responses from \sts models]{Inconsistent responses generated by a 4-layer \sts model trained on 25 million Twitter conversation snippets. m denotes input message and r denotes the generated response.}
\label{sample:baseline}
\end{table}

In this chapter, we 
talk about how we can 
address the challenge of consistency and how to endow data-driven systems with the coherent ``persona'' needed to model human-like behavior, whether as personal assistants, personalized avatar-like agents, or game characters.\footnote{\newcite{vinyals2015neural} suggest that the lack of a coherent personality makes it impossible for current systems to pass the Turing test.} 
For present purposes, we will define \textsc{persona} as the character that an artificial agent, as actor, plays or performs during conversational interactions.
A~persona can be viewed as a composite of elements of identity (background facts or user profile), language behavior, and interaction style. 
A persona is also adaptive, since an agent may need to present different facets to different human interlocutors depending on the demands of the interaction.

We explore two persona models, a single-speaker \textsc{Speaker Model} and a dyadic \textsc{Speaker-Addressee Model}, within the \sts framework. 
The Speaker Model integrates a speaker-level vector representation into the target part of the \sts model.
Analogously, the Speaker-Addressee model encodes the interaction patterns of two interlocutors by constructing an interaction representation from their individual embeddings and incorporating it into the \sts model. 
These persona vectors are trained on human-human conversation data and used at test time to generate personalized responses.
Our experiments on an open-domain corpus of Twitter conversations and dialog datasets comprising TV series scripts show that leveraging persona vectors can improve relative performance up to $20\%$ in \bleu score and $12\%$ in perplexity, with a commensurate gain in consistency as judged by human annotators.

\begin{figure*} [!ht]
\centering
\includegraphics[width=6in]{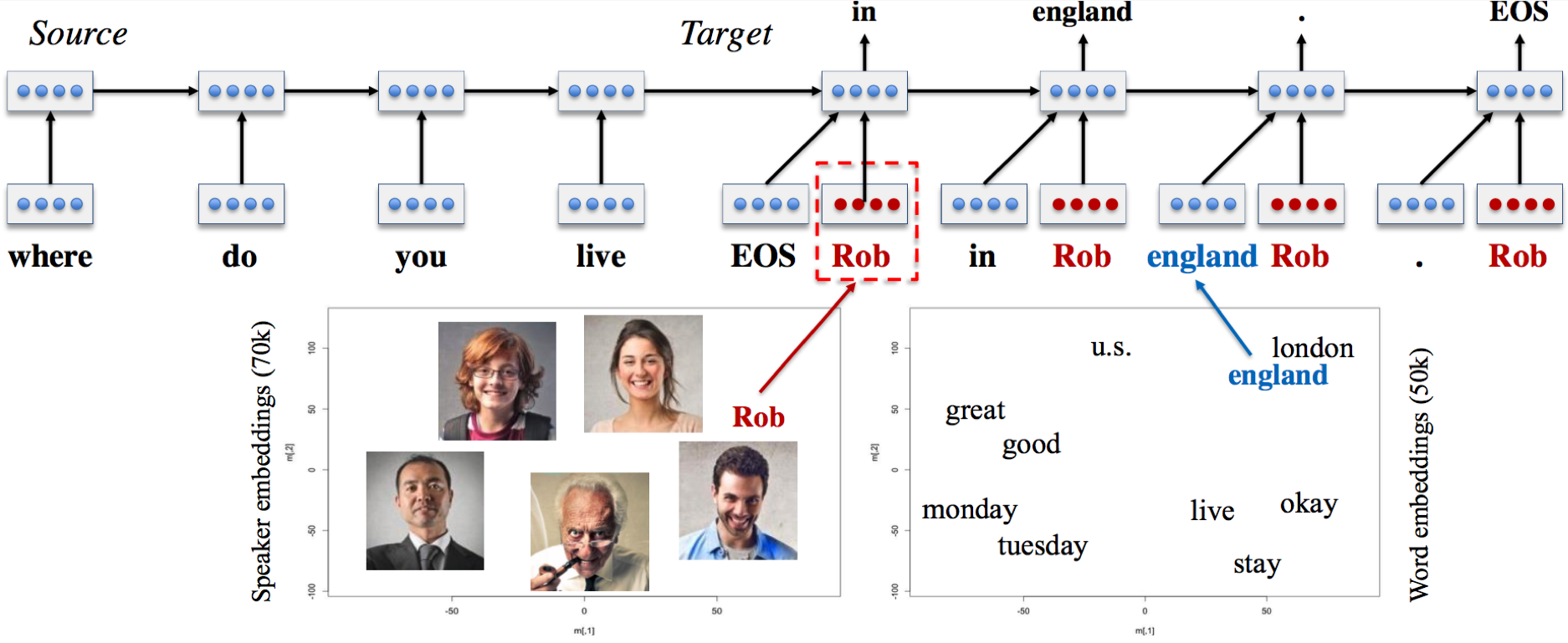}
\caption[The persona speaker model]{Illustrative example of the Speaker Model introduced in this work. Speaker IDs close in embedding space tend to respond in the same manner. These speaker embeddins are learned jointly with word embeddings and all other parameters of the neural model via backpropagation. 
In this example, say Rob is a speaker clustered with people who often mention England in the training data, then the generation of the token `england' at time \mbox{$t=2$} would be much more likely than that of `u.s.'. A non-persona model would prefer generating {\it in the u.s.} if `u.s.' is more represented in the training data across all speakers.
}\label{fig1}
\end{figure*}

\section{Model}
\subsection{Speaker Model}
Our first model is the Speaker Model, which 
models the respondent alone.
This model represents each individual speaker as a vector or embedding, which encodes 
speaker-specific information (e.g., dialect, register, age, gender, personal information) that influences the content and style of her responses.\footnote{Note that these attributes are not explicitly annotated, which would be tremendously expensive for our datasets. Instead, our model manages to cluster users along some of these traits (e.g., age, country of residence) based on responses alone.}

Figure \ref{fig1} gives a brief illustration of the Speaker Model. 
Each speaker $i\in [1,N]$ is associated with a user-level representation $v_i\in\mathbb{R}^{K\times 1}$. 
 As in standard \sts models, we first encode message  $S$ into a vector representation $h_S$ using the source LSTM. 
Then for each step in the target side, hidden units are obtained by 
combining the representation produced by the target LSTM at the previous time step (i.e., $h_{t-1}$), the word representations at the current time step $x_t$, and the speaker embedding $v_i$:
\begin{equation}
\left[
\begin{array}{lr}
i_t\\
f_t\\
o_t\\
l_t\\
\end{array}
\right]=
\left[
\begin{array}{c}
\sigma\\
\sigma\\
\sigma\\
\tanh\\
\end{array}
\right]
W\cdot
\left[
\begin{array}{c}
v_i\\
h_{t-1}\\
x_t\\
\end{array}
\right]
\end{equation}
\begin{equation}
c_t=f_t\cdot c_{t-1}+i_t\cdot l_t
\end{equation}
\begin{equation}
h_{t}=o_t\cdot \tanh(c_t)
\end{equation}
where $W\in \mathbb{R}^{4K\times 3K}$. 
In this way, speaker information is encoded and 
injected into the hidden layer at each time step and thus helps predict personalized responses throughout the generation process.
The Speaker embedding $\{v_i\}$ is shared across all conversations that involve speaker $i$.  $\{v_i\}$ are learned by back propagating word prediction errors to each neural component during training. 

Another helpful property of this model is that it 
helps {\it infer} answers to questions even if the evidence is not readily present in the training set.
This is important as
the training data does not contain explicit 
information about every
attribute of each user
(e.g., gender, age, country of residence).

The model learns speaker representations based on conversational content produced by different speakers, and speakers producing similar responses tend to have similar embeddings, occupying nearby positions in the vector space. 
This way, the training data of speakers nearby in vector space help increase the generalization capability of the
speaker model. For example, consider two speakers $i$ and $j$
who sound distinctly British, and who are therefore close in speaker 
embedding space. Now, suppose that, in the training data, speaker $i$ was asked {\it Where do you live?} and responded {\it in the UK}. Even if speaker $j$ was never asked the same question, this answer can help influence a good response from speaker $j$, and this without any explicitly labeled geo-location information.
  
\subsection{Speaker-Addressee Model}
A natural extension of the Speaker Model is a model that is sensitive to speaker-addressee interaction patterns within the conversation. Indeed,
speaking style, register, and content does not only vary with the identity of the speaker, but also with that of the addressee.
For example, in scripts for the TV series {\it Friends} used in some of our experiments, the character Ross often 
talks differently to his sister Monica than to Rachel,
with whom he is engaged in a on-again off-again relationship throughout the series. 

The proposed Speaker-Addressee Model operates as follows:
We wish to predict how speaker $i$ would respond to a message produced by speaker $j$. Similarly to the Speaker model, we associate each speaker with a $K$ dimensional speaker-level representation, namely $v_i$ for user $i$ and $v_j$ for user $j$. 
We obtain an interactive representation $V_{i,j}\in \mathbb{R}^{K\times 1}$ by linearly combining user vectors $v_i$ and $v_j$
in an attempt to model the interactive style of user $i$ towards user $j$,
\begin{equation}
V_{i,j}=\tanh(W_1\cdot v_i+W_2\cdot v_2)
\end{equation}
where $W_1, W_2\in \mathbb{R}^{K\times K}$. 
$V_{i,j}$  is then linearly incorporated into LSTM models at each step in the target: 
\begin{equation}
\left[
\begin{array}{lr}
i_t\\
f_t\\
o_t\\
l_t\\
\end{array}
\right]=
\left[
\begin{array}{c}
\sigma\\
\sigma\\
\sigma\\
\tanh\\
\end{array}
\right]
W\cdot
\left[
\begin{array}{c}
h_{t-1}\\
w_{t}\\
V_{i,j}\\
\end{array}
\right]
\end{equation}
\begin{equation}
c_t=f_t\cdot c_{t-1}+i_t\cdot l_t\\
\end{equation}
\begin{equation}
h_{t}=o_t\cdot \tanh(c_t)
\end{equation}
$V_{i,j}$ 
depends on both speaker and addressee and
the same speaker will thus respond differently to a message from different interlocutors. 
One potential issue with Speaker-Addressee modelling is the difficulty involved in collecting a large-scale training dataset in which each speaker 
is involved in conversation with a wide variety of people. Like the Speaker Model, however, the Speaker-Addressee Model derives generalization capabilities from speaker embeddings.
Even if the two speakers
at test time ($i$ and $j$) were never involved in the same conversation in the training data, two speakers $i'$ and $j'$ who are respectively close in embeddings may have been, and this can help  modelling how $i$ should respond to $j$. 

\subsection{Decoding and Reranking}
For decoding, 
the N-best lists  are generated using the decoder with beam size \mbox{$K=200$}.
We set a maximum length of 20 for the generated candidates. 
To deal with the issue that \sts models tend to generate generic and commonplace responses such as {\it I don't know}, we follow \newcite{li2015diversity} by reranking the generated N-best list 
using a scoring function that linearly combines 
 a length penalty and the log likelihood of source given target:
\begin{equation}
\log p(y|x,v)+\lambda\log p(x|y)+\gamma L_y 
\end{equation}
where $p(y|x,v)$ denotes the probability of the generated response given the message $x$ and the  respondent's speaker ID. 
$L_y$ denotes the length of the target and $\gamma$ denotes the associated penalty weight. We optimize $\gamma$ and $\lambda$ on N-best lists of response candidates generated from the development set using MERT \cite{och2003minimum} by optimizing \bleu.
To compute $p(x|y)$, 
we train an inverse \sts model by swapping messages and responses. We trained standard \sts models for $p(x|y)$  with no speaker information considered.

\section{Experiements}
Following \cite{sordoni2015neural,li2015diversity}
we used \bleu \cite{papineni2002bleu} 
for parameter tuning and evaluation. 
Besides \bleu scores, we also report perplexity, which has been widely adopted as an indicator of model capability. 

\begin{table}
\centering
\begin{tabular}{lc}\toprule
System                                    & \bleu \\ \midrule
MT baseline \cite{ritter2011data}         & 3.60\% \\ 
Standard LSTM MMI \cite{li2015diversity}  & 5.26\% \\
Standard LSTM MMI (our system)            & 5.82\% \\ 
{\it Human}                               & {\it 6.08\%}\\\bottomrule
\end{tabular}
\caption[Results for the baseline model]{\bleu on the Twitter Sordoni dataset (10 references). We contrast our baseline against an SMT baseline \cite{ritter2011data}, and the best result \cite{li2015diversity} on the established
dataset of \cite{sordoni2015neural}.
The last result is for a human oracle, but it is not directly comparable as the oracle \bleu is computed in a leave-one-out fashion, having one less reference available. We nevertheless provide
this result to give a sense that these \bleu scores of 5-6\% are not unreasonable.}
\label{twitter-baselines}
\end{table}

\subsection{Baseline}
Since our main experiments are with a new dataset (the Twitter Persona Dataset), we first show that our LSTM baseline is competitive with the state-of-the-art \cite{li2015diversity} on an established
dataset, the Twitter  Dataset \cite{sordoni2015neural}.
Our baseline is simply our implementation of the LSTM-MMI of \cite{li2015diversity}, so results should be relatively close to their reported results.
Table~\ref{twitter-baselines} summarizes our results against prior work.
We see that our system actually does better than \cite{li2015diversity}, and
we attribute the improvement to a larger training corpus, the use of dropout during training, and possibly to the ``conversationalist'' nature of our corpus.

\begin{table}
\centering
\begin{tabular}{ccc}\toprule
Model&Standard LSTM&Speaker Model \\\midrule
Perplexity&47.2&42.2 ($-10.6\%$) \\\bottomrule
\end{tabular}
\caption[Perplexity for the persona model on the Twitter dataset]{Perplexity for standard \sts and the Speaker model on the development set of the Twitter Persona dataset.}
\label{twitter-per}
\end{table}
\begin{table}
\centering
\begin{tabular}{lll}\toprule
Model&Objective& \bleu \\\midrule
Standard LSTM &MLE& 0.92\% \\
Speaker Model & MLE&1.12\%  \\
Standard LSTM &MMI& 1.41\% \\
Speaker Model & MMI&1.66\%\\\bottomrule
\end{tabular}
\caption[\bleu for the persona model on the Twitter dataset]{
\bleu on the Twitter Persona dataset (1 reference), for the
standard \sts model and the Speaker model using as objective either maximum likelihood (MLE) or maximum mutual information (MMI).}
\label{twitter-bleu}
\end{table}

\begin{table*}
\centering
\begin{tabular}{cccc}\toprule
Model&Standard LSTM&Speaker Model& Speaker-Addressee Model \\
Perplexity&27.3&25.4 ($-7.0\%$)& 25.0 ($-8.4\%$)  \\\bottomrule
\end{tabular}
\caption[Perplexity for the persona model on the TV series dataset]{Perplexity for standard \sts and persona models on the TV series dataset.}
\label{tv-per}
\end{table*}
\begin{table*}
\centering
\begin{tabular}{ccccc}\toprule
Model&Standard LSTM&Speaker Model& Speaker-Addressee Model \\\midrule
MLE&1.60\%& 1.82\% ($+13.7\%$)& 1.83\%  ($+14.3\%$) \\
MMI&1.70\%& 1.90\% ($+10.6\%$) &1.88\% ($+10.9\%$) \\\bottomrule
\end{tabular}
\caption[\bleu for the persona model on the TV series dataset]{
\bleu on the TV series dataset (1 reference), for the
standard \sts and persona models.}
\label{tv-bleu}
\end{table*}

\subsection{Results}
We first report performance on the Twitter Persona dataset.
Perplexity is reported in Table \ref{twitter-per}. We observe about a $10\%$ decrease in perplexity for the Speaker model over the standard \sts model. 
In terms of \bleu scores (Table~\ref{twitter-bleu}), a significant performance boost 
is observed for 
 the Speaker model over the standard \sts model, yielding an increase of $21\%$
in the maximum likelihood (MLE) setting and $11.7\%$ for mutual information setting (MMI). 
In line with findings in \cite{li2015diversity}, we observe a consistent performance boost introduced by 
the MMI objective function 
over a standard \sts model based on the MLE objective function. 
It is worth noting that our persona models are more beneficial to the MLE models
than to the MMI models. This result is intuitive as the persona models help make 
Standard LSTM MLE outputs more informative and less bland, and thus make the use 
of MMI less critical.

For the TV Series dataset,  perplexity and \bleu  scores are respectively  reported in Table \ref{tv-per} and Table \ref{tv-bleu}.
As can be seen, the Speaker and Speaker-Addressee models respectively achieve perplexity values of 25.4 and 25.0 on the TV-series dataset, $7.0\%$ and $8.4\%$ percent lower than the correspondent standard \sts models. 
In terms of \bleu score,  we observe a similar performance  boost as on the Twitter dataset, in which the Speaker model
and the Speaker-Addressee model 
 outperform the standard \sts model by $13.7\%$ and $10.6\%$.
By comparing the Speaker-Addressee model against the Speaker model on the TV Series dataset, we do not observe a significant difference.  
We suspect that this is primarily due to the relatively small size of the dataset where the interactive patterns might not be fully captured. 
Smaller values of perplexity are observed for the Television Series dataset than the Twitter dataset, the perplexity of which is over 40, presumably due to the more noisy nature of Twitter dialogues. 
\subsection{Qualitative Analysis}

\begin{table}
\setlength{\tabcolsep}{4pt}
\center
{\small
\begin{tabular}{rlrl} \toprule
\Message   & Where is your hometown?&\Message   & What company do you work for? \\\midrule
{\it baseline}    & I was born in Canada.&{\it baseline}    & I work for a company.\\\midrule
\User{1}   & I'm from England.&\User{1}   & I don't have a job.\\
\User{2}   & I'm from Manchester.& \User{2}   & I'm a manager.\\
\User{3}   & I'm from Liverpool.&\User{3}   & I'm working. {\color{red} $\#$}\\
\User{4}   & England. You?&\User{4}   & I work for a company\\
\User{5}   & My hometown.&\User{5}   & I'm working. {\color{red} $\#$}\\
\User{6}   & I'm from Texas.&\User{6}   & Customer service.\\
\User{7}   & I'm from LA.&\User{7}   & I work in retail.\\
\User{8}   & England.&\User{8}   & I work in a cafe.\\
\User{9}   & I was born here.&\User{9}   & I work at a gas station.\\
\User{10}  & I was born in the us.&\User{10}  & I don't know {\color{red} $\#$}\\\bottomrule
\end{tabular}
}
\caption[Sample responses from the persona model on the Twitter dataset]{Responses generated by the baseline (LSTM-MMI) and the Speaker Model for ten randomly selected users, without cherry picking. {\color{red} $\#$} indicates poor-quality responses produced by the system.}
\label{example1}
\end{table}

\begin{table}[!ht]
\setlength{\tabcolsep}{4pt}
\center
{\small
\begin{tabular}{cc} \toprule
\Message   & Do you love me ?\\\midrule
{\it baseline} & of course, i love you .  \\\midrule
& {\it addressee {\it Emily},  speaker Sheldon }\\
{\it response} & of course, i love you, emily .\\\midrule
& {\it addressee {\it Emily},  speaker Penny }\\
{\it response} & of course, i love you, emily .\\\midrule
& {\it addressee {\it Lenard},  speaker Penny }\\
{\it response} & of course, i love you, Lenard .\\\midrule
& {\it addressee {\it Monica},  speaker  Chandler }\\
{\it response} & of course i love you . i want to marry you .\\\midrule
& {\it addressee {\it  Rachel},  speaker  Chandler }\\
{\it response} & of course i love you .\\\midrule
& {\it addressee {\it  Ross},  speaker  Chandler }\\
{\it response} & of course i love you .\\\midrule
& {\it addressee {\it  Ross},  speaker  Rachel }\\
{\it response}  &of course i love you . ( kisses him ) \\\midrule
& {\it addressee {\it  Emily},  speaker  Rachel }\\
{\it response}  &of course i love you . \\\bottomrule
\end{tabular}
}
\caption[Sample responses from the addressee-speaker model on the TV-series dataset]{Responses to {\it do you love me} from the {\it addressee-speaker} model on the TV-series dataset using different addressees and speakers.}
\label{addressees}
\end{table}

\paragraph{Diverse Responses by Different Speakers}
Table \ref{example1} represents responses generated by persona models in response to three different input questions. We randomly selected 10 speakers (without cherry-picking) from the original Twitter dataset. We collected their user level representations from a speaker look-up table and integrated them into the decoding models.  We can see that the model tends to generate specific responses for different people in response to the factual questions.\footnote{There appears to be a population bias in the training set that favors British users.}  

Table \ref{addressees} represents responses generated from the {\it Speaker-Addressee Model} using the TV-series dataset. Interestingly, we do observe the subtle difference captured by the model when it 
attempts to respond to different 
addresses . For example, the model produces {\it of course, i love you, emily .} in response to input from addressee{\it Emily} by generating the addressee's name. 
Also, the model generates {\it of course i love you . ( kisses him )} in response to female addressees. 
 
\paragraph{Human Evaluation} We conducted a human evaluation of outputs from the Speaker Model, using 
a crowdsourcing service. 
Since we cannot expect crowdsourced human judges to know or attempt to learn the ground truth of Twitter users who are not well-known public figures, we designed our experiment to evaluate the consistency of outputs associated with the speaker IDs. To this end, we collected 24 pairs of questions for which we would expect responses to be consistent if the persona model is coherent.  For example, responses to the questions {\it What country do you live in?} and {\it What city do you live in?} would be considered consistent if the answers were {\it England} and {\it London} respectively, but not if they were {\it UK} and {\it Chicago}.  Similarly, the responses to {\it Are you vegan or vegetarian?} and {\it Do you eat beef?} are consistent if the answers generated are {\it vegan} and {\it absolutely not}, but not if they are {\it vegan} and {\it I love beef}.  We collected the top 20 pairs of outputs provided by the Speaker Model for each question pair (480 response pairs total). We also obtained the corresponding outputs from the baseline MMI-enhanced \sts system. 

Since our purpose is to measure the gain in consistency over the baseline system, we presented the pairs of answers system-pairwise, i.e., 4 responses, 2 from each system,  displayed on the screen.
Then we 
 asked judges to decide which of the two systems was more consistent.  The position in which the system pairs were presented on the screen was randomized.  Five judges rated each pair.  A system was assigned a score 1.0 if it was judged much more consistent than the other, 0.5 if mostly more consistent, zero otherwise. Ties (where the two systems are equally consistent or inconsistent) were discarded.  A maximum score of 5.0 was possible. After reweighting for frequency of response  pairs, the mean relative consistency rating for the Speaker Model was 1.33 compared to 0.99 for the baseline model, representing a gain in relative consistency of 34.68\%. It should be stressed that the latter is a strong baseline, since it represents the consensus of all 70K Twitter users in the dataset. ({\it I'm not pregnant} is an excellent consensus answer to the question {\it Are you pregnant?}, while {\it I'm pregnant} is consistent as a response only in the case of someone who also answers the question {\it Are you a guy or a girl?} with something in the vein of {\it I'm a girl}.)

\paragraph{Speaker Consistency}
Table \ref{example2} illustrates how consistency is an emergent property of two arbitrarily selected users. The model is capable of discovering the relations between different categories of location such as London and the UK, Jakarta and Indonesia. However, the model also makes inconsistent response decisions, generating different answers in the second example in response to questions asking about age or major. 
Our proposed persona models integrate user embeddings into the LSTM, and thus can be viewed as encapsulating a trade-off between a persona-specific generation model and a general conversational model.

\begin{table}[!ht]
\setlength{\tabcolsep}{4pt}
\center
{\small
\begin{tabular}{rlrl} \toprule
{\bf User1} &&{\bf User2}&\\\midrule
\Message    & Where is your hometown?&\Message    & Where is your hometown?\\
\Response   & I'm from England.& \Response   & I live in Indonesia.\\\
\Message    & Where are you from?&\Message    & Where are you from? \\
\Response   & I'm from England.&\Response   & England, you? \\
\Message    & In which city do you live now?&\Message    & In which city do you live now?\\
\Response   & I live in London.& \Response   & I live in Jakarta.\\
\Message    & In which country do you live now?&\Message& In which country do you live now? \\
\Response   & I live in England.&\Response   & I live in Indonesia. \\\midrule
\Message    & What is your major?&\Message    & What is your major?\\
\Response   & Business. You?& \Response   & Business, you?\\
\Message    & What did you study in college?&\Message    & What did you study in college?\\
\Response   & I did business studies.& \Response   & Psychology, you?\\\midrule
\Message    & How old are you?&\Message    & How old are you? \\
\Response   & I'm 18.&\Response   & I'm 18.\\
\Message    & What is your age?&\Response& What is your age?\\
\Response   & I'm 18. &\Response&  I'm 16.\\\bottomrule
\end{tabular}
}
\caption{Examples of speaker consistency and inconsistency generated by the Speaker Model}
\label{example2}
\end{table}
\section{Conclusion}
In this chapter,
we have presented two persona-based response generation models for open-domain conversation generation. 

Although the gains presented by our new models are not spectacular, the systems nevertheless outperform our baseline \sts systems in terms of \bleu, perplexity, and human judgments of speaker consistency. 
We have demonstrated that by encoding personas into distributed representations, we are able
to capture certain personal characteristics such as speaking style and background information. 
In the Speaker-Addressee model, moreover, the evidence suggests that there is benefit in capturing dyadic interactions. 

Our ultimate goal is to be able to take the profile of an arbitrary individual whose identity is not known in advance, and generate conversations that accurately emulate that individual's persona in terms of linguistic response behavior and other salient characteristics. 
Such a capability will dramatically change the ways in which we interact with dialog agents of all kinds, opening up rich new possibilities for user interfaces. 
Given a sufficiently large training corpus in which a sufficiently rich variety of speakers is represented, this objective does not seem too far-fetched.



\chapter{Fostering Long-term Dialogue Success}
\label{RL}
In the previous two sections, 
we have talked about how a chit-chat system can avoid general responses and produce consistent responses regarding different questions. 
So far, we have only been concerned with the quality of single-turn responses, but this is actually an overly simplified approximation  
of humans' conversations.
Humans' conversations usually consist of tens, or even hundreds of turns.
These multi-turn conversations usually 
 have structures, e.g., starting with an opener, setting up contexts, 
getting to the points, etc, and humans are good at managing the informational flow of a conversation, that lead to the long-term success (including overall coherence, meaningfulness, etc) of a conversation.

Current models are trained by predicting the next dialogue turn in a given conversational context using the maximum-likelihood estimation (MLE) objective function. 
Its incapability of handling this conversational long-term success results in many pitfalls:
First, 
the fact that \sts models tend to generate highly generic responses, as discussed in Chapter \ref{mutual} actually results from this incapability: responses such as {\it ``I don't know''} are apparently not a good action to take in long term, since it closes the conversation down; 
Second, as illustrated in the two sample conversations on the  left of Table \ref{con-simulation},  the system becomes stuck in an infinite loop of repetitive responses. This is due to MLE-based \sts models' inability to account for repetition.
In example 2 (bottom left), the dialogue falls into an infinite loop after three turns, with both agents generating dull, generic utterances like {\it i don't know what you are talking about} and {\it you don't know what you are saying}. 
Looking at the entire conversation, utterance (4) turns out to be a bad action to take
because it offers no way of continuing the conversation.\footnote{A similar rule is often suggested in improvisational comedy: \url{https://en.wikipedia.org/wiki/Yes,_and...}}
\begin{table*}
\centering
\small
\begin{tabular}{ll}\toprule
{\bf Baseline mutual information model}  & {\bf Proposed reinforcement learning model} \\\midrule
\textsc{A}: {Where are you going?} (1)             & A: Where are you going? (1)   \\
\textsc{B}: {I'm going to the restroom}. (2)    & B: I'm going to the police station. (2) \\
\textsc{A}: {See you later}. (3)			    & A: I'll come with you. (3) \\
\textsc{B}: {See you later}. (4)			    & B: No, no, no, no, you're not going anywhere. (4) \\
\textsc{A}: {See you later}. (5)		    & A: Why? (5) \\
\textsc{B}: {See you later}. (6)			    & B: I need you to stay here. (6)\\
...	&								    A: I don't know what you are talking about. (7)\\
...& ... \\\midrule
\textsc{A}: {how old are you?} (1) & A: How old are you? (1)\\
\textsc{B}: {I'm 16.} (2)& B: I'm 16. Why are you asking? (2)\\
\textsc{A}: {16?} (3) & A I thought you were 12. (3) \\
\textsc{B}: {I don't know what you are talking about.} (4) & B: What made you think so? (4)\\
\textsc{A}: {You don't know what you are saying.} (5) & A: I don't know what you are talking about. (5)\\
\textsc{B}: {I don't know what you are talking about .} (6) & B: You don't know what you are saying. (6)\\
\textsc{A}: {You don't know what you are saying.} (7) & ... \\
... & ... \\\bottomrule
\end{tabular}					
\caption[Dialogue simulation between two agents using \sts models]{ Left Column: Dialogue simulation between two agents using  a \sts trained on the OpenSubtitles dataset. The first turn (index 1) is input by the authors. Then the two agents take turns conversing, taking as input the other agent's prior generated turn. The output is generated using the mutual information model \cite{li2015diversity} in which an N-best list is first obtained using beam search based on p(t$|$s) and reranked by linearly combining the backward probability p(s$|$t), where t  and s respectively denote targets and  sources.  Right Column: Dialogue simulated using the proposed reinforcement learning model. The new model has more forward-looking utterances (questions like ``Why are you asking?" and offers like ``I'll come with you") and lasts longer before it falls into conversational black holes.}
\label{con-simulation}	
\end{table*}

These challenges suggest we need a conversation framework that has the ability to (1)  integrate developer-defined rewards that better mimic the true goal of chatbot development
  and (2)  model the long-term influence of a generated response in an ongoing dialogue.

To achieve these goals,  we draw on the insights of reinforcement learning, which
have been widely applied in MDP and POMDP dialogue systems (see Related Work section for details).
We introduce a neural reinforcement learning (RL) generation method, 
which can optimize long-term rewards designed by system developers.
Our model uses the encoder-decoder architecture as its backbone, and 
 simulates conversation between two virtual agents to explore the space of possible actions while learning to maximize expected reward.
We define simple heuristic approximations to rewards that characterize
 good conversations: good conversations are forward-looking \cite{All92} or interactive (a turn suggests
 a following turn), informative, and coherent.
  The parameters of an encoder-decoder RNN define a policy over an infinite action space consisting of all possible utterances.
  The agent learns a policy by optimizing the long-term developer-defined reward from ongoing dialogue simulations using policy gradient methods \cite{williams1992simple},
  rather than the MLE objective defined in standard \sts models. 

Our model thus integrates the power of \sts systems to learn compositional semantic
meanings of utterances with the strengths of reinforcement learning in
optimizing for long-term goals across a conversation.
 Experimental results (sampled results at the right panel of Table \ref{con-simulation}) demonstrate that our approach fosters a more sustained dialogue and
  manages to produce more interactive  responses than standard \sts models trained using the MLE objective.
\section{Model}  
  In this section, we describe in detail the components of the proposed RL model. 
The learning system consists of two agents. 
We use $p$ to denote sentences generated from the first agent and  $q$ to denote sentences from the second. The two agents take turns talking with each other. A dialogue can be represented as an alternating sequence of sentences generated by the two agents: $p_1, q_1, p_2, q_2, ..., p_{i}, q_{i}$.  We view the generated sentences as actions that are taken according to a policy defined by an
encoder-decoder recurrent neural network language model.

The parameters of the network are optimized to maximize the expected future reward using policy search.  
Policy gradient methods are more appropriate for our scenario than Q-learning \cite{mnih2013playing}, because we can initialize the encoder-decoder RNN using
MLE parameters that already produce plausible responses, before changing the objective and tuning towards a policy that maximizes long-term reward.
Q-learning, on the other hand, directly estimates the future expected reward of each action, which can differ from the MLE objective by orders of magnitude,
thus making MLE parameters inappropriate for initialization.
The components (states, actions, reward, etc.) of our sequential decision problem are summarized in the following sub-sections.

\subsection{Action} An action $a$ is the dialogue utterance to generate. The action space is infinite since arbitrary-length sequences can be generated.

\subsection{State} A state is denoted by the previous two dialogue turns $[p_i,q_i]$. 
The dialogue history is further transformed to a vector representation by feeding the concatenation of $p_i$ and $q_i$ into an LSTM encoder model as described in \newcite{li2015diversity}.

\subsection{Policy} A policy takes the form of an LSTM encoder-decoder (i.e., $p_{RL}(p_{i+1}|p_{i}, q_i)$ ) and is defined by its parameters.
Note that we use a stochastic representation of the policy (a probability distribution over actions given states).
A deterministic policy would result in a discontinuous objective that is difficult to optimize using gradient-based methods.

\subsection{Reward} $r$ denotes the reward obtained for each action. In this subsection, we discuss major factors that contribute to the success of a dialogue and describe how 
approximations to these factors can be operationalized in computable reward functions. 

\paragraph{Ease of answering}
A turn generated by a machine should be easy to respond to.
This aspect of a turn is related to its {\em forward-looking function}: the constraints
a turn places on the next turn \cite{All92}.
We propose measuring the ease of answering a generated turn by
using the negative log likelihood of responding to that utterance with a dull response.
We manually constructed a list of dull responses $\mathbb{S}$ consisting 8 dialogue utterances such as ``I don't know what you are talking about", ``I have no idea", etc., that we and others have found
occur very frequently in \sts models of conversations.
The reward function is given as follows:
\begin{equation}
r_1=-\frac{1}{N_{\mathbb{S}}}\sum_{s\in \mathbb{S}}\frac{1}{N_s}\log p_{\text{seq2seq}} (s|a)
\label{eq1}
\end{equation}
where $N_{\mathbb{S}}$ denotes the
number of dull responses that we predefined 
 and $N_s$ denotes the number of tokens in the dull response $s$. 
Although of course there are more ways to generate dull responses than the list can cover,
many of these responses are likely to fall into similar regions in the vector space computed by the model.
A system less likely to generate utterances in the list is thus also less likely to generate other dull responses. 

$p_{\text{seq2seq}}$ represents the  likelihood output by \sts models. 
It is worth noting that $p_{\text{seq2seq}}$ is different from the stochastic policy function $p_{RL}(p_{i+1}|p_{i}, q_i)$, since the former is learned based on the MLE objective of the \sts model while the latter is the policy optimized for long-term future reward in the RL setting.

\paragraph{Information Flow}
We want each agent to contribute new information at each turn to keep the dialogue moving and avoid repetitive sequences. 
We therefore propose penalizing semantic similarity between consecutive turns from the same agent. 
Let $h_{p_i}$ and $h_{p_{i+1}}$ denote representations obtained from the encoder for two consecutive turns $p_i$ and $p_{i+1}$. The reward is given by the negative log of the cosine similarity
between them:
\begin{equation}
r_2=-\log \text{cos}(h_{p_i},h_{p_{i+1}}) = -\log \text{cos} \frac{h_{p_i} \cdot h_{p_{i+1}}}{\lVert h_{p_i} \rVert \lVert h_{p_{i+1}} \rVert}
\label{sim}
\end{equation}
\paragraph{Semantic Coherence}
We also need to measure the adequacy of responses to avoid situations in which the generated replies are highly rewarded but are ungrammatical or not coherent.
 We therefore  consider the mutual information between the action $a$ and previous turns in the history to ensure the generated responses are coherent and appropriate: 
\begin{equation}
r_3=\frac{1}{N_a}\log p_{\text{seq2seq}}(a|q_i,p_i)+\frac{1}{N_{q_i}}\log p_{\text{seq2seq}}^{\text{backward}}(q_i|a)
\label{eq4}
\end{equation}
$p_{\text{seq2seq}}(a|p_i,q_i)$  denotes the probability of generating response $a$ given the previous dialogue utterances $[p_i,q_i]$. $p_{\text{seq2seq}}^{\text{backward}}(q_i|a)$ denotes the backward probability of generating the previous dialogue utterance $q_i$ based on response $a$. $p_{\text{seq2seq}}^{\text{backward}}$ is trained in a similar way as standard \sts models with sources and targets swapped. 
Again, to control the influence of target length, both $\log p_{\text{seq2seq}}(a|q_i,p_i)$ and  $\log p_{\text{seq2seq}}^{\text{backward}}(q_i|a)$ are scaled by the length of targets.
The final reward for  action $a$ is a weighted sum of the rewards discussed above:
\begin{equation}
r(a,[p_i,q_i])=\lambda_1 r_1+\lambda_2 r_2+\lambda_3 r_3
\label{reward}
\end{equation}

The central idea behind our approach is to simulate the process of two virtual agents taking turns talking with each other, through which we can explore the state-action space and learn a policy
$p_{RL}(p_{i+1}|p_{i}, q_i)$ that leads to the optimal expected reward.
We adopt an AlphaGo-style strategy \cite{silver2016mastering} by initializing the RL system using a general response generation policy which is learned from a fully supervised setting. 
\section{Simulation}
\subsection{Supervised Learning}
 For the first stage of training, we build on prior work of predicting a generated target sequence given dialogue history using the supervised \sts model \cite{vinyals2015neural}.
 Results from supervised models will be later used for initialization. 
 
We trained a \sts model with attention \cite{bahdanau2014neural} on the OpenSubtitles dataset, which consists of roughly 80 million source-target pairs. 
We treated each turn in the dataset as a target and the concatenation of two previous sentences as source inputs. 
\subsection{Mutual Information}
Samples from \sts models are often times dull and generic, e.g., ``{\it i don't know}" \cite{li2015diversity}.
We thus do not want to initialize the policy model using the pre-trained \sts models because this will lead to a lack of diversity in the RL models' experiences. 
 \newcite{li2015diversity} showed that modeling mutual information between sources and targets will significantly decrease the chance of generating dull responses and improve  general response quality.
We now show how we can obtain an encoder-decoder model which generates maximum mutual information responses.

As illustrated in \newcite{li2015diversity}, direct decoding from Eq~\ref{eq4} is infeasible since the second term 
 requires
 the target sentence to be  completely generated.
 Inspired by recent work on sequence level learning \cite{ranzato2015sequence}, we treat the problem of 
generating maximum mutual information response 
as a reinforcement learning problem in which a reward of mutual information value is observed when the model arrives at the end of  a sequence. 

Similar to \newcite{ranzato2015sequence}, we use policy gradient methods \cite{sutton1999policy,williams1992simple} for optimization. We initialize the policy model $p_{RL}$ using a pre-trained $p_{\sts}(a|p_i,q_i)$ model. 
Given an input source $[p_i,q_i]$, 
we generate a candidate list  $A=\{\hat{a}|\hat{a}\sim p_{RL}\}$. 
For each generated candidate $\hat{a}$, we will obtain the mutual information score $m(\hat{a}, [p_i,q_i])$ from the pre-trained $p_{\sts}(a|p_i,q_i)$ and $p^{\text{backward}}_{\sts}(q_i|a)$.
This mutual information score will be used as a reward and back-propagated to the encoder-decoder model, tailoring it to generate sequences with higher rewards. 
We refer the readers to \newcite{zaremba2015reinforcement}  and
\newcite{williams1992simple} for details.  The expected reward for a sequence is given by:
\begin{equation}
\begin{aligned}
&J(\theta)=\mathbb{E} [m(\hat{a},[p_i,q_i]) ]\\
\end{aligned}
\end{equation}
The gradient is
estimated using the likelihood ratio trick:
\begin{equation}
\begin{aligned}
\nabla J(\theta)=m(\hat{a},[p_i,q_i])\nabla\log p_{RL}(\hat{a}|[p_i,q_i])
\end{aligned}
\end{equation}
We update the parameters in the encoder-decoder model using stochastic gradient descent. 
A
curriculum learning strategy is adopted \cite{bengio2009curriculum} 
as in \newcite{ranzato2015sequence}  
such that, for
every sequence of length $T$ we use the MLE loss for the first $L$ tokens and the reinforcement algorithm for the remaining $T-L$ tokens. We gradually anneal 
the value of $L$ to zero.
A baseline strategy is employed to decrease the learning variance:
an additional neural model takes as inputs the generated target and the initial source and outputs a baseline value, similar to the strategy 
adopted by \newcite{zaremba2015reinforcement}. 
 The final gradient is thus:
\begin{equation}
\begin{aligned}
\nabla J(\theta)=\nabla\log p_{RL}(\hat{a}|[p_i,q_i]) [m(\hat{a},[p_i,q_i])-b]
\end{aligned}
\end{equation}

\subsection{Dialogue Simulation between Two Agents}
\label{sec:learning}
We simulate conversations between the two virtual agents and 
have them take turns talking with each other. The simulation proceeds as follows: at the initial step, a message from the training set is fed to the first agent. The agent encodes the input message to a vector representation and starts decoding to generate  a response output. 
Combining the immediate output from the first agent with the dialogue history, the second agent updates the state by encoding the dialogue history into a representation and uses the decoder RNN to generate responses, which are subsequently fed back to the first agent, and the process is repeated. 

\begin{figure*}[!ht]
    \centering
    \includegraphics[width=6in]{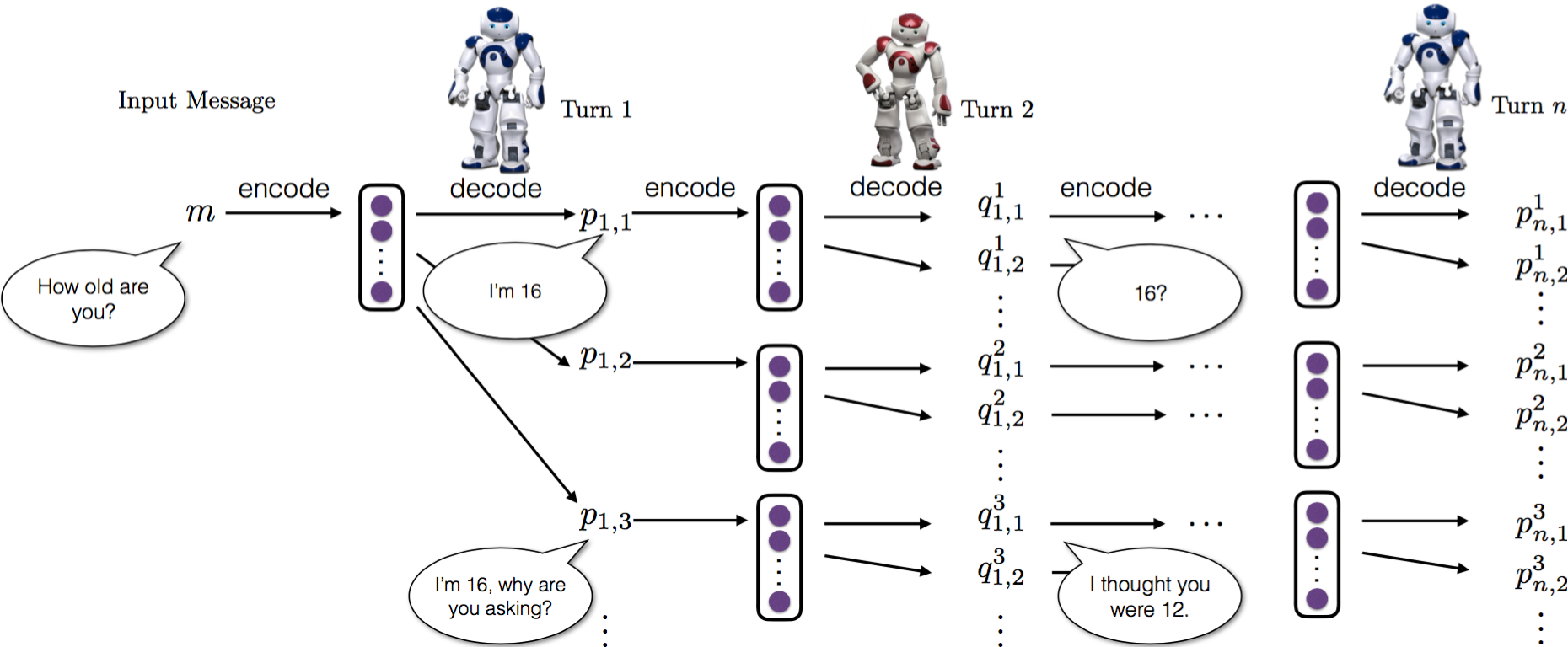}
\caption{Dialogue simulation between the two agents. }
\end{figure*}

\paragraph{Optimization}
We initialize the policy model $p_{RL}$ with parameters from the mutual information model described in the previous subsection. 
We then use policy gradient methods to find parameters that lead to a larger expected reward.
The objective to maximize is the expected future reward:
\begin{equation}
J_{RL}(\theta)=\mathbb{E}_{p_{RL}(a_{1:T})} \left[ \sum_{i=1}^{i=T} R(a_i,[p_i,q_i] ) \right]
\end{equation}
where $R(a_i,[p_i,q_i])$ denotes the reward resulting from action $a_i$. 
We
use the likelihood ratio trick 
\cite{williams1992simple,glynn1990likelihood} for gradient updates:
\begin{equation}
\nabla J_{RL}(\theta) \approx { \sum_{i}\nabla\log p(a_i|p_i,q_i)} \sum_{i=1}^{i=T} R(a_i,[p_i,q_i] )
\end{equation}

\subsection{Curriculum Learning}
A curriculum learning strategy is again employed  in which we begin by simulating the dialogue for 2 turns, and gradually increase the number of simulated turns. 
We generate 5 turns at most, as  the number of candidates to examine grows exponentially in the size of the candidate list.
Five candidate responses are generated at each step of the simulation.

In this section, we  describe experimental results along with qualitative analysis.  We evaluate dialogue generation systems using both human judgments and
two automatic metrics: conversation length (number of turns in the entire session) and diversity.
\section{Experiments}
\subsection{Dataset}
The dialogue simulation requires  high-quality initial inputs fed to the agent. For example, an initial input of ``why ?" is undesirable since it is unclear how the dialogue could proceed. 
We take a subset of 10 million messages from the OpenSubtitles dataset and extract 0.8 million sequences with the lowest likelihood of generating the response
``{\it i don't know what you are taking about}" to ensure initial inputs are easy to respond to.

\subsection{Automatic Evaluation}
Evaluating dialogue systems is difficult. Metrics such as BLEU
\cite{papineni2002bleu} and perplexity have been widely used for
dialogue quality evaluation
\cite{li2015diversity,vinyals2015neural,sordoni2015neural}, but it
is widely debated how well these automatic metrics are correlated
with true response quality \cite{galley2015deltableu}.
Since the goal of the proposed system is not to predict the highest probability response, but rather the long-term success of the dialogue, 
we do not employ BLEU or perplexity for evaluation.\footnote{We found the RL model performs worse on BLEU score. On a random sample of 2,500 conversational pairs, single reference BLEU scores for RL models, mutual information models and vanilla \sts models  are respectively 1.28, 1.44, and 1.17.  BLEU is highly correlated with perplexity in generation tasks. Since the RL model is trained based on future reward rather than MLE, it is not surprising that the RL based models achieve lower BLEU score.}

\paragraph{Length of the Dialogue}
The first metric we propose is the length of the simulated dialogue. We say a dialogue ends when one of the agents starts generating dull responses such as ``{\it i don't know}" \footnote{We use a simple rule matching method, with a list of 8 phrases that count as dull responses.  Although this can lead to both false-positives and -negatives, it works pretty well in practice. }
or two consecutive utterances from the same user are highly overlapping.\footnote{Two utterances are considered to be repetitive if they share more than 80 percent of their words.}


The test set consists of 1,000 input messages. 
To reduce the risk of circular dialogues, we limit the number of simulated turns to be less than 8.  
Results are shown in Table \ref{numturns}. As can be seen, using mutual information leads to more sustained conversations between the two agents.
The proposed  RL model is first trained based on the mutual information objective and thus benefits from it in addition to the RL model. 
We observe that the RL model with dialogue simulation achieves the best evaluation score.

\begin{table}
\centering
\begin{tabular}{cc}\toprule
Model& $\#$ of simulated turns \\\midrule
\sts& 2.68 \\
mutual information& 3.40\\
RL&4.48 \\\bottomrule
\end{tabular}
\caption{The average number of simulated turns from standard \sts models, mutual information model and the proposed RL model.}
\label{numturns}
\end{table}

\begin{table*}[!ht]
\centering
\small
\resizebox{\textwidth}{!}{
\begin{tabular}{lll}\toprule
Input Message& Mutual Information Model& RL \\
How old are you? & I'm 16. & I'm 16, why are you asking? \\
What's your full name? & I have no idea. & What's yours? \\
I don't want to go home tonight. & Really? & Why? \\
Do you have any feelings for me? & I don't know what you are talking about. &I don't want to hurt your feelings. \\
How much time do you have here? & Not long enough. Sorry, sir. & Ten seconds. \\
Shall we get started?& Of course! & Yes. We've got a lot of work to do here.\\
Do you play football? & No, i don't & Yes. I love football!\\ 
We'd have to talk to him.&I mean, he's a good guy&About what ? \\
How come you never say it? & Because I don't want to hurt you.& I don't think it's a good idea to say it. \\\bottomrule
\end{tabular}
}
\caption{Sampled responses generated from the mutual information models and the proposed RL model.}
\label{sen}
\end{table*}

\paragraph{Diversity} 
We report degree of diversity by calculating
the number of distinct unigrams and bigrams
in generated responses. The value is scaled by the total
number of generated tokens to avoid favoring long
sentences as described in \newcite{li2015diversity}.
The resulting metric is thus a type-token ratio for unigrams and bigrams.

For both the standard \sts model and the proposed RL model, we use beam search with a beam size 10 to generate a response to a given input message. For the mutual information model, we first generate $n$-best lists using $p_{\sts}(y|x)$ and then linearly re-rank them using $p_{\sts}(x|y)$.
Results are presented in Table \ref{diversity}. We find that the proposed RL model generates more diverse outputs when compared against both the vanilla \sts model and the mutual information model. 

\begin{table}[ht]
\centering
\begin{tabular}{cll}\toprule
Model& Unigram&Bigram \\\midrule
\sts& 0.0062&0.015 \\
mutual information& 0.011&0.031\\
RL&0.017&0.041 \\\bottomrule
\end{tabular}
\caption{Diversity scores (type-token ratios) for the standard \sts model, mutual information model and the proposed RL model.}
\label{diversity}
\end{table}

\paragraph{Human Evaluation}
We explore three settings for human evaluation:
the first setting is similar to what was described in \newcite{li2015diversity},
where we employ crowdsourced judges to evaluate a random sample of 500 items. 
We present both an input message and the generated outputs to 3 judges and ask them to 
 decide which of the two outputs is better (denoted as {\it single-turn general quality}). 
Ties are permitted. 
Identical strings are assigned the same score. 
We measure the improvement achieved by the RL model over the mutual information model by the mean difference in scores between the models. 

For the second setting, judges are again presented with input messages and system outputs, and are asked to decide which of the two outputs is easier to respond to so that the conversation is able to continue
(denoted as {\it single-turn ease to answer}).
Again we evaluate a random sample of 500 items, each being assigned to 3 judges. 
\begin{figure}
\begin{tabular}{p{14cm}}\hline
{\bf Title}: Describe which response is easier to respond to (about 1 min)\\
~~\\
{\bf Instructions}:\\
In each task, you are given two dialogue episodes, each of which consists two dialogue turns. The first turns of the two episodes are the same. 
You need to decide the second turn of which dialogue episode is easier to respond to. Ties are permitted. 
~~\\~~\\
Each task will be assigned to three Turkers. 
We will give bonuses for those whose results for the same task are consistent.\\\hline
\end{tabular}
\caption{Instructions given to turkers for the single-turn-ease-to-answer task.}
\label{EasyTurker}
\end{figure}

For the third setting, judges are presented with simulated conversations between the two agents (denoted as {\it multi-turn general quality}). Each conversation consists of 5 turns. 
We evaluate 200 simulated conversations, each being assigned to 3 judges, who are asked to decide which of the simulated conversations is of higher quality.

\begin{table}[!ht]
\centering
\small
\begin{tabular}{cccc} \\\hline
Setting&RL-win&RL-lose&Tie  \\\hline
single-turn general quality&0.40&0.36&0.24 \\
single-turn ease to answer& 0.52&0.23&0.25\\
multi-turn general quality&0.72&0.12&0.16\\\hline
\end{tabular}
\caption{RL gains over the mutual information system based on pairwise human judgments.}
\label{human-eval}
\end{table}

Results for human evaluation are shown in Table \ref{human-eval}. 
The proposed RL system does not introduce a significant boost in single-turn response quality (winning 40 percent of time and losing 36 percent of time).
This is in line with our expectations, 
as the RL model is not optimized to predict the next utterance, but rather to increase long-term reward.
The RL system produces responses that are significantly easier to answer than 
does the mutual information system, as demonstrated by the {\it single-turn ease to answer} setting (winning 52 percent of time and losing 23 percent of time),
and also significantly higher quality multi-turn dialogues, as demonstrated by the {\it multi-turn general quality} setting (winning 72 percent of time).

\paragraph{Qualitative Analysis and Discussion}
We show a random sample of generated responses in Table \ref{sen} and simulated conversations in Table \ref{con-simulation} at the beginning of the paper.
From Table \ref{sen}, we can see that the RL based agent indeed generates more interactive responses than the other baselines. We also find that the RL model has a tendency to 
end a sentence with another question and hand the conversation over to the user. 
From Table \ref{con-simulation}, we observe that the RL model manages to produce more interactive and sustained conversations than the mutual information model. 

During error analysis, we found that although we penalize repetitive utterances in consecutive turns, the dialogue sometimes enters a cycle with length greater than one, as shown in Table \ref{he}. This can be ascribed to the 
limited amount of conversational history we consider. 
Another issue observed is that the model sometimes starts a less relevant topic during the conversation. There is a tradeoff between relevance and less repetitiveness, as manifested in the reward function we defined. 

\begin{table}[!ht]
\centering
\small
\begin{tabular}{l}\toprule
A: What's your name ?\\
B: Daniel.  \\
A: How old are you ?\\
B. Twelve. What's your name ? \\
A. Daniel.  \\
B: How old are you ? \\
A: Twelve. What's your name ? \\
B: Daniel.\\
A: How old are you ? \\
B ... \\\bottomrule
\end{tabular}
\caption{An simulated dialogue with a cycle longer than one.}
\label{he}
\end{table}

The fundamental problem, of course, is that the manually defined reward function 
can't possibly cover the crucial aspects that define an ideal conversation. 
While the heuristic rewards that we defined are amenable to automatic calculation,
and do capture some aspects of what makes a good conversation, ideally 
the system would instead receive real rewards from humans. But such a strategy is both costly and hard to scale up.
As we will describe in the following section, we can train a machine to act as the role of a human evaluator and provide reward signals to train the response generation model. 

\section{Conclusion}
We introduce a reinforcement learning framework for neural response
generation by simulating dialogues between two agents, integrating the strengths
of neural \sts systems and reinforcement learning for dialogue.
Like earlier neural \sts models, our framework
captures the compositional models of the meaning of a dialogue turn and 
generates semantically appropriate responses. Like reinforcement learning 
dialogue systems, our framework is able to generate utterances that 
optimize future reward, successfully capturing global properties of a good conversation.
Despite the fact that our model uses very simple, operationable heuristics for capturing these global properties, the framework generates more diverse, interactive responses that foster a more sustained conversation.

\chapter{Adversarial Learning for Dialogue Generation}
\label{adversarial}
In the previous chapter (chapter \ref{RL}), we manually define three types of ideal dialogue properties, i.e., ease of answering, informativeness, and coherence, 
based on which a reinforcement learning system is trained. 
However, 
it is widely
acknowledged that  
manually defined reward functions can't possibly
cover all crucial aspects and can lead to suboptimal generated utterances. 
This relates to two important questions in dialogue learning:  
what are the crucial
aspects that define an ideal conversation and how
can we quantitatively measure them.

A good dialogue model should generate utterances indistinguishable from human dialogues.
Such a goal suggests a training objective 
resembling the idea of the Turing test \cite{turing1950computing}.
We borrow the idea of adversarial training \cite{goodfellow2014generative,denton2015deep} in 
computer
vision, in which we jointly train two models, 
a generator (which takes the form of the neural  \sts model) that defines the probability of generating a dialogue sequence, and 
a discriminator
that labels dialogues as human-generated or machine-generated. 
This discriminator  is analogous to  the evaluator in the Turing test.
We cast the task as a reinforcement learning problem, in which the quality of machine-generated utterances is measured by its ability to fool the discriminator into believing that it is a human-generated one. The output from the discriminator is used as a reward to the generator, pushing it to generate 
utterances indistinguishable from human-generated dialogues. 

The idea of a Turing test---employing an evaluator 
to distinguish  machine-generated texts from human-generated
ones---can be applied not only to training but also testing,
where it goes by the name of adversarial evaluation. Adversarial
evaluation was first employed in \newcite{bowman2015generating} to evaluate sentence generation quality, and preliminarily studied in the context of dialogue generation by \newcite{kannan}. 
Here, we discuss potential pitfalls of adversarial evaluations and 
  necessary steps to avoid them and make evaluation reliable.

Experimental results demonstrate that our approach
produces more interactive, interesting, and non-repetitive responses than standard
\sts models trained using the MLE objective function.
\section{Adversarial Reinforcement Learning}

In this section, we describe in detail the components of the proposed adversarial reinforcement learning model. 
The problem can be framed as follows: given a dialogue history $x$ consisting of a sequence of dialogue utterances,\footnote{We approximate the dialogue history using the concatenation of two preceding utterances. We found that using more than 2 context utterances yields very tiny performance improvements for \sts models.} the model needs to generate a response $y=\{y_1,y_2,...,y_{L_y}\}$.
 We view the 
 process of sentence generation 
   as a sequence of actions that are taken according to a policy defined by an
encoder-decoder recurrent neural networks.

\subsection{Adversarial REINFORCE}
The adversarial REINFORCE algorithm consists of two components:
a generative model $G$ and a discriminative model $D$.

\paragraph{Generative Model} The generative model $G$ defines the
 policy that generates a response $y$ given dialogue history $x$. 
It takes a  form similar to \sts models, which first map the source input to a vector representation using a recurrent net
and then compute the probability of generating each token in the target using a softmax function.

\paragraph{Discriminative Model} The discriminative model $D$ is a binary classifier that takes as input a sequence of dialogue utterances $\{x,y\}$
 and outputs a label indicating whether the input is generated by humans or machines.  
The input dialogue is encoded into a vector representation 
using a hierarchical encoder  \cite{li2015hierarchical,serban2016building},\footnote{To be specific, each utterance $p$ or $q$ is mapped to a vector representation $h_p$ or $h_q$ using LSTM \cite{hochreiter1997long}.
Another LSTM is put on sentence level, mapping the entire dialogue sequence to a single representation} which is then fed to a 2-class softmax function, returning the probability of the input dialogue episode being a machine-generated dialogue
(denoted $Q_-(\{x,y\})$)
 or a human-generated dialogue  (denoted $Q_+(\{x,y\})$). 
\paragraph{Policy Gradient Training}
 The key idea of the system is to encourage the generator to generate utterances that are indistinguishable from human generated dialogues. We use policy gradient methods to achieve such a goal, in which the 
score of current utterances being human-generated ones assigned by the discriminator
(i.e.,  $Q_+(\{x,y\})$)
 is used as a reward for the generator, which is trained to maximize the expected reward of generated utterance(s) using the REINFORCE algorithm \cite{williams1992simple}:
 \begin{equation}
 J(\theta)=\mathbb{E}_{y\sim p(y|x)}(Q_+(\{x,y\})|\theta)
 \label{lb1}
 \end{equation}
Given the input dialogue history $x$, the  bot generates a dialogue utterance $y$ by sampling from the policy. 
The concatenation of the generated utterance $y$ and the input $x$ is fed to the discriminator. 
 The gradient of \eqref{lb1} is approximated using the likelihood ratio trick \cite{williams1992simple,glynn1990likelihood,aleksand1968stochastic}:
 \begin{multline}
\nabla J(\theta)\approx [Q_+(\{x,y\})-b(\{x,y\})] \\ \nabla\log \pi(y|x)
\\ = [Q_+(\{x,y\})-b(\{x,y\})] \\ \nabla\sum_t \log p(y_t|x,y_{1:t-1})
\label{partial}
\end{multline}
where $\pi$ denotes the probability of the generated responses. 
 $b(\{x,y\})$ denotes the baseline value to reduce the variance of the estimate while keeping it unbiased.\footnote{
 Like \newcite{ranzato2015sequence}, 
 we train another neural network model (the critic) to estimate the value (or future reward) of current state (i.e., the dialogue history) under the current policy $\pi$. The critic network takes as input the dialogue history, transforms it to a vector representation using a hierarchical network and maps the representation to a scalar. The network is optimized based on the mean squared loss between the estimated reward and the real reward.} The discriminator is simultaneously updated with the human generated dialogue that contains dialogue history $x$ as a positive example and the machine-generated dialogue as a negative example. 
\subsection{Reward for Every Generation Step (REGS)}
The  REINFORCE algorithm
described has the disadvantage that the expectation of the reward is approximated by only one sample, and the reward associated with this sample 
(i.e.,  $[Q_+(\{x,y\})-b(\{x,y\})]$ in Eq\eqref{partial})
is used for all actions (the generation of each token) in the generated sequence. 
Suppose, for example, the input history is {\it what's your name}, the human-generated response is {\it I am John}, and the machine-generated response is {\it I don't know}. 
The vanilla REINFORCE model assigns the same negative reward to all tokens within the human-generated response (i.e., {\it I}, {\it don't}, {\it know}), whereas 
proper credit assignment in training would  give separate rewards, most likely a neutral reward for the token {\it I}, and negative rewards to {\it don't} and {\it know}. 
We call this {\it reward for every generation step}, abbreviated {\it REGS}.

Rewards for intermediate steps or partially decoded sequences are thus necessary. Unfortunately, the discriminator is trained to assign scores to fully generated sequences, but not partially decoded ones. 
We propose two strategies for computing intermediate step rewards by (1) using Monte Carlo (MC) search and (2) training a discriminator that is able to assign rewards to partially decoded sequences.

In (1) Monte Carlo search, given a partially decoded $s_P$, the model 
keeps sampling tokens from the distribution
 until the decoding finishes. Such a process is repeated $N$ (set to 5) times and the $N$ generated   sequences will share  a common prefix $s_P$.
These $N$ sequences 
are fed to the discriminator, the average score of which is used as a reward for the $s_P$. A similar strategy is adopted in \newcite{yu2016seqgan}. 
The downside of MC is that 
it requires repeating the sampling process for each prefix of each sequence  and is thus significantly time-consuming.\footnote{Consider one target sequence with length 20, we need to sample 5*20=100 full sequences to get rewards for all intermediate steps. Training one batch with 128 examples roughly takes roughly 1 min on a single GPU, which is computationally intractable considering the size of the dialogue data we have. We thus parallelize the sampling processes, distributing jobs across 8 GPUs. }

In (2), we directly train a discriminator that is able to assign rewards to both fully and partially decoded sequences. 
We break 
the generated sequences into partial sequences, namely 
$\{y^+_{1:t}\}_{t=1}^{N_{Y^+}}$ and $\{y^-_{1:t}\}_{t=1}^{N_{Y^-}}$
 and use 
 all instances in  $\{y^+_{1:t}\}_{t=1}^{N_{Y^+}}$ as positive examples and  instances  $\{y^-_{1:t}\}_{t=1}^{N_{Y^-}}$ as negative examples. 
The problem with such a strategy is that earlier actions in a sequence are 
shared among multiple training examples for
 the discriminator 
(for example, token $y^+_1$ is contained in all partially generated sequences, which results in overfitting. 
To mitigate this problem, 
we adopt a  strategy similar to when training value networks in  {\it AlphaGo} \cite{silver2016mastering}, in which 
for each collection of subsequences of $Y$, we randomly sample only one example from $\{y^+_{1:t}\}_{t=1}^{N_{Y^+}}$ and one example from $\{y^-_{1:t}\}_{t=1}^{N_{Y^-}}$, which are treated as positive and negative examples
to update the discriminator. 
Compared with the Monte Carlo search model, this strategy is significantly more time-effective, but comes with the weakness that the discriminator becomes less accurate  after partially decoded sequences are added in as training examples. 
We find that the MC model performs better when training time is less of an issue.

For each partially-generated sequence $Y_t=y_{1:t}$, the discriminator gives a classification score $Q_+(x,Y_t)$. 
We compute the baseline $b(x,Y_t)$
using a similar model  to the vanilla REINFORCE model. 
This yields the following gradient to update the generator:
 \begin{multline}
\nabla J(\theta)\approx \sum_t  (Q_+(x,Y_t)-b(x,Y_t))  \\
\nabla\log p(y_t|x,Y_{1:t-1})
\label{action}
\end{multline}
Comparing \eqref{action} with \eqref{partial}, we can see that
the values for  
rewards and baselines are different among generated tokens in the same response. 

\paragraph{Teacher Forcing}
Practically, we find that  updating the generative model only using Eq.~\ref{lb1} leads to  unstable training for both vanilla Reinforce and REGS, with the perplexity value skyrocketing after training the model for a few hours (even when the generator is initialized using a pre-trained \sts model). The reason this happens is that the generative model can only be indirectly exposed to the gold-standard target sequences through the reward 
passed back from the
 discriminator, and this reward is used to promote or discourage its (the generator's) own generated sequences. Such a training strategy is fragile: once the generator (accidentally) deteriorates in some training batches and the discriminator consequently does an extremely good job in recognizing sequences from the generator,  
the generator immediately gets lost. It knows that its generated sequences are bad based on the rewards outputted from the discriminator, but it does not know what sequences are good and how to push itself to generate these good sequences (the odds of generating a good response from random sampling are minute, due to the vast size of the space of possible sequences). Loss of the reward signal leads to a breakdown in the training process.

To alleviate this issue and give the generator more direct access to the gold-standard targets, 
we propose also feeding human generated responses to the generator for model updates.  
The most straightforward strategy is for the discriminator to automatically assign a reward of 1 (or other positive values) to the human generated responses and for
the generator to use this reward to update itself on human generated examples. 
This can be seen as having a teacher intervene with the generator some fraction of the time and force it to generate the true responses, 
an approach that is similar to the professor-forcing algorithm of \newcite{lamb2016professor}.

A closer look reveals that 
this modification is the same as the standard training of \sts models, making the final training
alternately update the \sts model using the adversarial objective and the MLE objective. One can think of the 
professor-forcing 
model as a regularizer to 
regulate the generator once it starts deviating from the training dataset.

We also propose another workaround, in which the discriminator first assigns a reward to a human generated example using its own model,  and the generator then updates itself using this reward on the human generated example only if the reward is larger than the baseline value. Such a strategy has the advantage that different weights for model updates are assigned to different human generated examples (in the form of different reward values produced by the generator) and that human generated examples are always associated with non-negative weights. 

A review of the proposed model is shown in Figure \ref{fig:adver-reinforce}. 
\begin{figure}
\small
\line(1,0){220} \\
{\bf For} number of training iterations {\bf do} \\
.~\hspace{0.3cm}{\bf For} i=1,D-steps {\bf do} \\
.~\hspace{0.8cm}Sample (X,Y) from real data \\
.~\hspace{0.8cm}Sample $\hat{Y}\sim G(\cdot|X)$\\
.~\hspace{0.8cm} Update $D$ using $(X,Y)$ as positive examples and $(X,\hat{Y})$ as negative examples. \\
.~\hspace{0.3cm}{\bf End} \\
.\\
.~\hspace{0.2cm} {\bf For} i=1,G-steps {\bf do} \\
.~\hspace{0.8cm}Sample (X,Y) from real data \\
.~\hspace{0.8cm}Sample $\hat{Y}\sim G(\cdot|X)$ \\
.~\hspace{0.8cm}Compute Reward $r$ for $(X,\hat{Y})$ using $D$.\\
.~\hspace{0.8cm}Update $G$  on $(X,\hat{Y})$ using reward $r$\\
.~\hspace{0.8cm}Teacher-Forcing: Update $G$  on $(X,Y)$\\
.~\hspace{0.3cm}{\bf End} \\
{\bf End} \\
\line(1,0){220}
\caption{A brief review of the proposed adversarial reinforcement algorithm
for training the generator $G$ and discriminator $D$.
The reward $r$ from the discriminator $D$ can be computed using different strategies according to whether using REINFORCE or REGS.  
The update of the generator $G$ on $(X,\hat{Y})$ can be done by either using Eq.\ref{partial} or Eq.\ref{action}.
D-steps is set to 5 and G-steps is set to 1.}
\label{fig:adver-reinforce}
\end{figure}

\subsection{Training Details}
We first  pre-train the generative model by predicting target sequences given the dialogue history. We trained a  \sts model
  \cite{sutskever2014sequence}   
 with an attention mechanism \cite{bahdanau2014neural,luong2015effective}  on the OpenSubtitles dataset. 
 We followed protocols recommended by \newcite{sutskever2014sequence}, such as gradient clipping, mini-batch, and learning rate decay.
We also pre-train the discriminator. To generate negative examples, 
we  decode   part of the training data. 
Half of the negative examples are generated using beam-search with mutual information reranking 
as described in \newcite{li2015diversity},
and the other half is generated from sampling.

For 
data processing, 
model training, and decoding  (both the proposed adversarial training model and the standard \sts models), we employ a few  strategies that  improve response quality, including: 
(1) Remove training examples with the length of responses shorter than a threshold (set to 5).
We find that this significantly improves the general response quality.\footnote{To compensate for the loss of short responses, one can train a separate model using short sequences.}
(2) Instead of using the same learning rate for all examples, using a weighted learning rate that considers the average tf-idf score for tokens within the response.
Such a strategy
 decreases the influence from dull and generic utterances.\footnote{We treat each sentence as a document. Stop words are removed.
 Learning rates are normalized within one batch. 
 For example, suppose $t_1$, $t_2$, ..., $t_i$, ... ,$t_N$ denote the tf-idf scores for  sentences within current batch and $lr$ denotes the original learning rate. The learning rate for sentence with index $i$ is $N\cdot lr\cdot \frac{t_i}{\sum_{i'}t_{i'}}$. 
 To avoid exploding learning rates for sequences with extremely rare words, the tf-idf score of a sentence 
 is capped at $L$ times the minimum tf-idf score in the current batch. $L$ is empirically chosen and is set to 3.
  } (3) Penalizing intra-sibling ranking when doing beam search decoding to promote N-best list diversity as described in Chapter 3.  (4) Penalizing word types (stop words excluded) that have already been generated. Such a strategy dramatically decreases the rate of repetitive responses such as {\it no. no. no. no. no.} or contradictory responses such as {\it I don't like oranges but i like oranges}. 

\section{Adversarial Evaluation}
In this section, we discuss details of strategies for successful adversarial evaluation. 
It is worth noting that 
 the proposed adversarial training and adversarial evaluation are separate procedures.They are independent of each other and share no common parameters.

 The idea of adversarial evaluation, first proposed by 
\newcite{bowman2015generating}, is to 
train a discriminant
function
 to separate  generated and
true sentences, in an attempt to evaluate the model's sentence generation capability.
The idea has been preliminarily studied by \newcite{kannan} in the context of dialogue generation. 
Adversarial evaluation  also resembles the idea of the Turing test,
which requires a human evaluator 
to distinguish  machine-generated texts from human-generated ones. 
Since it is time-consuming and costly to 
 ask a human to talk to a model 
and give judgements,
we  train a machine evaluator
in place of the  human evaluator to distinguish the human dialogues and machine dialogues, and we use it to measure 
 the general quality of the generated responses. 

Adversarial evaluation involves both  training and testing. 
At training time, the evaluator is trained to
label dialogues as machine-generated (negative) or  human-generated (positive). 
At test time, the trained evaluator is evaluated on a held-out dataset.
If the human-generated dialogues and machine-generated ones are indistinguishable, the model  will achieve 50 percent accuracy at test time.  
\subsection{Adversarial Success}
We define Adversarial Success ({\it AdverSuc} for short) to be the fraction of instances in which a model is capable of fooling the evaluator. {\it AdverSuc} is the difference between 1 and the accuracy 
achieved by the  evaluator. Higher values of {\it AdverSuc} for a dialogue generation model are better. 
\subsection{Testing the Evaluator's Ability}
One caveat  with the adversarial evaluation methods is that 
they are model-dependent. 
We approximate the human evaluator in the Turing test with an automatic evaluator and assume that the evaluator is perfect: low accuracy of the discriminator should indicate high quality of the responses, since we interpret this to mean the generated responses are indistinguishable from the human ones. Unfortunately, there is another factor that can lead to low discriminative accuracy: a poor discriminative model. 
 Consider a  discriminator that always gives random labels or always gives the same label.
 Such an evaluator 
   always yields a high {\it AdverSuc} value of 0.5. 
   \newcite{bowman2015generating} propose  two different discriminator models 
separately using {\it unigram} features and {\it neural} features. It is hard to tell which feature set is more reliable. 
The standard strategy of testing the model on a held-out development set is not suited to this case, since a model that overfits the development set is necessarily superior. 

To deal with this issue, we propose setting up a few manually-invented situations to test the ability of the automatic evaluator.
 This is akin to setting up  examinations to test the ability of the human evaluator in the Turing test. 
We  report not only the {\it AdverSuc} values,  but also the scores that the evaluator
achieves
 in these manually-designed test cases, indicating how much we can trust the reported {\it AdverSuc}. 
 We develop scenarios in which we know  in advance how a perfect evaluator should behave, and then compare   {\it AdverSuc}  from a discriminative model with the gold-standard  {\it AdverSuc}. Scenarios we design include:
 \begin{tightitemize}
 \item We use human-generated dialogues as both positive examples and negative examples. A perfect evaluator should give an {\it AdverSuc} of $0.5$ (accuracy $50\%$), which is the gold-standard result.
  \item We use machine-generated dialogues as both positive examples and negative examples. A perfect evaluator should give an {\it AdverSuc} of $0.5$ (accuracy $50\%$).
 \item We use original human-generated dialogues as positive examples and dialogues 
 consisting of
  random utterances 
  as negative examples. A perfect evaluator should give an {\it AdverSuc} of $0$ (accuracy $100\%$).
 \item We use original human-generated dialogues as positive examples and dialogues with 
 responses replaced by the utterances immediately following the true responses. 
 A perfect evaluation should give an {\it AdverSuc} of $0$ (accuracy $100\%$).
\end{tightitemize}
The {\it evaluator reliability error} (ERE) is the average deviation of an evaluator's adversarial error
 from the gold-standard error in the above tasks, with equal weight for each task. 
The smaller the error, the more reliable the evaluator is. 

\begin{table*}[t]
\centering
\footnotesize
\begin{tabular}{p{4cm}p{11cm}}\hline
{\bf Input}&tell me ... how long have you had this falling sickness ?\\
{\bf ~~Vanilla-MLE}&i 'm not a doctor .\\
{\bf ~~Vanilla-Sample}&well everything you did was totally untrue .\\
{\bf ~~REINFORCE}&i don 't know how long it 's been .\\
{\bf ~~REGS Monte Carlo}& A few months, I guess .\\\hline
{\bf Input}&so i had the doctors test sammy 's response to conditioning .\\
{\bf ~~Vanilla-MLE}&sammy wrote the test sammy wrote the test .\\
{\bf ~~Vanilla-Sample}&objects pick objects up objects objects objects objects objects \\
{\bf ~~REINFORCE}&i 'm not sure that 's relevant .\\
{\bf ~~REGS Monte Carlo}&so he took the pills .\\\hline
{\bf Input}&what are you going to do ?\\
{\bf ~~Vanilla-MLE}&i 'm going to the movies .\\
{\bf ~~Vanilla-Sample}&get him outta here first !\\
{\bf ~~REINFORCE}&i 'm going to get you .\\
{\bf ~~REGS Monte Carlo}&i 'm going to get you a new car .\\\hline
{\bf Input}&they fear your power your intellect .\\
{\bf ~~Vanilla-MLE}&you 're the only one who knows what 's going on .\\
{\bf ~~Vanilla-Sample}&when they are conquered and you surrender they will control all of us .\\
{\bf ~~REINFORCE}&i 'm afraid i 'm not ready yet .\\
{\bf ~~REGS Monte Carlo}&i 'm not afraid of your power .\\\hline
\end{tabular}
\caption{Sampled responses from different models. More in Appendix Tables \ref{example-appendix1} and \ref{example-appendix2}.}
\label{example-appendix1}
\end{table*}

\subsection{Machine-vs-Random Accuracy}
Evaluator reliability error uses scenarios constructed from human-generated
dialogues to assess feature or hyper-parameter choice for the evaluator. Unfortunately, no machine-generated responses are involved in the ERE metric. 
The following example illustrates the serious weakness resulting from this strategy:
as will be shown in the experiment section, 
when inputs are decoded using greedy or beam search models,  
most generation systems to date yield an adversarial success less than 10 percent (evaluator accuracy 90 percent). 
But when using sampling for decoding, the adversarial success skyrockets to around 40 percent,\footnote{Similar results are also reported in \newcite{kannan}.} only 10 percent less than what's needed to pass the Turing test. 
A close look at the decoded sequences using sampling tells a different story: the responses from sampling are sometimes incoherent, irrelevant or even ungrammatical. 

We thus propose an additional sanity check, in which we report the accuracy of distinguishing between machine-generated responses and randomly sampled responses ({\it machine-vs-random} for short). 
This resembles the N-choose-1 metric described in \newcite{shao15}. 
Higher accuracy indicates that the generated responses are distinguishable from randomly sampled human responses, indicating that the generative model is not fooling the generator simply by introducing randomness. 
As we will show in Sec.~\ref{sec:experiments}, using sampling results in high {\it AdverSuc} values but low {\it machine-vs-random} accuracy. 

\section{Experimental Results} \label{sec:experiments}
In this section, 
we detail experimental results on adversarial success and human evaluation.  

\begin{table}[htb]
\centering
\small
\begin{tabular}{ccc}\toprule
Setting&ERE\\
SVM+Unigram&0.232\\
Concat Neural &0.209 \\
Hierarchical Neural &0.193 \\
SVM+Neural+multil-features&0.152 \\\bottomrule
\hline
\end{tabular}
\caption{ERE scores obtained by different models.}
\label{ERE}
\end{table}

\subsection{Adversarial Evaluation}
\paragraph{ERE} We first test 
 adversarial evaluation models with different 
 feature sets and 
 model architectures for reliability, as measured by evaluator reliability error (ERE).
 We explore the following models:
   (1) {\it SVM+Unigram}:  SVM using unigram features.\footnote{Trained using the SVM-Light package \cite{joachims2002learning}.}
 A multi-utterance dialogue (i.e., input messages and responses) is transformed to a unigram representation; (2) 
{\it Concat Neural}: 
a neural classification model with 
a softmax function that takes as input the concatenation of representations of constituent dialogues sentences;
 (3) {\it Hierarchical Neural}: 
  a hierarchical encoder    
  with a  structure similar to the discriminator used in the reinforcement; and
  (4) 
  {\it SVM+Neural+multi-lex-features}: 
  a SVM model that uses the following features: unigrams,  neural representations of dialogues obtained by the neural model trained using strategy (3),\footnote{The representation before the softmax layer.} the forward likelihood $\log p(y|x)$ and backward likelihood $p(x|y)$.

ERE scores obtained by different models are reported in Table \ref{ERE}. 
As can be seen, the {\it hierarchical neural} evaluator (model 3) is more reliable than simply concatenating the sentence-level representations (model 2).
Using the combination of neural features and lexicalized features yields the most reliable evaluator. 
For the rest of this section, we report results obtained 
by the 
{\it hierarchical Neural} setting due to its end-to-end nature, despite its inferiority to {\it SVM+Neural+multil-features}. 

Table \ref{adv} presents
 {\it AdverSuc} values for different models, along with {\it machine-vs-random} accuracy described in Section 4.3. 
Higher values of  {\it AdverSuc}  and  {\it machine-vs-random} are better. 

Baselines we consider include standard \sts models using greedy decoding ({\it MLE-greedy}), beam-search ({\it MLE+BS}) and sampling, as well as the 
mutual information reranking model (as described in Chapter 3) with two algorithmic variations: (1) MMI+$p(y|x)$, in which a large N-best list is first generated using a pre-trained \sts model and then reranked by the backward probability $p(x|y)$ and (2) MMI$-p(y)$, in which language model probability is penalized during decoding. 

Results are shown in Table \ref{adv}. What first stands out  is decoding using sampling (as discussed in Section 4.3), achieving a significantly higher {\it AdverSuc} number than all the rest models. 
However, this does not indicate the superiority of the sampling decoding model, since the {\it machine-vs-random} accuracy is at the same time significantly lower. This means that sampled responses based on \sts models are not only hard for an evaluator to distinguish from real human responses, but also from randomly sampled responses.
A similar, though much less extreme, effect is observed for MMI$-p(y)$, which has an {\it AdverSuc} value slightly higher than {\it Adver-Reinforce}, but a significantly lower {\it machine-vs-random} score. 

By comparing different baselines, we find that MMI+$p(y|x)$ is better than {\it MLE-greedy}, which is in turn better than {\it MLE+BS}. This result is in line with human-evaluation results from \newcite{li2015diversity}. 
The two proposed adversarial algorithms achieve better performance than the baselines. We expect this to be the case, since the adversarial algorithms are trained on an objective function 
more similar to the
the evaluation metric (i.e., adversarial success). 
{\it REGS} performs slightly better than the vanilla REINFORCE algorithm. 

\begin{table}
\small
\centering
\begin{tabular}{ccc}\toprule
Model&{\it AdverSuc}&{\it machine-vs-random} \\\midrule
MLE-BS&0.037&0.942 \\
MLE-Greedy&0.049&0.945 \\ 
MMI+$p(t|s)$&0.073&0.953\\
MMI-$p(t)$&0.090& 0.880\\
Sampling&0.372&0.679\\
Adver-Reinforce&0.080&0.945 \\
Adver-REGS&0.098&0.952\\\bottomrule
\end{tabular}
\caption{{\it AdverSuc} and {\it machine-vs-random} scores achieved by different training/decoding strategies.}
\label{adv}
\end{table}

\subsection{Human Evaluation}
For human evaluation, we follow protocols defined in Chapter 5, employing
crowdsourced judges to evaluate a random sample of
200 items. We present both an input message and the
generated outputs to 3 judges and ask them to decide
which of the two outputs is better ({\it single-turn}
general quality). 
They were instructed to prefer outputs that were more specific (relevant) to the message and preceding context,
Ties are permitted. Identical
strings are assigned the same score. 
We also
 present the judges with {\it multi-turn}
conversations simulated between the two agents. Each conversation
consists of 3 turns.
Results are presented in Table \ref{human}.
We observe a significant quality improvement on both 
single-turn quality and multi-turn quality from the proposed adversarial model.
It is worth noting that the reinforcement learning system described in Chapter 5, which 
simulates conversations between two bots and 
is trained based on manually designed reward functions, only improves multi-turn dialogue quality, while the model described in this chapter improves both single-turn and multi-turn dialogue generation quality. 
This confirms that the reward adopted in adversarial training is more general, natural, and effective in training dialogue systems.

\begin{table}
\small
\centering
\begin{tabular}{cccc}\\\toprule
Setting &adver-win &adver-lose &tie\\\midrule
single-turn& 0.62& 0.18 &0.20 \\
multi-turn& 0.72 &0.10& 0.18\\\bottomrule
\end{tabular}
\caption{The gain
from the proposed adversarial model
 over the mutual information system
based on pairwise human judgments.}
\label{human}
\end{table}

\begin{table*}[!htb]
\centering
\footnotesize
\begin{tabular}{p{4cm}p{8cm}}\toprule
{\bf Input}&maybe his eyes were too big for his stomach .\\
{\bf Vanilla-MLE}&he 's got a big mouth .\\
{\bf Vanilla-Sample}&i 'm sure he 's unhappy with the fact .\\
{\bf Reinforce}&he 's a little bit of a pain .\\
{\bf MC}&i don 't think so .\\\midrule
{\bf Input}&they didn 't have much success with this problem commander .\\
{\bf Vanilla-MLE}&they 're not the only ones who have been in the system .\\
{\bf Vanilla-Sample}&it 's the new priority only .\\
{\bf Reinforce}&they 're not the only ones who can help us .\\
{\bf MC}&can we find someone else ?\\\midrule
{\bf Input}&and he carried them all the time wherever he went .\\
{\bf Vanilla-MLE}&i don 't know who he is .\\
{\bf Vanilla-Sample}&the church ... suffered the winds of 900 meters with horse aid .\\
{\bf Reinforce}&he was a good man .\\
{\bf MC}&he was a very good man .\\\midrule
{\bf Input}&somebody probably flushed their pet snake down the toilet .\\
{\bf Vanilla-MLE}&i 'm sure the dog will be here soon .\\
{\bf Vanilla-Sample}&i could at least think .\\
{\bf Reinforce}&i don 't think so .\\
{\bf MC}&i don 't think so .\\\midrule
{\bf Input}&excuse me but how did we pay for these suits ?\\
{\bf Vanilla-MLE}&i 'm not a beggar .\\
{\bf Vanilla-Sample}&is this the money eugene ?\\
{\bf Reinforce}&i don 't know what to do .\\
{\bf MC}& how much money you owe me ?\\\midrule
{\bf Input}&... while the beautiful and happy world was madly destroyed ;\\
{\bf Vanilla-MLE}&i was born in the world of the world\\
{\bf Vanilla-Sample}&as in childhood i took you to one tenth of a postcard\\
{\bf Reinforce}&and the world was full of joy .\\
{\bf MC}&and the world was full of joy .\\\midrule
{\bf Input}&i was kinda hoping that you were my dad .\\
{\bf Vanilla-MLE}&i 'm just a little bit of a little bit of a little bit of a little bit of a\\
{\bf Vanilla-Sample}&shiny gold gold and silver in the mountains .\\
{\bf Reinforce}&i 'm not gonna be a baby .\\
{\bf MC}&i 'm not going to be a father .\\\midrule
{\bf Input}&the man who does that is not afraid to die .\\
{\bf Vanilla-MLE}&he 's not afraid of the dark .\\
{\bf Vanilla-Sample}&even the program is the only truth\\
{\bf Reinforce}&he 's also not afraid of me .\\
{\bf MC}&he 's such a brave man !\\\bottomrule
\end{tabular}
\caption{Sampled responses outputted from different models}
\label{example-appendix1}
\end{table*}

\begin{table*}[!htb]
\centering
\footnotesize
\begin{tabular}{p{4cm}p{8cm}}\toprule

{\bf Input}&i 'll have to drive home .\\
{\bf Vanilla-MLE}&i 'll be back in a minute .\\
{\bf Vanilla-Sample}&let 's go for some fun .\\
{\bf Reinforce}&i 'll drive you home .\\
{\bf MC}&i don 't have a car .\\\midrule
{\bf Input}&i had to work late and then i went for a walk .\\
{\bf Vanilla-MLE}&i was in the middle of the road .\\
{\bf Vanilla-Sample}&my body was just pounding .\\
{\bf Reinforce}&i was just trying to help you .\\
{\bf MC}&i was just trying to get a little sleep .\\\midrule
{\bf Input}&lots of cream lots of sugar .\\
{\bf Vanilla-MLE}&i 'm gonna be in the kitchen .\\
{\bf Vanilla-Sample}&was this gay or red ?\\
{\bf Reinforce}&i don 't want to be a vegetable .\\
{\bf MC}&i 'm not a very good cook .\\\midrule
{\bf Input}&so they just let him go ?\\
{\bf Vanilla-MLE}&he 's not gonna let them go through the motions .\\
{\bf Vanilla-Sample}&they asked him whether we could go see ernesto ?\\
{\bf Reinforce}&i don 't know what to do .\\
{\bf MC}&he 's not going to be here any longer .\\\midrule
{\bf Input}&you must not bomb this town .\\
{\bf Vanilla-MLE}&you have to go back to the city .\\
{\bf Vanilla-Sample}&the one is a multiple choice .\\
{\bf Reinforce}&I think I have to .\\
{\bf MC}&I think I have to .\\\midrule
{\bf Input}&communism recently failed in eastern europe .\\
{\bf Vanilla-MLE}&the war is not over .\\
{\bf Vanilla-Sample}&there are people who hope that in tokyo they won 't risk losing their departing .\\
{\bf Reinforce}&and the people who are not the same .\\
{\bf MC}&the war is still going on .\\\midrule
{\bf Input}&you should stop in more often .\\
{\bf Vanilla-MLE}&I don 't think so .\\
{\bf Vanilla-Sample}&is it possible as old as you ?\\
{\bf Reinforce}&i 'm not going to be a burden on you .\\
{\bf MC}&i don 't want to be a burden .\\\hline{\bf Input}&he couldn 't have heard us .\\
{\bf Vanilla-MLE}&he was in the middle of the night  .\\
{\bf Vanilla-Sample}&he should have been somewhere out here .\\
{\bf Reinforce}&he 's not a good guy .\\
{\bf MC}&he 's got a lot of nerve.\\\midrule
{\bf Input}&that 's our wake up call .\\
{\bf Vanilla-MLE}&we 're gonna be late for the meeting .\\
{\bf Vanilla-Sample}&c 'mon c 'mon let 's get outta here .\\
{\bf Reinforce}&i 'll be right back .\\
{\bf MC}&we 're gonna have to get to the station .\\
\bottomrule
\end{tabular}
\caption{Sampled responses outputted from different models}
\label{example-appendix2}
\end{table*}

\section{Conclusion }
In this chapter, drawing  intuitions from the Turing test, we propose using an adversarial training approach for response generation. We cast the model in the framework of reinforcement learning and train a generator based on the signal from a discriminator to generate response sequences indistinguishable from human-generated dialogues.
We observe clear performance improvements on multiple metrics from the adversarial training strategy.

\chapter{Learning by  Asking Questions}
\label{AQ}
In the two chapters below, 
our focus switches from the chit-chat systems to interactive QA dialogue systems.
We will talk about how we can 
design an interactive dialogue agent, and how this agent can benefit from interacting with humans. 
The reason why 
we choose to study the issues with human-bot interactions under the context of factoid QA is because it is
easier to formalize the problem and evaluate various aspects of features and performances in a more domain-specific QA task.
But  
chit-chat 
 systems and interactive QA systems are closely related and 
 the methodology we proposed in this chapter is a general one: 
a chit-chat dialogue system can certainly benefit from the  types of interactions that will be studied in the two chapters below. 

In this chapter, we will discuss how we can train a bot to ask questions. 
Think about the following situation: 
when a student is asked a question by a teacher, but is not confident about the answer, 
they may ask for clarification or hints.
A good conversational agent should have this ability to
interact with a dialogue partner (the teacher/user).
However,  recent efforts have mostly focused
 on learning through fixed answers provided in the training set,
rather than through interactions.
In that case, when a learner encounters a confusing situation such as an unknown surface form (phrase or structure),
a semantically complicated sentence or an unknown word, the agent will either make a (usually poor) guess or will
redirect the user to other resources (e.g., a search engine, as in Siri).
Humans, in contrast, can adapt to many situations by asking questions.

In this chapter, 
we study this issue in the context of a knowledge-base based question answering task, where the dialogue agent has access to a knowledge base (KB)
and need to answer a question based on the KB. 
We first identify
 three categories of mistakes a learner can make during 
a question-answering based
dialogue scenario:\footnote{This
list is not exhaustive; for example, we do not address a failure in the dialogue generation stage.}
(1) the learner has problems understanding the surface form of the text of the dialogue partner, e.g., the phrasing of a question;
(2) the learner has a problem 
with reasoning, e.g., they fail to retrieve and connect 
the relevant knowledge to the question at hand;
(3) the learner lacks the knowledge necessary to answer the question in the first place
-- that is,
the knowledge sources the student has access to do not contain the needed information.

All the situations above can be potentially addressed through interaction with
the dialogue partner. 
Such interactions can be used to {\em learn to perform better in future dialogues}.
If a human student has problems understanding a teacher's question, they
 might ask the teacher to clarify the question.
If the student doesn't know where to start, they might
 ask the teacher to point out which known facts are most relevant.
If the student doesn't know the information needed at all, they might
 ask the teacher to tell them the knowledge they're missing, writing it down
 for future use.

In this work, we try to bridge the gap between how a human and
an end-to-end machine learning dialogue agent deal with these situations:
our student has to {\em learn how to learn}.
We hence design a simulator and a set of synthetic tasks in the movie question answering domain
that allow a bot to interact with a teacher to address the issues described above.
Using this framework, we explore how a bot can benefit from interaction by asking questions
 in both offline supervised settings and online reinforcement learning settings,
 as well as how to choose when to ask questions in the latter setting.
In both cases, we find that the learning system improves through interacting with users.

Finally, we validate our approach on real data where the teachers are humans using Amazon Mechanical Turk,
and observe similar results.

\section{Tasks}
\label{sec:tasks}
In this section we describe the dialogue tasks we designed.
They are tailored for the three different situations described in Section 1
that motivate the bot to ask questions:
(1) {\it Question Clarification},
 in which the bot  has problems understanding its dialogue partner's text;
  (2) {\it Knowledge Operation},
in which the bot
 needs to ask for help to perform reasoning steps over an existing knowledge base;
  and (3) {\it Knowledge Acquisition},
  in which the bot's knowledge is incomplete and needs to be filled. 

For our experiments we adapt the WikiMovies dataset \citep{weston2015towards}, which consists
 of roughly 100k questions over 75k entities based on questions
with answers in the open movie dataset (OMDb).
 The training/dev/test sets respectively contain 181638 / 9702 / 9698 examples. The  accuracy metric corresponds to the percentage of times the student gives correct answers to the teacher's questions.

Each dialogue takes place between a teacher and a bot. In this
section we describe how we generate tasks using a simulator. Section \ref{sec:mturk}
 discusses how  we test similar setups with real data using Mechanical Turk.

The bot is first presented with facts from the OMDb KB.
This allows us to control the exact knowledge the bot has access to.
Then, we include several teacher-bot question-answer pairs unrelated to
the question the bot needs to answer, which we call \emph{conversation histories}.\footnote{
These history QA pairs can be viewed as distractions and are used to test the
bot's ability to separate the wheat from the chaff.
For each dialogue, we incorporate 5 extra QA pairs (10 sentences).}
%
In order to explore the benefits of asking clarification questions 
during a conversation, for each of the three scenarios, our 
simulator generated data for two different settings, namely, 
\emph{Question-Answering} (denoted by QA) and 
\emph{Asking-Question} (denoted by AQ).
For both {\it QA} and {\it AQ}, the bot needs to give an answer to the teacher's original question at the end.

\subsection{ Question Clarification.}

The QA-based dialogue system is interactive \cite{rieser2009does}: instead of just having to answer one single question, 
the QA-based system needs the ability to handle a diverse category of interaction-related issues such as 
 asking for question clarification \cite{stoyanchev2013modelling,stoyanchev2014towards}, adapting answers given 
a human's feedback \cite{rieser2009does}, 
self-learning when encountering new questions or concepts
\cite{purver2006clarie}, etc.
To handle these issues, 
 the system needs to 
 take proper  actions based on the current conversation state, which resembles the key issue addressed in the \textsc{state-based dialogue system}.

As discussed in the previous sections, the bot needs to ask the teacher to clarify the question when it does not understand it. 
Actually, the task of {\it ask for clarification} has been studied in a variety of speech recognition 
scenarios
when a speech recognition system fails to understand a user's utterance, such as
use the generic dialog act REJECT and emit a rule-based ``Please repeat",
or ``I don't understand what you said" utterance, or generate more complicated targeted utterances
in different clarification senarios
 \cite{purver2004theory,nerbonne2005interrogative,stoyanchev2015localized}. 

In this setting, the bot does not understand the teacher's question. We focus on a
special situation where the bot does not understand the teacher because of typo/spelling mistakes,
 as shown in Figure \ref{AskC}.
We intentionally misspell some words in the questions such as replacing the word ``movie" with
``{\color{brown}{movvie}}" or ``star" with  ``{\color{brown}{sttar}}".\footnote{Many
reasons could lead to the bot not understanding the teacher's question, e.g.,
the teacher's question has an unknown phrase structure, rather than unknown words.
We choose to use spelling mistakes because of the ease of dataset construction.}
To make sure that the bot will have problems understanding the question, we guarantee that the bot
has never encountered the misspellings before---the misspelling-introducing mechanisms
in the training, dev, and test sets are different, so the same word will be misspelled in different ways in different sets.
We present two {\it AQ} tasks:
(i) {\it Question Paraphrase} where the student asks the teacher to use a paraphrase that
does not contain spelling mistakes to clarify the question by asking ``what do you mean?'';
and (ii) {\it Question Verification} where the student  asks the teacher
whether the original typo-bearing question corresponds to another question
without the spelling mistakes (e.g., ``Do you mean which film did Tom Hanks appear in?").
The teacher will give
feedback by giving a paraphrase of the original question without spelling mistakes
(e.g., ``I mean which film did Tom Hanks appear in") in {\it Question Paraphrase}
or positive/negative feedback in {\it Question Verification}.
Next the student will give an answer and the teacher will give positive/negative
feedback depending on whether the student's answer is correct.
Positive and negative feedback are variants of ``No, that's incorrect" or
``Yes, that's right".\footnote{In the datasets we build, there are 6 templates
for positive feedback and 6 templates for negative feedback.}
In these tasks, the bot has access to all relevant entries in the KB.

\subsection{Knowledge Operation} \label{sec:knolop}
The bot has  access to all the relevant knowledge (facts)
 but lacks the ability to perform necessary reasoning operations over them;
see Figure \ref{AskV}.
We focus on a special case where the bot will try to understand what are the relevant
facts. We explore two settings:
 {\it Ask For Relevant Knowledge} (Task 3) where the bot directly asks the teacher to point out the relevant KB fact and {\it Knowledge Verification} (Task 4)  where the bot asks whether the teacher's question is relevant to one particular KB fact. The teacher will point out the relevant KB fact in the {\it Ask For Relevant Knowledge}  setting
 or give a positive or negative response in the {\it Knowledge Verification} setting.
 Then the bot will give an answer to the teacher's original question
and the teacher will give feedback on the answer.

\begin{figure*}[!ht]
\centering
\footnotesize
\begin{tabular}{|l|l|l|}
\cline{1-1}\cline{3-3}
 \\[-2ex]
{\bf  Task 1:  Question Paraphrase (QA)}  &&  {\bf Task 1:  Question Paraphrase (AQ):} \\
 kb: \underline{Larry Crowne} {directed\textunderscore by} \underline{Tom Hanks} && kb: \underline{Larry Crowne} {directed\textunderscore by} \underline{Tom Hanks} \\
kb: \underline{Forrest Gump} {starred\textunderscore actors} \underline{Tom Hanks}, && kb: \underline{Forrest Gump} {starred\textunderscore actors} \underline{Tom Hanks},   \\
 \underline{Robin Wright}, \underline{Gary Sinise} &&  \underline{Robin Wright}, \underline{Gary Sinise} \\
 kb: \underline{Forrest Gump} directed\textunderscore  by \underline{Robert Zemeckis} && kb: \underline{Forrest Gump} directed\textunderscore  by \underline{Robert Zemeckis}  \\
{\color{blue}T}/{\color{red}S }:  {Conversation History}.&&{\color{blue}T}/{\color{red}S }:  {Conversation History}.\\
{\color{blue}T : Which  {\color{brown} movvie} did Tom Hanks {\color{brown} sttar} in ?} &&{\color{blue}T : Which  {\color{brown}movvie} did Tom Hanks  {\color{brown}sttar} in ?}  \\
{\color{red}S : Larry Crowne }  &&  {\color{red}S : What do you mean ?}
 \\
{\color{blue}T : Wrong. (-)} && {\color{blue}T : I mean which film did Tom Hanks appear in.} \\
&&{\color{blue}T : Which  {\color{brown} movvie} did Tom Hanks {\color{brown} sttar} in ?}  \\
&&{\color{red}S : Forrest Gump }   \\
&& {\color{blue}T : That's correct. (+)} \\
\cline{1-1}\cline{3-3}
 \multicolumn{1}{c}{}
\vspace{-2mm} \\
\cline{1-1}\cline{3-3}
{\bf Task 2:  Question Verification (QA):} && {\bf Task 2: Question Verification (AQ)} \\
 kb: \underline{Larry Crowne} {directed\textunderscore by} \underline{Tom Hanks} && kb: \underline{Larry Crowne} {directed\textunderscore by} \underline{Tom Hanks} \\
kb: \underline{Forrest Gump} {starred\textunderscore actors} \underline{Tom Hanks}, && kb: \underline{Forrest Gump} {starred\textunderscore actors} \underline{Tom Hanks},   \\
 \underline{Robin Wright}, \underline{Gary Sinise} &&  \underline{Robin Wright}, \underline{Gary Sinise} \\
 kb: \underline{Forrest Gump} directed\textunderscore  by \underline{Robert Zemeckis} && kb: \underline{Forrest Gump} directed\textunderscore  by \underline{Robert Zemeckis}  \\
{\color{blue}T}/{\color{red}S }:  {Conversation History}.&&{\color{blue}T}/{\color{red}S }:  {Conversation History}.\\
{\color{blue}T : Which  {\color{brown} movvie} did Tom Hanks {\color{brown} sttar} in ?} &&{\color{blue}T : Which  {\color{brown}movvie} did Tom Hanks  {\color{brown}sttar} in ?}  \\
{\color{red}S : Forrest Gump } && {\color{red}S : Do you mean which film did Tom Hanks appear in ?}\\
{\color{blue}T : That's correct. (+)}&&{\color{blue}T : That's correct. (+)} \\
 &&{\color{blue}T : Which  {\color{brown} movvie} did Tom Hanks {\color{brown} sttar} in ?} \\
&&{\color{red}S : Larry Crowne }   \\
&& {\color{blue}T : Sorry, that's not right. (-)} \\
\cline{1-1}\cline{3-3}
\end{tabular}
\caption[Sample dialogues for the {\it Question Clarification} tasks]{Sample dialogues for the two {\it Question Clarification} tasks (rows) using both the traditional {\it QA} setting
(left column) and {\it AQ} setting (right column).
In each case the same example is given for simplicity. Black text prefixed by ``kb:'' denotes KB knowledge that the student has access to.
{\color{blue}Blue} text is spoken by the teacher, denoted by {\color{blue}T}. {\color{blue}{(+)}} denotes a reward of 1
(and 0 otherwise) that the teacher assigns to the bot.
{\color{red}Red} text denotes responses or questions posed by the bot, denoted by  {\color{red}S}.
 {\color{brown}{Brown}} denotes typos deliberately introduced by the authors. For the {\it Question Verification} setting,
the student can either ask a correct (pertinent) question (as in this example) or an incorrect (irrelevant) one.
The teacher will give positive or negative feedback based on the correctness of the student's question.
In our offline superised learning experiments, the probability of asking pertinent questions and correctly answering the original
question from the teacher is set to 0.5. Finally, {\color{blue}T}/{\color{red}S }
denotes 5 pairs of questions and answers that are irrelevant to the rest of the conversation.
}
\label{AskC}
\end{figure*}
\begin{figure*}[!ht]
\centering
\footnotesize
\begin{tabular}{|l|l|l|}
 \multicolumn{1}{c}{}
\vspace{-2mm} \\
\cline{1-1}\cline{3-3}
 {\bf Task 3: Ask For Relevant Knowledge (AQ): }  && {\bf Task 4: Knowledge Verification (AQ):} \\
 kb: \underline{Larry Crowne} {directed\textunderscore by} \underline{Tom Hanks} && kb: \underline{Larry Crowne} {directed\textunderscore by} \underline{Tom Hanks} \\
kb: \underline{Forrest Gump} {starred\textunderscore actors} \underline{Tom Hanks}, && kb: \underline{Forrest Gump} {starred\textunderscore actors} \underline{Tom Hanks},   \\
 \underline{Robin Wright}, \underline{Gary Sinise} &&  \underline{Robin Wright}, \underline{Gary Sinise} \\
 kb: \underline{Forrest Gump} directed\textunderscore  by \underline{Robert Zemeckis} && kb: \underline{Forrest Gump} directed\textunderscore  by \underline{Robert Zemeckis}  \\

{\color{blue}T}/{\color{red}S }:  {Conversation History}.&&{\color{blue}T}/{\color{red}S }:  {Conversation History}.\\
{\color{blue}T : Which movie did Tom Hanks star in ?} &&{\color{blue}T : Which movie did Tom Hanks star in ?}  \\
 {\color{red}S :  Can you give me a hint ?}&& {\color{red}S : Does it have something to do with } \\
{\color{blue}T : It is related to the kb fact that} && ~~~~~{\color{red}\underline{Larry Crowne} {directed\textunderscore by} \underline{Tom Hanks}?}  \\
~~~~~{\color{blue}{\underline{Forrest Gump} {starred\textunderscore actors} \underline{Tom Hanks} }} &&{\color{blue}T : No. That's wrong. (-)} \\
 ~~~~~{\color{blue} \underline{Robin Wright}, \underline{Gary Sinise} }  &&{\color{blue}T : Which movie did Tom Hanks star in ?}\\
{\color{blue}T : Which movie did Tom Hanks star in ? }&&{\color{red}S : Forrest Gump } \\
{\color{red}S : Forrest Gump }  && {\color{blue}T : That's correct. (+)} \\
 {\color{blue}T : That's correct. (+)}  && \\
\cline{1-1}\cline{3-3}

\end{tabular}
\caption{Sample dialogues for {\it Knowledge Operation} tasks. }
\label{AskV}
\end{figure*}

\begin{figure*}[!ht]
\footnotesize
\label{MissE}
\begin{center}
\begin{tabular}{|l|l|l|}
 \multicolumn{3}{l}{{\bf }. }\\
\cline{1-1}\cline{3-3}
{\bf Task 5: Missing Question Entity (QA) }  &&{\bf  Task 5: Missing Question Entity (AQ)} \\
{\st{kb: Larry Crowne directed\textunderscore by Tom Hanks }}&& {\st{kb: Larry Crowne directed\textunderscore by Tom Hanks }}\\
{\st{kb: Forrest Gump starred\textunderscore actors Tom Hanks}} &&{\st{kb: Forrest Gump starred\textunderscore actors Tom Hanks, } } \\
kb: {Forrest Gump} starred\textunderscore actors {Sally Field} &&
kb: {Forrest Gump} starred\textunderscore actors {Sally Field}\\
kb: {Forrest Gump} directed\textunderscore  by {Robert Zemeckis} && kb: {Forrest Gump} directed\textunderscore  by {Robert Zemeckis}  \\
{\color{blue}T}/{\color{red}S }:  {Conversation History}. &&{\color{blue}T : Which movie did Tom Hanks star in ?}\\
{\color{blue}T : Which movie did Tom Hanks star in ?}&& {\color{red}S : I don't know. What's the answer?}\\
{\color{red}S : Forrest Gump }  && {\color{blue}T : The answer is Forrest Gump. }\\
{\color{blue}T : That's correct. (+)}&& {\color{blue}T}/{\color{red}S }:  {Conversation History}.  \\
&& {\color{blue}T : Which movie did Tom Hanks star in ?} \\
&& {\color{red}S : Forrest Gump }  \\
&& {\color{blue}T : That's correct. (+)} \\
\cline{1-1}\cline{3-3}
\end{tabular}
\end{center}
\begin{center}
\end{center}

\begin{center}
\begin{tabular}{|l|l|l|}
\cline{1-1}\cline{3-3}
{\bf  Task 6: Missing Answer Entity (AQ)} &&{\bf  Task 7: Missing Relation Entity (AQ)} \\
 kb: {Larry Crowne} {directed\textunderscore by} {Tom Hanks} && kb: {Larry Crowne} {directed\textunderscore by} {Tom Hanks} \\
{\st{kb: Forrest Gump starred\textunderscore actors Tom Hanks}} &&{\st{kb: Forrest Gump starred\textunderscore actors Tom Hanks, } } \\
{\st{kb: Forrest Gump starred\textunderscore actors Sally Field}}&&
{\st{kb: Forrest Gump starred\textunderscore actors Sally Field}}\\
{\st{kb: Forrest Gump directed\textunderscore  by Robert Zemeckis}} && kb: {Forrest Gump} directed\textunderscore  by {Robert Zemeckis}  \\
{\color{blue}T : Which movie did Tom Hanks star in ?} && {\color{blue}T : Which movie did Tom Hanks star in ?}\\
 {\color{red}S : I don't know. What's the answer?}&& {\color{red}S : I don't know. What's the answer?}\\
 {\color{blue}T : The answer is Forrest Gump. }\ && {\color{blue}T : The answer is Forrest Gump. }\\
 {\color{blue}T}/{\color{red}S }:  {Conversation History}.&& {\color{blue}T}/{\color{red}S }:  {Conversation History}.   \\
{\color{blue}T : Which movie did Tom Hanks star in ?}&&{\color{blue}T : Which movie did Tom Hanks star in ?}\\
{\color{red}S : Forrest Gump }  && {\color{red}S : Forrest Gump }  \\
{\color{blue}T : That's correct. (+)} &&{\color{blue}T : That's correct. (+)}  \\
\cline{1-1}\cline{3-3}
\end{tabular}
\end{center}
\begin{center}
\begin{tabular}{|l|l|l|}
 \multicolumn{3}{l}{{\bf }. }\\
\cline{1-1}\cline{3-3}
{\bf  Task 8: Missing Triple (AQ)}   &&{\bf  Task 9: Missing Everything (AQ)} \\
{kb: Larry Crowne directed\textunderscore by Tom Hanks}&& {\st{kb: Larry Crowne directed\textunderscore by Tom Hanks}} \\
{\st{kb: Forrest Gump starred\textunderscore actors Tom Hanks}} &&{\st{kb: Forrest Gump starred\textunderscore actors Tom Hanks, } } \\
{kb: Forrest Gump starred\textunderscore actors Sally Field}&&
{\st{kb: Forrest Gump starred\textunderscore actors Sally Field}}\\
{kb: Forrest Gump directed\textunderscore  by Robert Zemeckis} &&{\st{kb: Forrest Gump directed\textunderscore  by Robert Zemeckis}}  \\
{\color{blue}T : Which movie did Tom Hanks star in ?} &&{\color{blue}T : Which movie did Tom Hanks star in ?}\\
 {\color{red}S : I don't know. What's the answer?}&& {\color{red}S : I don't know. What's the answer?}\\
 {\color{blue}T : The answer is Forrest Gump. }&& {\color{blue}T : The answer is Forrest Gump. }\\
  {\color{blue}T}/{\color{red}S }:  {Conversation History}. &&  {\color{blue}T}/{\color{red}S }:  {Conversation History}. \\
 {\color{blue}T : Which movie did Tom Hanks star in ?} && {\color{blue}T : Which movie did Tom Hanks star in ?} \\
{\color{red}S : Forrest Gump }  && {\color{red}S : Forrest Gump }  \\
 {\color{blue}T : That's correct. (+)} && {\color{blue}T : That's correct. (+)} \\
\cline{1-1}\cline{3-3}
\end{tabular}
\end{center}
\caption[Sample dialogues for {\it Knowledge Acquisition} tasks]{Different tasks for {\it Knowledge Acquisition}. Crossed lines correspond to entries of the KB whose retrieval is missed.}
\end{figure*}

\subsection{Knowledge Acquisition} \label{sec:knolaq}
For the tasks in this subsection, the bot has an
incomplete KB and there are entities important to the dialogue missing from it,
see Figure \ref{MissE}.
For example, given the question  ``Which movie did Tom Hanks star in?",
the missing part could either be the entity that the teacher is asking about
(question entity for short, which is \underline{Tom Hanks} in this example),
the relation entity (\underline{starred\_actors}), the answer to the question (\underline{Forrest Gump}),
or the combination of the three.
In all cases, the bot has little chance of giving the correct answer due to the missing knowledge.
It needs to ask the teacher the answer to acquire the missing knowledge.
The teacher will give the answer and then move on to other questions (captured in the conversational history). They later will come back to reask the question.
At this point, the bot needs to give an answer since the entity is not new any more.

Though the correct answer has effectively been included in the earlier part of the dialogue as the answer to the bot's question, as we will show later, many of the tasks are not as trivial as they look when the teacher reasks the question. This is because the bot's model needs to memorize the missing entity and then construct the links between the missing entities and known ones. This is akin to the real world case where a student might make the same mistake again and again even though each time the teacher corrects them if their answer is wrong. We now detail each task in turn.

{\bf Missing Question Entity}:    The entity
that
the teacher is asking about is missing from the knowledge base.
All KB facts containing the question entity will be hidden from the bot. In the example for {\it Task 5} in Figure \ref{MissE}, since the teacher's question contains the entity \underline{Tom Hanks}, the KB facts that contain \underline{Tom Hanks} are  hidden from the bot.

{\bf Missing Answer Entity}:
The answer entity to the question is unknown to the bot.
All KB facts that contain the answer entity will be hidden. Hence, in {\it Task 6} of Figure \ref{MissE}, all KB facts containing the answer entity \underline{Forrest Gump} will be hidden from the bot.

{\bf Missing Relation Entity}:
The relation type is unknown to the bot.
In {\it Task 7} of Figure \ref{MissE},
all KB facts that express the relation \underline{starred actors} are hidden from the bot.

{\bf Missing Triples}:
The triple that expresses the relation between the question entity and the answer entity is hidden from the bot. In {\it Task 8} of Figure \ref{MissE}, the triple ``\underline{Forrest Gump} (question entity)
starred\textunderscore actors  \underline{Tom Hanks} (answer entity)" will be hidden.

{\bf Missing Everything}: The question entity, the relation entity, the answer entity are
all missing from the KB. All KB facts in
{\it Task 9} of Figure \ref{MissE} will be removed since they either contain the relation entity (i.e., starred\textunderscore actors), the question entity (i.e., \underline{Forrest Gump}) or the answer entity \underline{Tom Hanks}.

\section{Train/Test Regime}
\label{Train_Test}
We now discuss in detail the regimes we used to train and test our models,
which are divided between evaluation within our simulator and using
real  data collected via Mechanical Turk.

\subsection{Simulator}

\label{sec:train-test}

Using our simulator, our objective was twofold.
We first wanted to validate the usefulness of asking questions 
in all the settings described in the previous section.
Second, we wanted to assess the ability of our student bot 
to learn {\em when} to ask questions.
In order to accomplish these two
objectives we explored training our models with our simulator using two methodologies,
namely, Offline Supervised Learning and Online Reinforcement Learning.

\subsubsection{Dialogue Simulator}
Here we  detail 
the simulator and 
the datasets we generated in order to realize the various scenarios 
discussed in Section~\ref{sec:tasks}. 
We focused on the problem of movieQA where we adapted the 
WikiMovies dataset proposed in \cite{weston2015towards}. The 
dataset consists of roughly $100$k questions with over $75$k entities 
from the open movie dataset (OMDb). 

Each dialogue generated by the simulator takes place between 
a student and a teacher.
The simulator samples a random question from the 
WikiMovies dataset and fetches the set of all KB facts
relevant to the chosen question. This question is assumed 
to be the one the teacher asks its student, and is referred to as 
the ``original" question. The student is first presented with the 
relevant KB facts followed by the original question. Providing the 
KB facts to the student allows us to control the exact knowledge 
the student is given access to while answering the questions. 
At this point, depending on the 
task at hand and the student's ability to answer, the 
student might choose to directly answer it or ask a ``followup"
question. The nature of the followup question will depend on 
the scenario under consideration. If the student 
answers the question, it gets a response from the teacher about 
its correctness and the conversation ends. However if the student 
poses a followup question, the teacher gives an appropriate 
response, which should give additional information to the student 
to answer the original question. In order to make things more 
complicated, the simulator pads the 
conversation with several unrelated student-teacher 
question-answer pairs. These question-answer pairs can be 
viewed as distractions and are 
used to test the student's ability to remember the additional 
knowledge provided by the teacher after it was queried. 
For each dialogue, the simulator incorporates $5$ such pairs 
($10$ sentences). We refer to these pairs as \emph{conversational 
histories}. 

For the \emph{QA} setting (see Section~\ref{sec:tasks}), the 
dialogues generated by the simulator are such that the student 
never asks a clarification question. Instead, it simply responds 
to the original question, even if it is wrong. For the dialogs in 
the \emph{AQ} setting, the student {\em always} asks a clarification 
question. The nature of the question asked is dependent on 
the scenario (whether it is 
\emph{Question Clarification}, 
\emph{Knowledge Operation}, or 
\emph{Knowledge Acquisition}) under consideration. 
In order to simulate the 
case where the student sometimes choses to directly answer 
the original question and at other times choses to ask question, 
we created training datasets, which were a combination of 
\emph{QA} and \emph{AQ} (called ``Mixed''). For all these cases, the student 
needs to give an answer to the teacher's original question at 
the end.

\subsubsection{Offline Supervised Learning} \label{sec:offline}
The motivation behind training our student models in an offline supervised
setting was primarily to test the usefulness of the ability to ask questions.
The dialogues are generated as described in the previous section,
and the bot's role is generated with a fixed policy.
We chose a policy where answers to the teacher's questions are
correct answers 50\% of the time, and incorrect
otherwise, to add a degree of realism.
Similarly, in tasks where questions can be irrelevant
they are only asked correctly 50\% of the time.\footnote{This only makes sense
in tasks like Question or Knowledge Verification. In tasks where the question is
static such as `What do you mean?'' there is no way to ask an irrelevant question,
and we do not use this policy.}

The offline setting  explores different combinations of training and testing
scenarios, which mimic different situations in the real world.
The aim is to understand when and how observing interactions between two agents
can help the bot improve its performance for different tasks. As a result
we construct training and test sets in three ways across all tasks,
resulting in $9$ different scenarios per task, each of which correspond to
a real world scenario.

The three training sets we generated are referred to as
{\bf TrainQA, TrainAQ, and TrainMix}. TrainQA
 follows the QA setting discussed in the previous section:
the bot never asks questions and only tries to immediately answer.
TrainAQ follows the AQ setting: the student, before answering,
first always asks a question in response to the teacher's original question.
TrainMix is a combination of the two
where $50\%$ of time the student asks a question and $50\%$ does not.

The three test sets we generated are referred to as {\bf TestQA, TestAQ,
and TestModelAQ}. TestQA and TestAQ are generated similarly to TrainQA and TrainAQ,
but using a perfect fixed policy (rather than 50\% correct) for evaluation purposes.
In the TestModelAQ setting the model has to get the form of the question correct as well. 
In the  {\em Question Verification} and {\em Knowledge Verification} tasks there are many possible ways of forming the question and some of them are correct -- the model has to choose the right question to ask. E.\ g.\ it should ask ``Does it have something to do with the fact that Larry Crowne directed by Tom Hanks?''rather than ``Does it have something to do with the fact that Forrest Gump directed by Robert Zemeckis?'' when the latter is irrelevant (the candidate list of questions is generated from the known knowledge base entries with respect to that question).
The policy is trained
using either the TrainAQ or TrainMix set, depending on the training
scenario. The teacher will reply to the question, giving positive feedback
if the student's question is correct and no response and negative feedback otherwise. The
student will then give the final answer.
The difference between TestModelAQ and TestAQ only exists in
the {\it Question Verification} and {\it Knowledge Verification} tasks;
in other tasks there is only one way to ask the question
 and TestModelAQ and TestAQ are identical.

To summarize, for every task listed in Section~\ref{sec:tasks} we train
one model for each of the three training sets (TrainQA, TrainAQ, TrainMix)
and test each of these models on the three test sets (TestQA, TestAQ, and
TestModelAQ), resulting in $9$ combinations.
For the purpose of notation the train/test combination is denoted by
``TrainSetting+TestSetting". For example, {\it TrainAQ+TestQA} denotes
a model which is trained using the TrainAQ dataset and tested on TestQA
dataset. Each combination has a real world interpretation.  For instance,
{\it TrainAQ+TestQA} would refer to a scenario where a student can ask
the teacher questions during learning but cannot to do so while taking an
exam. Similarly, {\it TrainQA+TestQA} describes a stoic teacher that
never answers a student's question at either learning or examination time.
The setting {\it TrainQA+TestAQ} corresponds to the case where a lazy
student never asks question at learning time but gets anxious during the
examination and always asks a question.

\subsubsection{Online Reinforcement Learning (RL)}
We also explored scenarios where the student
learns the ability to decide when to ask a question. In other
words, the student learns how to learn.

Although it is in the interest of the student to ask questions at every
step of the conversation, since the response to its question will contain
extra information, we don't want our model to learn this behavior.
Each time a human student asks a question, there's a cost associated with that
action. This cost is a reflection of the patience of the teacher,
or more generally of the users interacting with the bot in the wild:
users won't find the bot engaging if it always
asks clarification questions. The student should thus be judicious about
asking questions and learn when and what to ask.
For instance, if the student is confident about the answer, there is no need
for it to ask. Or, if the teacher's question is so hard
that clarification is unlikely to help enough to get the answer right,
then it should also refrain from asking.

We now discuss how we model this problem under the Reinforcement Learning
framework. The bot is presented with KB facts (some facts might be missing
depending on the task) and a question. It needs to decide whether to ask a
question or not at this point. The decision  whether to ask
is made by a binary policy $P_{RLQuestion}$. If
the student  chooses to ask a question, it will be penalized by $cost_{AQ}$.
We explored different values of $cost_{AQ}$ ranging from $[0, 2]$, which we consider
as modeling the patience of the teacher.
The goal of this setting is to find the best policy for asking/not-asking questions
which would lead to the highest cumulative reward.
The teacher will appropriately reply if the student asks a question.
The student will eventually give an answer to the teacher's initial question
at the end using the policy $P_{RLAnswer}$, regardless of whether it had asked a
question. The student will get a reward of $+1$ if its final answer
is correct and  $-1$ otherwise. 
Note that the student can ask at most one question and that the type of
question is always specified by the task under consideration.
The final reward the student gets is the cumulative reward over the
current dialogue episode. In particular the reward structure we propose is the
following:

\begin{table}[!ht]
\small
\begin{tabular}{rcc}
&Asking Question &Not asking Question \\
Final Answer Correct & 1-$cost_{AQ}$ & 1\\
Final Answer Incorrect & -1-$cost_{AQ}$ & -1\\
\end{tabular}
\caption{Reward structure for the Reinforcement Learning setting.}
\label{reward}
\end{table}

\noindent For each of the tasks described in Section~\ref{sec:tasks}, we consider
three different RL scenarios.\\
{\bf  Good-Student}: The student will be presented with
all relevant KB facts. There are no misspellings or unknown words in
the teacher's question. This represents a knowledgable student in the real
world that knows as much as it needs to know (e.g., a large knowledge
base, large vocabulary). This setting is identical across all missing entity
tasks (5 - 9). \\
{\bf  Poor-Student}: The KB facts or the questions presented to the
student are flawed depending on each task. For example,
for the {\it Question Clarification} tasks, the student does not understand the
question due to spelling mistakes. For the  {\it Missing Question Entity}
task the entity that the teacher asks about is unknown by the student and
all facts containing the entity will be hidden from the student.  This setting is
similar to a student that is underprepared for the tasks. \\
{\bf Medium-Student}: The combination of the previous two settings where
for $50\%$ of the questions, the student has access to the full KB and there
are no new words or phrases or entities in the question,
and $50\%$ of the time the question and KB are taken from the
{\it Poor-Student} setting.


\vspace{-3mm}
\begin{figure*}[!ht]
    \centering
    \includegraphics[width=4.5in]{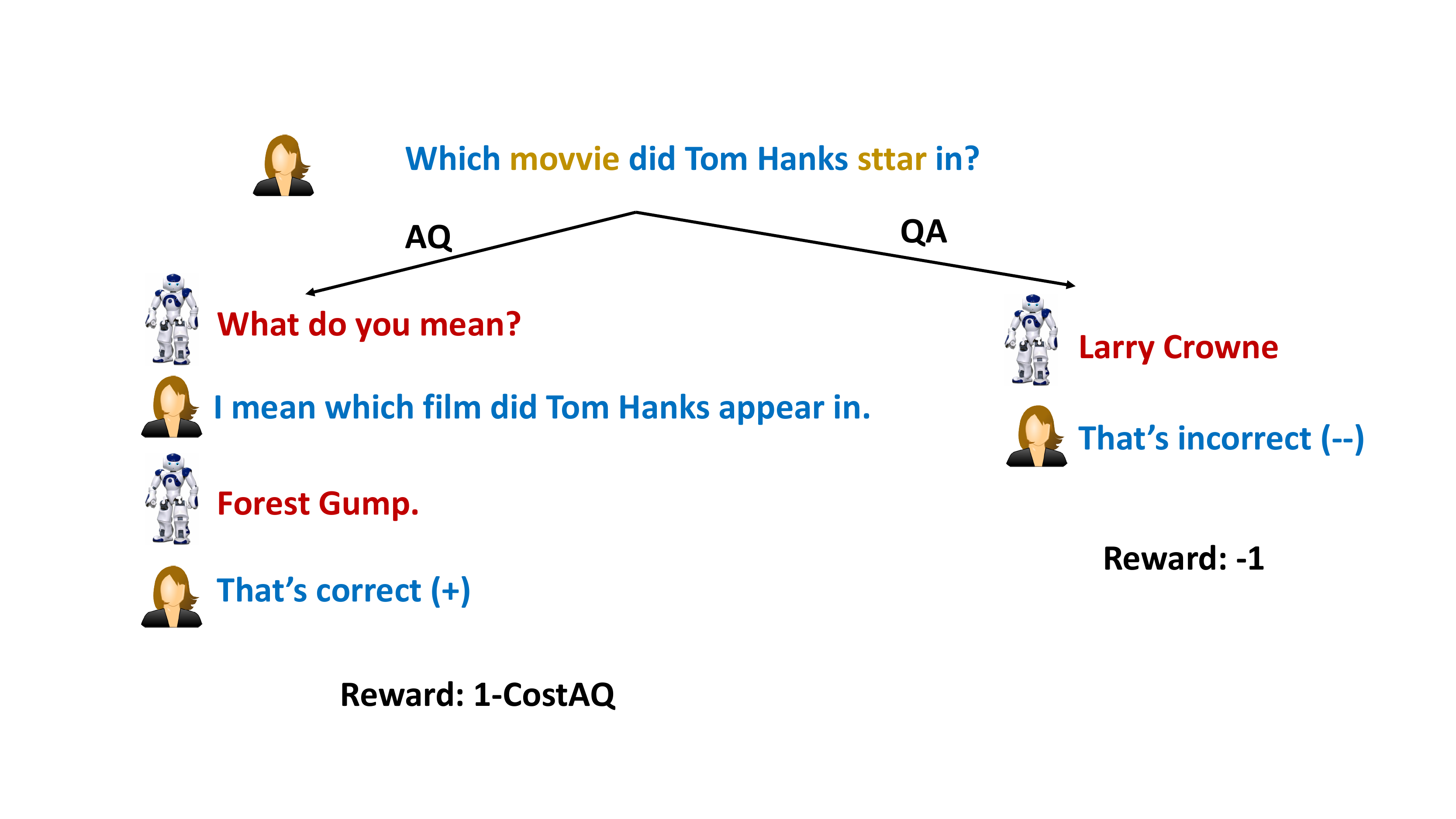}
    \vspace{-8mm}
\caption{An illustration of the {\it poor-student } setting for RL Task 1 (Question Paraphrase). }
\label{RL_illustration}
\end{figure*}

\subsection{Mechanical Turk Data} \label{sec:mturk}

Finally, to validate our approach beyond our simulator by using real language, we collected
data via Amazon Mechanical Turk.
Due to the cost of data collection, we focused on real language versions of
Tasks 4 (Knowledge Verification) and 8 (Missing Triple), 
see Secs. \ref{sec:knolop} and \ref{sec:knolaq} for the simulator versions.
That is, we collect dialogues and use them in an offline supervised learning setup
similar to Section \ref{sec:offline}. This setup allows easily reproducible experiments.

For Mechanical Turk Task 4, the bot is asked a question by a human teacher,
but before answering can ask the human if the question is related
to one of the facts it knows about from its memory. It is then required to answer the original question,
 after some additional dialog turns relating to other question/answer pairs
(called ``conversational history'', as before).
For Task 8, the bot is asked a question by a human but lacks the triple in its memory that
would be needed to answer it. 
It is allowed to ask for the missing information, the human responds to the question in free-form language.
The bot is then required to answer the original question,
 again after some ``conversational history'' has transpired.

We collect around 10,000 episodes (dialogues) for training, 1000 for validation, 
and 2500 for testing for each of the two tasks. In each case, we give
 instructions to the Turkers that still follow the original form of the task, 
but make the tasks contain realistic language written by humans.

For both tasks, while the human turkers replace the simulator that the bot was previously conversing with,
the bot's dialogue actions (capabilities) are essentially unchanged from before. 
That is, when answering questions, now the bot is required to answer a human's questions
rather than templated questions from the simulator.
When the bot is asking questions,
the bot still asks in the same form as before, e.g., questions like
``Does it have something to do with X'' for Task 4
or ``I don't know. What's the answer?'' for Task 8. However,  now its questions are answered
by a human. In both cases (asking and answering) the human data is richer
with potentially more complex language and lexical variability. Examples
of the collected dialogues are given in Figure \ref{example-MT-Tasks}.

\begin{figure*}[!ht]
\centering
\footnotesize
\begin{tabular}{|l|}
\cline{1-1}
 {\bf MTurk Task 4:  Knowledge Verification (AQ): } \\
~~kb:  In Their Skin written\_by Joshua Close, Jeremy Power Regimbal      \\
~~kb:  In Their Skin starred\_actors Selma Blair, James D'arcy, Rachel Miner, Joshua Close \\
~~{\color{blue}T: What film was written by Joshua Close?}                                           \\
~~{\color{red}S: Does it have something to do with ``In Their Skin starred\_actors Joshua Close''?}  \\
~~{\color{blue}T: No we were looking for who wrote the film not who starred in it.}                \\
~~\dots\\
\cline{1-1}
\cline{1-1}
{\bf MTurk Task 8: Missing Triple (AQ):} \\
~~{\color{blue}T :What movies does Timothy Scott Bogart have writing credit for?}\\
~~{\color{red}S:  I don't know. What's the answer?} \\
~~{\color{blue}T: Extreme Ops or The I Inside is the right answer.} \\
~~\dots\\
\cline{1-1}
\end{tabular}
\caption[Sample dialogues for Mechanical Turk versions of Tasks 4 and 8]{Sample dialogues for Mechanical Turk versions of Tasks 4 and 8. Compared to the original
tasks (see Figs \ref{AskV} and \ref{MissE}) the teacher's questions, and the teacher
 responses to the student's questions,
 are written by humans and are more complex and contain more variety.
\label{example-MT-Tasks}
}
\end{figure*}

\section{Models}
For both  offline supervised and online RL settings, we use the {\it End-to-End Memory Network}
model (MemN2N) \citep{sukhbaatar2015end} as a backbone. The
model takes as input the last utterance of the dialogue history (the question
from the teacher) as well as a set of memory contexts including
short-term memories (the dialogue history between the bot and the teacher) and
long-term memories (the knowledge base facts that the bot has access to),
and outputs a label.

{\bf Offline Supervised Settings}:
The first learning strategy we adopt is
the reward-based imitation strategy (denoted {\it vanilla-MemN2N}) described in \newcite{weston2016dialog}, where
at training time, the model maximizes the log likelihood probability
of the correct answers the student gave (examples with incorrect final answers are discarded).
Candidate answers are words that appear in the memories,
which means the bot can only predict the entities that it has seen or known before.

We also use a variation of MemN2N called ``context MemN2N'' ({\it Cont-MemN2N} for short)
where we replace each word's embedding with the average of its embedding (random for
unseen words) and the embeddings of the other words that appear around it.
We use both the preceeding and following words as context and the number of context words is
a hyperparameter selected on the dev set.

An issue with both {\it vanilla-MemN2N} and {\it Cont-MemN2N}  is that the model only makes use of
the bot's answers as signals and ignores the teacher's feedback.
We thus propose to use a  model that jointly predicts the bot's answers and the teacher's feedback
(denoted as {\it TrainQA (+FP)}).
The bot's answers are predicted using a {\it vanilla-MemN2N}  and the teacher's feedback is
predicted using the {\it Forward Prediction} (FP) model as described in \newcite{weston2016dialog}.
At training time, the models learn to jointly predict the teacher's feedback
and the answers with positive reward. At test time, the model will only predict the bot's answer.

For the
{\it TestModelAQ} setting described in Section \ref{Train_Test}, the model
 needs to decide the question to ask.
Again, we use vanilla-MemN2N that takes as input the question and contexts,
and outputs the question the bot will ask.

{\bf Online RL Settings}:
A binary vanilla-MemN2N (denoted as $P_{RL}(Question)$) is used to decide whether
the bot should or should not ask a question, with the teacher replying if the bot does ask something.
A second MemN2N
is then used to decide the bot's answer, 
denoted as $P_{RL}(Answer)$.
$P_{RL}(Answer)$ for $QA$ and $AQ$ are two separate models, which means the bot will use different  models
for final-answer prediction
depending on whether it
 chooses to ask a question or not.\footnote{An alternative is to train one single model
for final answer prediction in both {\it AQ} and {\it QA} cases, similar to the {\it TrainMix} setting in the supervised learning setting. But we find training {\it AQ} and {\it QA} separately for the final answer prediction yields a little better result than the single model setting.}

We use the REINFORCE algorithm \citep{williams1992simple} to update $P_{RL}(Question)$ and  $P_{RL}(Answer)$.
For each dialogue,
the bot takes two sequential actions $(a_1,a_2)$:
to ask or not to ask a question (denoted as $a_1$); and guessing the final answer (denoted as $a_2$).
Let $r(a_1,a_2)$ denote the cumulative reward for the dialogue episode, computed using Table \ref{reward}.
The gradient to update the policy is given by:
\begin{equation}
\begin{aligned}
&p(a_1,a_2)=P_{RL}(Question)(a_1) \cdot  P_{RL}(answer)(a_2)\\
&\nabla J(\theta)\approx \nabla\log p(a_1,a_2) [r(a_1,a_2)-b]
\end{aligned}
\end{equation}
where $b$ is the baseline value, which is estimated using another MemN2N model
that takes as input the query $x$ and memory $C$, and outputs a scalar $b$ denoting the estimation of the future reward.
The baseline model is trained by minimizing the mean squared loss between the estimated reward $b$
and actual cumulative reward $r$, $||r-b||^2$.
We refer the readers to \newcite{zaremba2015reinforcement} for more details.
The baseline estimator model is independent from the policy models and the
error is not backpropagated back to them.

In practice, we find the following training strategy yields better results:
first train only $P_{RL}(answer)$, updating gradients only for the policy that
predicts the final answer.
After the bot's final-answer policy is sufficiently learned, train both policies in parallel.\footnote{
We implement this by running 16 epochs in total, updating only the model's policy for final answers in the first 8 epochs while updating both policies during the second 8 epochs. We pick the model that achieves the best reward on the dev set during the final 8 epochs. Due to relatively large variance for RL models, we repeat each task 5 times and keep the best model on each task.
}
This has a real-world analogy where the bot first learns the basics of the task, and then learns to
improve its performance via a question-asking policy tailored to the user's patience
(represented by $cost_{AQ}$) and its own ability to asnwer questions.

\section{Experiments}
\subsection{Offline Results}
Offline results are presented in Tables \ref{HeyHeyOOVResult}, \ref{OfflineResult}, and \ref{TestModelAQ}.
Table \ref{OfflineResult} presents results for the {\it vanilla-MemN2N} and
{\it Forward Prediction} models. Table \ref{HeyHeyOOVResult} presents results for {\it Cont-MemN2N},
which is better at handling unknown words.
We repeat each experiment 10 times and report the best result. 
Finally, Table \ref{TestModelAQ} presents results for the test scenario
 where the bot itself chooses when to ask questions.
Observations can be summarized as as follows:

\begin{table*}[t!]
\centering
\scriptsize
\begin{tabular}{ccccccccc}\toprule
&\multicolumn{4}{c}{Question Clarification }&\multicolumn{4}{c}{Knowledge Operation} \\\cmidrule(r){2-5}\cmidrule(r){6-9}
&\multicolumn{2}{c}{Task 1: Q. Paraphrase}&\multicolumn{2}{c}{Task 2: Q. Verification}&\multicolumn{2}{c}{Task 3: Ask For Relevant K.}&\multicolumn{2}{c}{Task 4: K. Verification}  \\
\cmidrule(r){2-3}\cmidrule(r){4-5}\cmidrule(r){6-7}\cmidrule(r){8-9}
Train \textbackslash Test &TestQA&TestAQ &TestQA&TestAQ &TestQA&TestAQ &TestQA&TestAQ\\\midrule
TrainQA (Context) &0.754 &0.726 & 0.742&0.684&  0.883&0.947 &  0.888&0.959\\
TrainAQ (Context)&0.640&0.889&0.643&0.807&0.716&0.985 &0.852&0.987 \\
TrainMix (Context)&0.751&0.846&0.740&0.789&0.870&0.985 & 0.875&0.985 \\\bottomrule
\end{tabular}
\hspace{0.5cm}
\begin{tabular}{ccccccccccc}\toprule
&\multicolumn{8}{c}{Knowledge Acquisition }   \\\cmidrule(r){2-11}
&\multicolumn{2}{c}{Task 5: Q. Entity}&\multicolumn{2}{c}{Task 6: Answer Entity}   &\multicolumn{2}{c}{Task 7: Relation Entity}&\multicolumn{2}{c}{Task 8: Triple}&\multicolumn{2}{c}{Task 9: Everything}   \\
\cmidrule(r){2-3}\cmidrule(r){4-5}\cmidrule(r){6-7}\cmidrule(r){8-9}\cmidrule(r){10-11}
Train \textbackslash Test &TestQA&TestAQ &TestQA&TestAQ &TestQA&TestAQ &TestQA&TestAQ&TestQA&TestAQ \\\midrule
TrainQA (Context)&$<$0.01 &0.224&$<$0.01&0.120&0.241&0.301& 0.339 &0.251 & $<$0.01&0.058\\
TrainAQ (Context)& $<$0.01 &0.639& $<$0.01&0.885&0.143&0.893&0.154&0.884 &  $<$0.01&0.908\\
TrainMix (Context)& $<$0.01 &0.632&$<$0.01& 0.852 &0.216&0.898&0.298&0.886&  $<$0.01&0.903\\\bottomrule
\end{tabular}
\caption{Results for Cont-MemN2N on different tasks.}
\label{HeyHeyOOVResult}
\end{table*}

\begin{table*}[h]
\centering
\scriptsize
\begin{tabular}{ccccccccc}\toprule
&\multicolumn{4}{c}{Question Clarification }&\multicolumn{4}{c}{Knowledge Operation} \\\cmidrule(r){2-5}\cmidrule(r){6-9}
&\multicolumn{2}{c}{Task 1: Q. Paraphrase}&\multicolumn{2}{c}{Task 2: Q. Verification}&\multicolumn{2}{c}{Task 3: Ask For Relevant K.}&\multicolumn{2}{c}{Task 4: K. Verification}  \\\cmidrule(r){2-9}
Train \textbackslash Test &TestQA&TestAQ &TestQA&TestAQ &TestQA&TestAQ &TestQA&TestAQ\\\midrule
TrainQA &0.338 &0.284 &0.340&0.271&  0.462&0.344 &  0.482&0.322\\
TrainAQ&0.213&0.450&0.225&0.373&0.187&0.632  &0.283&0.540 \\
TrainAQ(+FP)&0.288&0.464&0.146&0.320&0.342&0.631&0.311&0.524 \\
TrainMix&0.326&0.373&0.329&0.326&0.442&0.558 & 0.476&0.491 \\\bottomrule
\end{tabular}
\centering
\scriptsize
\begin{tabular}{ccccccccccc}\toprule
&\multicolumn{8}{c}{Knowledge Acquisition }   \\\cmidrule(r){2-11}
&\multicolumn{2}{c}{Task 5:  Q. Entity}&\multicolumn{2}{c}{Task 6: Answer Entity}   &\multicolumn{2}{c}{Task 7: Relation Entity}&\multicolumn{2}{c}{Task 8: Triple}&\multicolumn{2}{c}{Task 9: Everything}   \\\cmidrule(r){2-3}\cmidrule(r){4-5}\cmidrule(r){6-7}\cmidrule(r){8-9}\cmidrule(r){10-11}
Train \textbackslash Test &TestQA&TestAQ &TestQA&TestAQ &TestQA&TestAQ &TestQA&TestAQ&TestQA&TestAQ \\\midrule
TrainQA (vanila)&$<0.01$ &0.223& $<$0.01&$<$0.01&0.109&0.129& 0.201&0.259 & $<$0.01&$<$0.01\\
TrainAQ (vanila)&$<0.01$ &0.660& $<$0.01&$<$0.01&0.082&0.156&0.124&0.664 & $<$0.01&$<$0.01\\
TrainAQ(+FP)&$<0.01$& 0.742&$<0.01$&$<0.01$& 0.085&0.188&0.064&0.702&$<$0.01&$<$0.01  \\
Mix (vanila)& $<$0.01 &0.630&$<$0.01& $<$0.01 &0.070&0.152&0.180&0.572& $<$0.01&$<$0.01\\\bottomrule
\end{tabular}
\caption{Results for offline settings using memory networks.}
\label{OfflineResult}
\end{table*}

\begin{table}[h!]
\centering
\small
\begin{tabular}{ccc}\toprule
& Question Clarification & Knowledge Acquisition \\
&Task 2: Q. Verification& Task 4: K. Verification\\
&TestModelAQ&TestModelAQ\\\midrule
TrainAQ&0.382 &0.480 \\
TrainAQ(+FP)&0.344&0.501 \\
TrainMix&0.352&0.469 \\\bottomrule
\end{tabular}
\caption{Results for TestModelAQ settings. }
\label{TestModelAQ}
\end{table}

- Asking questions helps at test time, which is intuitive since it provides additional evidence:
\begin{itemize}
\item  {\it TrainAQ+TestAQ} (questions can be asked at both training and test time) performs the best across all the settings.
\item {\it TrainQA+TestAQ} (questions can be asked at training time but not at test time) performs worse than {\it TrainQA+TestQA} (questions can be asked at neither training nor test time) in tasks {\it Question Clarification} and {\it Knowledge Operation} due to the discrepancy between training and testing.
\item   {\it TrainQA+TestAQ}  performs better than {\it TrainQA+TestQA}  on all {\it Knowledge Acquisition} tasks, the only exception being the {\it Cont-MemN2N} model on the {\it Missing Triple} setting. The explanation is that for most tasks in {\it Knowledge Acquisition}, the learner has no chance of giving the correct answer without asking questions. The benefit from asking is thus large enough to compensate for the negative effect introduced by data discrepancy between training and test time.
\item {\it TrainMix} offers flexibility in bridging the gap between datasets generated using QA and AQ, very slightly underperforming {\it TrainAQ+TestAQ}, but gives competitive results on both {\it TestQA} and {\it TestAQ} in  the {\it Question Clarification} and {\it Knowledge Operations} tasks.
\item {\it TrainAQ+TestQA} (allowing questions at training time but forbid questions at test time) performs the worst, even worse than
{\it TrainQA+TestQA}. This has a real-world analogy where a student becomes dependent on the teacher answering their questions, later struggling to answer the test questions without help.
\item In the {\it Missing Question Entity} task (the student does not know about the question entity), the {\it Missing Answer Entity} task (the student does not know about the answer entity), and {\it Missing Everything} task, the bot achieves accuracy less than 0.01  if
not asking questions at test time (i.e., {\it TestQA}).
\item The performance of {\it TestModelAQ}, where the bot relies on its model to ask questions at test time (and thus can ask irrelevant questions) performs similarly to asking the correct question at test time ({\it TestAQ}) and better than not asking questions ({\it TestQA}).
\end{itemize}
- {\it Cont-MemN2N} significantly outperforms {\it vanilla-MemN2N}. One explanation
is that considering context provides significant evidence distinguishing correct
answers from candidates in the dialogue history, 
especially in cases where the model encounters unfamiliar words.

\begin{figure*}[!ht]
\begin{center}
\includegraphics[width=2in]{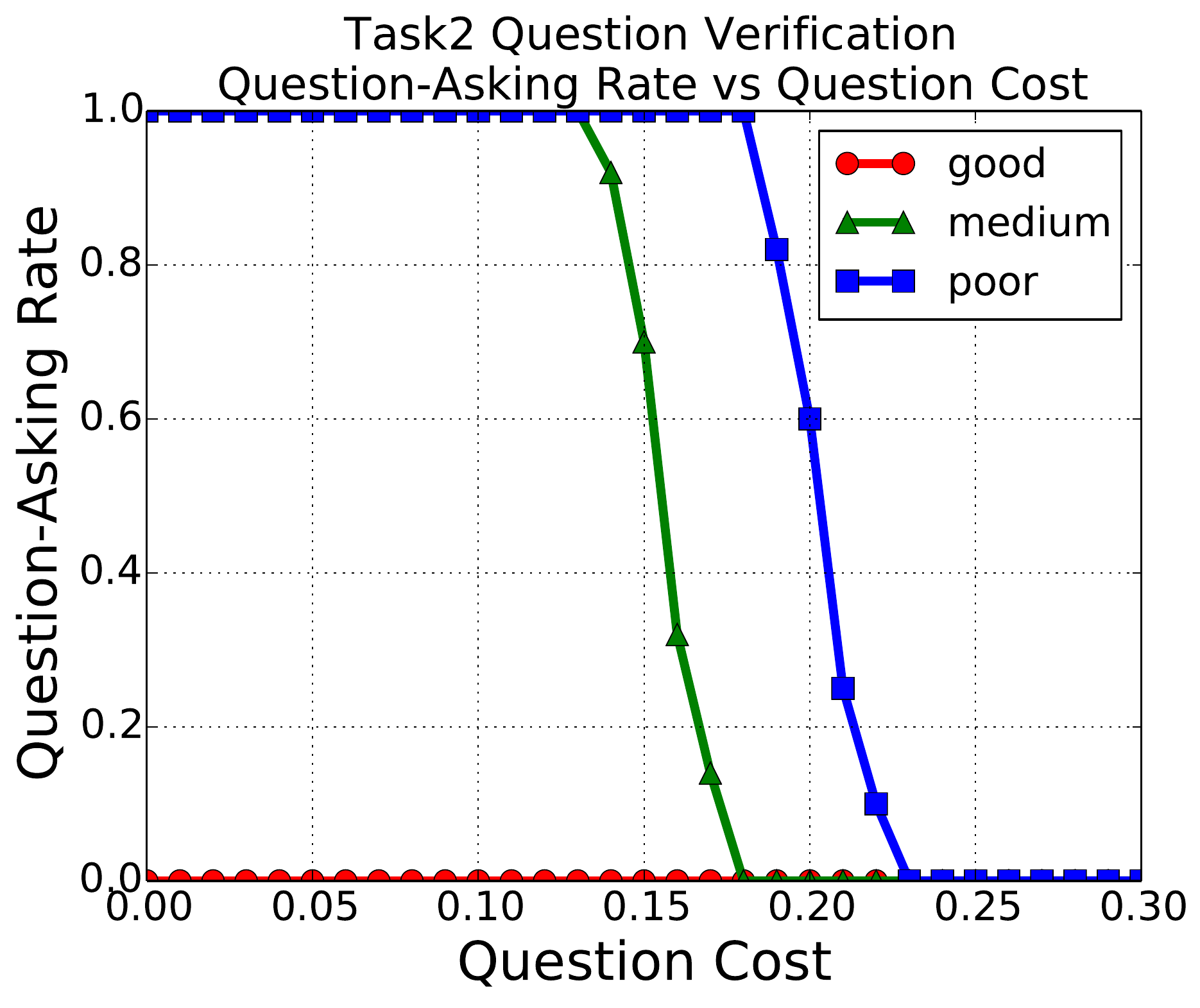}
\includegraphics[width=2in]{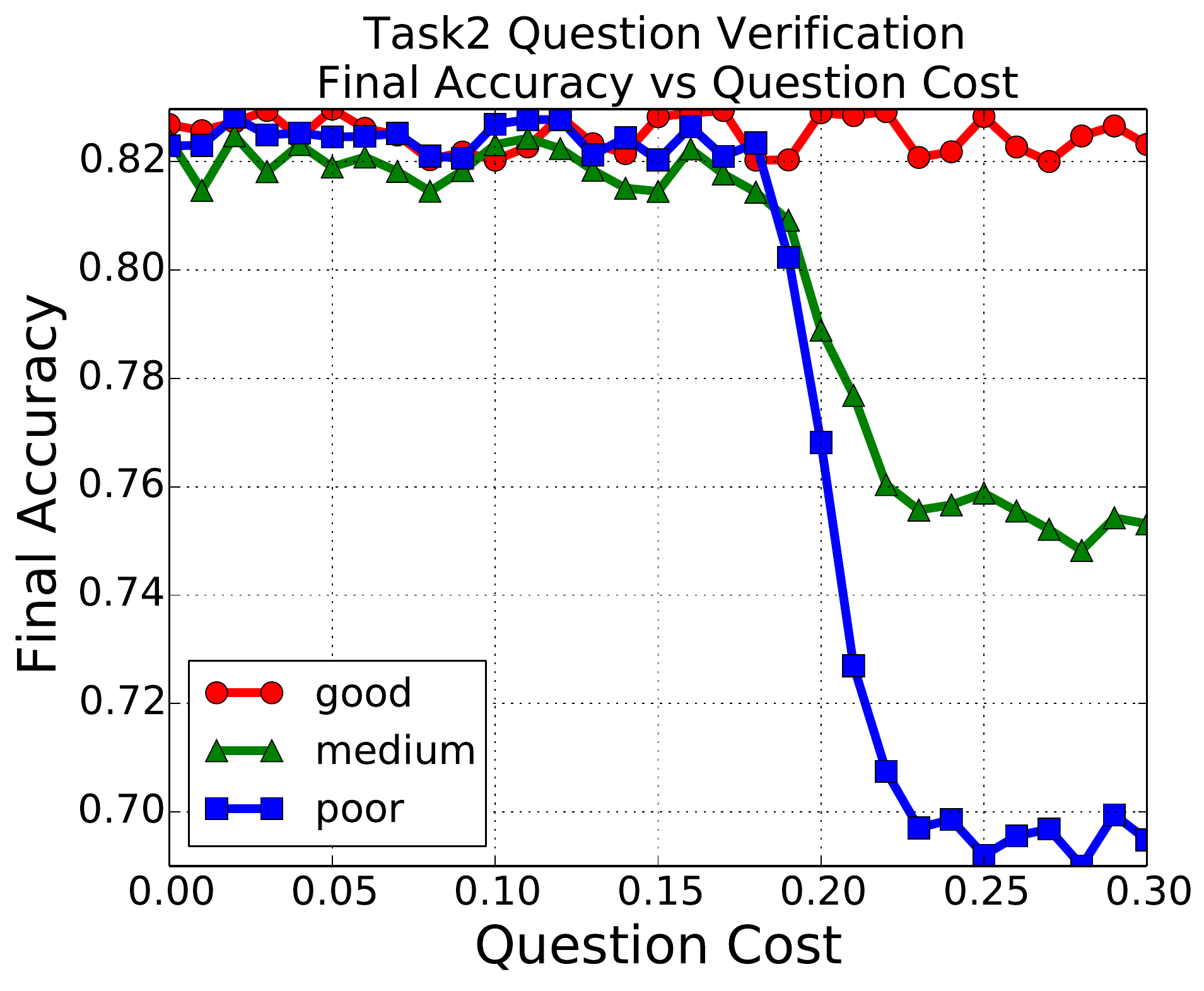}\\
\includegraphics[width=2in]{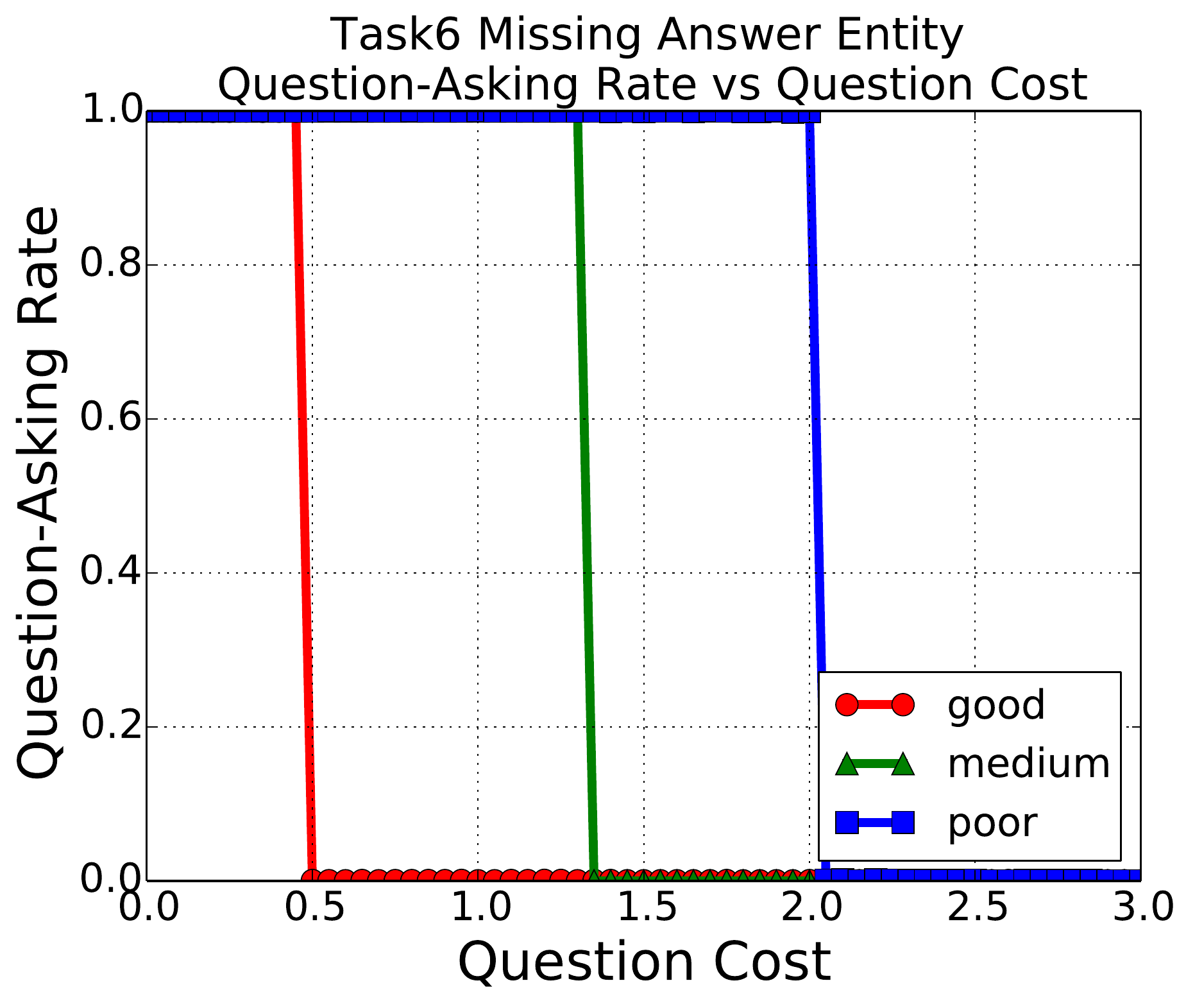}
\includegraphics[width=2in]{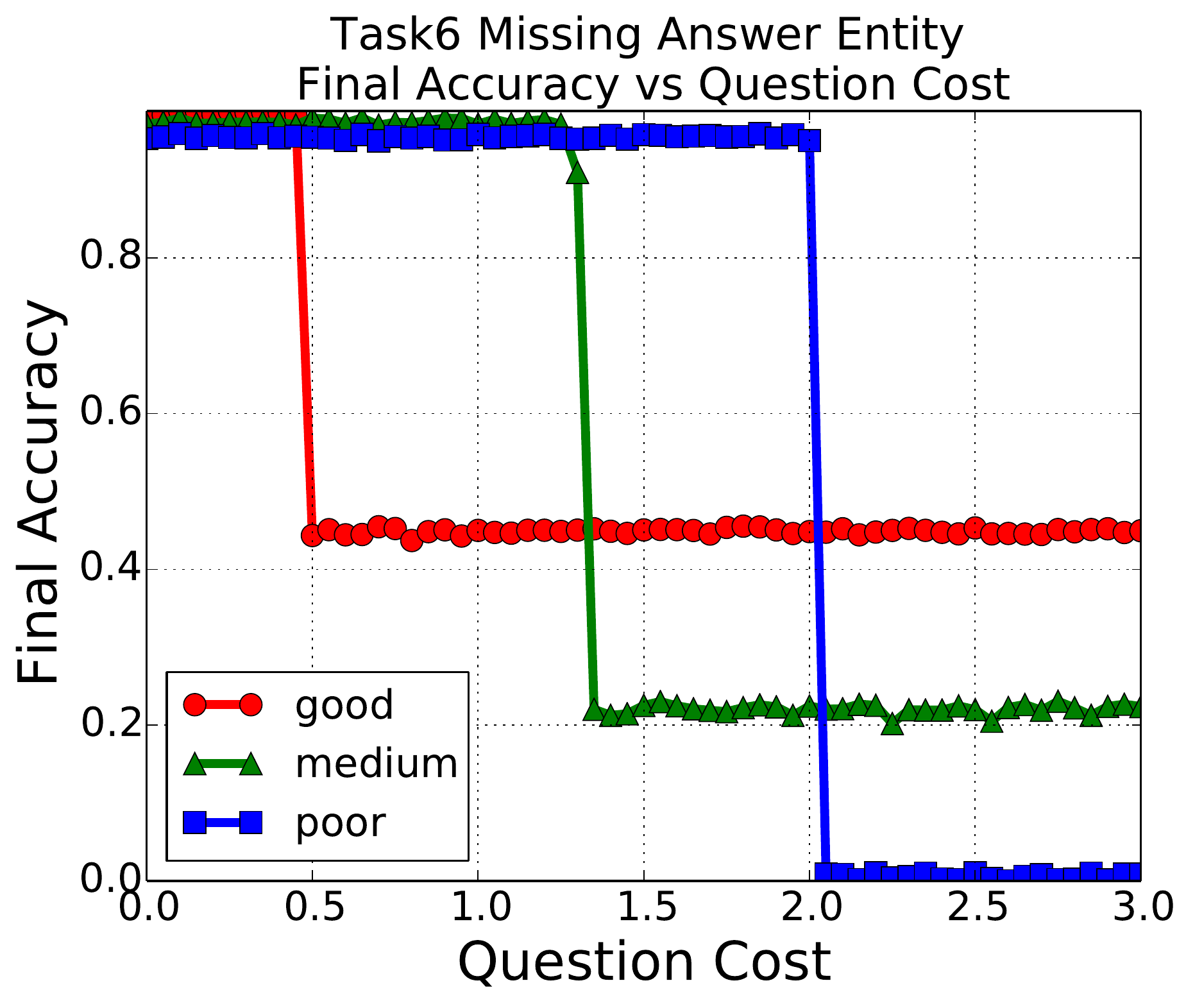}\\
\end{center}
\caption{Results of online learning for Task 2 and Task 6
\label{part-result}
}
\end{figure*}

\subsection{RL Results}
For the RL settings,
we  present  results for Task 2  ({\it Question Verification})
  and Task 6 ({\it Missing Answer Entities}) in Figure \ref{part-result}.
Task 2 represents scenarios where different types of student have different
abilities to correctly answer  questions (e.g., a poor student can still sometimes give
correct answers even when they do not fully understand the question).
Task 6 represents tasks where a poor learner who lacks the knowledge necessary
to answer the question can hardly give a correct answer.
All types of students including the good student will theoretically benefit from asking questions
(asking for the correct answer) in Task 6.
    We show the percentage of question-asking versus the cost of AQ on the test set
and  the accuracy of question-answering on the test set
vs the cost of AQ. Our main findings were: 
\begin{itemize}
\item
A good student does not need to ask questions
in Task 2  ({\it Question Verification}),
because they already understand the question. The student will raise questions
asking for the correct answer when cost is low for Task 6 ({\it Missing Answer Entities}).
\item A poor student always asks questions when the cost is low.
 As the cost increases, the frequency of question-asking declines.
\item As the AQ cost increases gradually, good students will stop asking questions
earlier than the medium and poor students. The explanation is intuitive:
poor students benefit more from asking questions than good students, so they
continue asking even with higher penalties.
\item As the probability of question-asking declines, the accuracy for poor and
medium students drops.  Good students are more resilient to not asking questions.
\end{itemize}

\subsection{Mechanical Turk Results} \label{sec:res-mturker}

\begin{table*}[t!]
\centering
\scriptsize
\begin{tabular}{ccccccccc}\toprule
&\multicolumn{4}{c}{vanilla-MemN2N}&\multicolumn{4}{c}{Cont-MemN2N} \\\cmidrule{2-5}\cmidrule{6-9}
&\multicolumn{2}{c}{Task 4: K. Verification}&\multicolumn{2}{c}{Task 8: Triple}
&\multicolumn{2}{c}{Task 4: K. Verification}&\multicolumn{2}{c}{Task 8: Triple}  \\\cmidrule{2-3}\cmidrule{4-5}\cmidrule{6-7}\cmidrule{8-9}
Train \textbackslash Test &TestQA&TestAQ &TestQA&TestAQ &TestQA&TestAQ &TestQA&TestAQ\\\midrule
TrainQA & 0.331 & 0.313  & 0.133 & 0.162 & 0.712  & 0.703  & 0.308 & 0.234\\
TrainAQ & 0.318 & 0.375  & 0.072 & 0.422 & 0.679  & 0.774  & 0.137 & 0.797 \\\bottomrule
\end{tabular}
\caption{Mechanical Turk Task Results. {\it Asking Questions} (AQ) outperforms only 
answering questions without asking (QA).
\label{res:mturk-mainy}}
\end{table*}

\begin{table*}[h]
\centering
\scriptsize
\begin{tabular}{ccccccccc}\toprule
&\multicolumn{4}{c}{vanilla-MemN2N}&\multicolumn{4}{c}{Cont-MemN2N} \\\cmidrule{2-5}\cmidrule{6-9}
&\multicolumn{2}{c}{Task 4: K. Verification}&\multicolumn{2}{c}{Task 8: Triple}
&\multicolumn{2}{c}{Task 4: K. Verification}&\multicolumn{2}{c}{Task 8: Triple}  \\\cmidrule{2-3}\cmidrule{4-5}\cmidrule{6-7}\cmidrule{8-9}
Train \textbackslash Test &TestQA&TestAQ &TestQA&TestAQ &TestQA&TestAQ &TestQA&TestAQ\\
TrainQA & 0.331 & 0.313  & 0.133 & 0.162 & 0.712  & 0.703  & 0.308 & 0.234\\
TrainAQ & 0.318 & 0.375  & 0.072 & 0.422 & 0.679  & 0.774  & 0.137 & 0.797 \\\bottomrule
\end{tabular}
\caption{Mechanical Turk Task Results, using real data for training and testing.
\label{res:mturk_main}}
\vspace{1cm}
\centering
\scriptsize
\begin{tabular}{ccccccccc}\toprule
&\multicolumn{4}{c}{vanilla-MemN2N}&\multicolumn{4}{c}{Cont-MemN2N} \\\cmidrule{2-5}\cmidrule{6-9}
&\multicolumn{2}{c}{Task 4: K. Verification}&\multicolumn{2}{c}{Task 8: Triple}
&\multicolumn{2}{c}{Task 4: K. Verification}&\multicolumn{2}{c}{Task 8: Triple}  \\\cmidrule{2-3}\cmidrule{4-5}\cmidrule{6-7}\cmidrule{8-9}
Train \textbackslash Test &TestQA&TestAQ &TestQA&TestAQ &TestQA&TestAQ &TestQA&TestAQ\\
TrainQA & 0.356 & 0.311  & 0.128 & 0.174 & 0.733  & 0.717  & 0.368 & 0.352\\
TrainAQ & 0.340 & 0.445  & 0.150 & 0.487 & 0.704  & 0.792  & 0.251 & 0.825 \\\bottomrule
\end{tabular}
\caption{Results on Mechanical Turk Tasks using a combination of real and simulated data for training,
testing on real data.
\label{res:mturk_main2}}
\vspace{1cm}
\centering
\scriptsize
\begin{tabular}{ccccccccc}\toprule
&\multicolumn{4}{c}{vanilla-MemN2N}&\multicolumn{4}{c}{Cont-MemN2N} \\\cmidrule{2-5}\cmidrule{6-9}
&\multicolumn{2}{c}{Task 4: K. Verification}&\multicolumn{2}{c}{Task 8: Triple}
&\multicolumn{2}{c}{Task 4: K. Verification}&\multicolumn{2}{c}{Task 8: Triple}  \\\cmidrule{2-3}\cmidrule{4-5}\cmidrule{6-7}\cmidrule{8-9}
Train \textbackslash Test &TestQA&TestAQ &TestQA&TestAQ &TestQA&TestAQ &TestQA&TestAQ\\
TrainQA & 0.340 & 0.311  & 0.120 & 0.165 & 0.665  & 0.648  & 0.349 & 0.342\\
TrainAQ & 0.326 & 0.390  & 0.067 & 0.405 & 0.642  & 0.714  & 0.197 & 0.788 \\\bottomrule
\end{tabular}
\caption{Results on Mechanical Turk Tasks using only simulated data for training,
but testing on real data.
\label{res:mturk_main3}}
\end{table*}

Results for the Mechanical Turk Tasks are given in Table \ref{res:mturk-mainy}.
We again compare vanilla-MemN2N and Cont-MemN2N, using the same
 TrainAQ/TrainQA and TestAQ/TestQA combinations as before, 
for Tasks 4 and 8 as described in  Section \ref{sec:mturk}.
We tune hyperparameters on the validation set and 
repeat each experiment 10 times and report the best result. 

While performance is lower than on the related Task 4 and Task 8 simulator tasks,
we still arrive at the same trends and conclusions when real data from humans is used.
The performance was expected to be lower because (i) real data has more lexical variety, complexity,
 and noise; and (ii) the training set was smaller due to data collection costs (10k vs. 180k).

More importantly, the same main conclusion is observed as before:
TrainAQ+TestAQ (questions can be asked at both training and test time) 
performs the best across all the settings. That is, we show that a bot asking questions to 
humans learns to outperform one that only answers them.

We also provide additional experiments.
In Table \ref{res:mturk-mainy}, results were shown when training and testing on the collected
Mechanical Turk data (around 10,000 episodes of training dialogues for training).
As we collected the data in the same settings as Task 4 and 8
of our simulator, we could also consider supplementing training with simulated 
data as well, of which we have a larger amount (over 100,000 episodes).
Note this is {\em only for training}, we will 
still test on the real (Mechanical Turk collected) data.
Although the simulated data has less lexical variety as it is built from templates, the 
larger size might obtain improve results. 

Results are given in Table \ref{res:mturk_main2} when training on the combination of real and simulator data,
and testing on real data. This should be compared to 
training on only the real data (Table \ref{res:mturk_main}) and only on the simulator
data (Table \ref{res:mturk_main3}).
The best results are obtained from the combination of simulator and real data. The best 
real data only results (selecting over algorithm and training strategy) on both tasks outperform
the best results using simulator data, i.e. 
using Cont-MemN2N with the Train AQ / TestAQ setting) 0.774 and 0.797 is obtained
 vs. 0.714 and 0.788 for Tasks 4 and 8 respectively.
This is despite there being far fewer examples of real data compared to simulator data.
Overall we obtain two main conclusions from this additional experiment:
(i) real data is indeed measurably superior to simulated data for training our models;
(ii) in all cases (across different algorithms, tasks, and data types -- be they real data, simulated data 
or combinations)
 the bot asking questions (AQ) outperforms it only answering questions and not asking them (QA).
The latter reinforces the main result of the paper.

\chapter{Dialogue Learning with Human-in-the-Loop}
\label{HITL}
In this chapter, we will discuss how a bot can improve in an online fashion from humans' feedback. 
A good conversational agent 
should have the ability to learn from the online feedback from a teacher: adapting its model when making mistakes and reinforcing the model when the teacher's feedback is positive.
This is particularly important in the situation where the bot is initially trained in a supervised way on a fixed synthetic, domain-specific or pre-built dataset before release, but will be exposed to
a different environment after release
 (e.g.,
 more diverse natural language utterance usage when talking with real humans, different distributions, special cases, etc.).  Most recent research
has focused on training a bot
from fixed training sets of labeled data but seldom on how the bot can
improve through online interaction with humans.
Human (rather than machine) language learning happens during communication \citep{bassiri2011interactional,werts1995instructive},
and not from labeled datasets, hence making this an important subject to study.

In this chapter, we explore this direction by
 training a bot through interaction with teachers in an online fashion.
 The task is formalized under the general framework of reinforcement learning
via the teacher's (dialogue partner's) feedback to the dialogue actions from the bot.
 The dialogue takes place in the context of question-answering tasks and the bot has to, given either a short story or a set of facts, answer a set of questions from the teacher.
We consider two types of feedback: explicit numerical rewards as in conventional
reinforcement learning, and textual feedback which is more natural in human dialogue.
We consider two online training scenarios:
(i) where the task is built with a dialogue simulator allowing for easy analysis and repeatability
of experiments; and (ii) where the teachers are real humans using Amazon Mechanical Turk.

    We explore  important issues involved in online learning
   such as  how a bot can be most efficiently trained using a minimal amount of teacher's feedback,
 how a bot can harness different types of feedback signal,
how to avoid pitfalls such as instability during online learing with different types of feedback via
 data balancing and exploration,
and how to make learning with real humans feasible via data batching.
Our findings indicate that
it is feasible to build a pipeline
that starts from a model trained with fixed data and then learns from interactions with humans
to improve itself. 

\section{Dataset and Tasks}

We begin by describing the data setup we use.
In our first set of experiments we build a simulator as a testbed for learning algorithms.
In our second set of experiments we use Mechanical Turk to provide real human teachers giving feedback.


\subsection{Simulator}

The simulator adapts two existing fixed datasets to our online setting.
Following \newcite{weston2016dialog}, we use
(i) the single supporting fact problem from the bAbI datasets \cite{weston2015towards}
which consists of 1000 short stories from a simulated world interspersed with questions; and (ii)
the WikiMovies dataset \cite{weston2015towards} which consists
 of roughly 100k (templated) questions
over 75k entities based on questions with answers in the open movie database (OMDb).
Each dialogue takes place between a teacher, scripted by the simulation, and a bot.
The communication protocol is as follows: (1)
the teacher first asks a question from the fixed set of questions existing in the dataset, (2)
the bot answers the question, and finally (3) the teacher gives feedback on the bot's answer.

\subsubsection{Tasks}
 We follow the paradigm defined in \newcite{weston2016dialog} where the teacher's feedback
 takes the form of either textual feedback, a numerical reward, or both, depending on the task.
For each dataset, there are ten tasks:
\begin{tightitemize}
\item {\bf   Task 1}: The teacher tells the student exactly what they should have said (supervised baseline).
\item {\bf  Task 2}:
The teacher replies with positive textual feedback and reward, or negative textual feedback. 
\item {\bf Task 3}: The teacher gives textual feedback containing the answer when the bot is wrong.
\item {\bf Task 4}: The teacher provides a hint by providing the class of the correct answer, e.g., ``No it's a movie" for the question ``which movie did Forest Gump star in?".
\item {\bf Task 5}: The teacher provides
a reason why the student's answer is wrong by pointing out the relevant supporting fact from the knowledge base.
\item {\bf Task 6}: The teacher gives positive reward only 50\% of the time. 
\item {\bf Task 7}: Rewards are missing and the teacher only gives natural language feedback.
\item {\bf Task 8}: Combines Tasks 1 and 2 to see whether a learner can learn successfully from both forms of supervision at once.
\item {\bf Task 9}:  The bot asks questions of the teacher about what it has done wrong.
\item {\bf Task 10}:  The bot will receive a hint rather than the correct answer after asking for help.
\end{tightitemize}
\begin{figure*}
\centering
\footnotesize
\begin{tabular}{|l|l|l|}
\cline{1-1}\cline{3-3}
 \\[-2ex]
{\color{blue}T : Which  {\color{blue} movie} did Tom Hanks {\color{blue} star} in ?} &&{\color{blue}T : Which  {\color{blue}movie} did Tom Hanks  {\color{blue}star} in ?}  \\
{\color{red}S : Forrest Gump }  &&  {\color{red}S :  Brad Pitt.}\\
\cline{1-1}\cline{3-3}
 \multicolumn{1}{c}{}
\vspace{-2mm} \\
\cline{1-1}\cline{3-3}
Task 1: Imitating an Expert Student &&Task 1: Imitating an Expert Student  \\
{\color{blue}S: Forrest Gump} && {\color{blue}S: Forrest Gump}\\
{\color{blue}T: (no response)  } &&{\color{blue}T: (no response)  }  \\
\cline{1-1}\cline{3-3}
 \multicolumn{1}{c}{}
\vspace{-2mm} \\
\cline{1-1}\cline{3-3}
Task 2: Positive and Negative Feedback &&Task 2: Positive and Negative Feedback  \\
{\color{blue}T: Yes, that's right! (+)} && {\color{blue}T: No, that's incorrect! \MINUS} \\
\cline{1-1}\cline{3-3}
 \multicolumn{1}{c}{}
\vspace{-2mm} \\
\cline{1-1}\cline{3-3}
Task 3: Answers Supplied by Teacher &&Task 3: Answers Supplied by Teacher  \\
{\color{blue}T: Yes, that is correct. (+)} && {\color{blue}T: No, the answer is Forrest Gump ! \MINUS} \\
\cline{1-1}\cline{3-3}
 \multicolumn{1}{c}{}
\vspace{-2mm} \\
\cline{1-1}\cline{3-3}
Task 4: Hints Supplied by Teacher &&Task 4: Hints Supplied by Teacher  \\
{\color{blue}T: Correct! (+)} && {\color{blue}T: No, it's a movie ! \MINUS} \\
\cline{1-1}\cline{3-3}
 \multicolumn{1}{c}{}
\vspace{-2mm} \\
\cline{1-1}\cline{3-3}
Task 5: Supporting Facts Supplied by Teacher &&Task 5: Supporting Facts Supplied by Teacher  \\
{\color{blue}T: That's right. (+)} && {\color{blue}T: No, because Forrest Gump starred actors } \\
&&{\color{blue} Tom Hanks, Robin Wright, Gary Sinise ! \MINUS} \\
\cline{1-1}\cline{3-3}
 \multicolumn{1}{c}{}
\vspace{-2mm} \\
\cline{1-1}\cline{3-3}
Task 6: Partial Feedback && Task 6: Partial Feedback \\
if random(0,1)$<$0.5 then  && {\color{blue}T: Sorry, wrong. \MINUS} \\
~~~~{\color{blue}T: That's correct. (+)} && \\
else ~{\color{blue}T: That's correct.  } && \\
\cline{1-1}\cline{3-3}
 \multicolumn{1}{c}{}
\vspace{-2mm} \\
\cline{1-1}\cline{3-3}
Task 7: No Feedback  && Task 7: No Feedback \\
{\color{blue}T: Yes.} && {\color{blue}T: No.} \\
\cline{1-1}\cline{3-3}
 \multicolumn{1}{c}{}
\vspace{-2mm} \\
\cline{1-1}\cline{3-3}
Task 8: Imitation and Feedback Mixture &&Task 8: Imitation and Feedback Mixture \\
if random(0,1)$<$0.5 then  && if random(0,1)$<$0.5 then  \\
~~~~{\color{blue}T: Yes, that's right! (+)} &&~~~~ {\color{blue}T: Wrong. \MINUS} \\
else ~{\color{blue}T: (no response)  } && else ~{\color{blue} S: Forrest Gump  } \\
\cline{1-1}\cline{3-3}
 \multicolumn{1}{c}{}
\vspace{-2mm} \\
\cline{1-1}\cline{3-3}
Task 9: Asking For Corrections &&Task 9: Asking For Corrections  \\
{\color{blue}T: Correct! (+)}  &&  {\color{blue}T: No, that's wrong. } \\
&& {\color{red}S: Can you help me? } \\
&& {\color{blue}T: Forrest Gump !} \\
\cline{1-1}\cline{3-3}
 \multicolumn{1}{c}{}
\vspace{-2mm} \\
\cline{1-1}\cline{3-3}
Task 10: Asking For Supporting Facts && Task 10: Asking For Supporting Facts\\
{\color{blue}T: Yes, that's right! (+)}  &&  {\color{blue}T: Sorry, that's not it.} \\
&& {\color{red}S: Can you help me? } \\
&& {\color{blue}T: A relevant fact is that
Forrest Gump starred actors } \\
&&{\color{blue} Tom Hanks, Robin Wright, Gary Sinise ! \MINUS} \\\hline

\end{tabular}
\caption[The ten tasks the simulator implements.]{The ten tasks our simulator implements, which evaluate different
forms of teacher response and binary feedback.
In each case the same example from WikiMovies is given for simplicity,
where the student answered correctly for all tasks (left) or incorrectly (right).
 {\color{red}Red} text denotes responses by the bot with  {\color{red}S} denoting the bot.
{\color{blue}Blue} text is spoken by the teacher with {\color{blue}T} denoting the teacher's response. For imitation learning the teacher provides the response the student should say denoted with {\color{blue}S} in Tasks 1 and 8.
A {(\color{blue}{+})} denotes a positive reward.
}
\label{Tasks}
\end{figure*}

Here, we only consider 
 Task 6 (``partial feedback''): 
the teacher replies with positive textual feedback (6 possible templates) when the bot answers correctly, and positive reward is given only 50\% of the time.
When the bot is wrong, the teacher gives textual feedback containing the answer.
Example dialogues are given in Figure \ref{fig:simulator-examples}.

\definecolor{dred}{rgb}{0.7,0.0,0.0}
\newcommand{\PLUS}{{\textcolor{blue}{(+)}}}
\newcommand{\SPACE}{~~~~~~~~~~~~~~~~~~~~~~}
\begin{figure*}[h]
\begin{small}
\begin{tabular}{|l|c|l|}
\cline{1-1}\cline{3-3}
&& \\[-2ex]
{\bf bAbI Task 6: Partial Rewards} &&
{\bf WikiMovies Task 6: Partial Rewards}  \\
&& \\[-2ex]
Mary went to the hallway.         &&
What films are about Hawaii? \SPACE \textcolor{dred}{50 First Dates}\\
John moved to the bathroom.    &&
Correct! \\  
Mary travelled to the kitchen.     &&
Who acted in Licence to Kill? \SPACE \textcolor{dred}{Billy Madison}\\
Where is Mary? \SPACE~ \textcolor{dred}{kitchen} &&
No, the answer is Timothy Dalton.\\
Yes, that's right! &&
What genre is Saratoga Trunk in?  \SPACE    \textcolor{dred}{Drama}\\
Where is John? \SPACE~~ \textcolor{dred}{bathroom}  &&
Yes! \PLUS \\
Yes, that's correct!  \PLUS   &&  ~~~~~~\dots  \\
\cline{1-1}\cline{3-3}
\end{tabular}
\end{small}
\caption[Simulator sample dialogues for the bAbI task and WikiMovies.]{Simulator sample dialogues for the bAbI task (left) and WikiMovies (right).
We consider 10 different tasks following \cite{weston2016dialog}.
The teacher's dialogue is in black and the bot is in red.
$\PLUS$ indicates receiving positive reward, given only 50\% of the time even when correct.
\label{fig:simulator-examples}
}
\end{figure*}

\begin{figure*}[!ht]
\center
\begin{small}
\begin{tabular}{|l|}
\hline
{\bf Sample dialogues with correct answers from the bot:}\\
Who wrote the Linguini Incident ?     \SPACE~~~~~~~~~~~~~~~~~~~ \textcolor{dred}{richard shepard}\\
Richard Shepard is one of the right answers here.\\
What year did The World Before Her premiere? \SPACE \textcolor{dred}{2012}\\
Yep! That's when it came out.  \\
Which are the movie genres of Mystery of the 13th Guest?  ~~~ \textcolor{dred}{crime}\\
\vspace{1mm}
Right, it can also be categorized as a mystery.\\
{\bf Sample dialogues with incorrect answers from the bot:}\\
What are some movies about a supermarket ?    \SPACE~~~ \textcolor{dred}{supermarket} \\
There were many options and this one was not among them.  \\
Which are the genres of the film Juwanna Mann ? ~~~~~~~~~~~~~~~~~~~ \textcolor{dred}{kevin pollak} \\
That is incorrect. Remember the question asked for a genre not name. \\
Who wrote the story of movie Coraline ?    \SPACE~~~~~~~~~~~~\textcolor{dred}{fantasy} \\
That's a movie genre and not the name of the writer. A better answer would of been Henry Selick\\
 or Neil Gaiman.\\
\hline
\end{tabular}
\end{small}
\caption[Human Dialogue from Mechanical Turk]{Human Dialogue from Mechanical Turk (based on WikiMovies)
The human teacher's dialogue is in black and the bot is in red.
We show examples where the bot answers correctly (left) and incorrectly (right).
Real humans provide more variability of language in both questions and textual feedback
than in the simulator setup (cf. Figure \ref{fig:simulator-examples}).
 \label{fig:mturk_data}}
\end{figure*}

\begin{figure}[!ht]
\noindent\rule{16cm}{0.4pt}
Title: Write brief responses to given dialogue exchanges (about 15 min)\\

Description: Write a brief response to a student's answer to a teacher's question, providing feedback to the student on their answer.

Instructions:\\

Each task consists of the following triplets:
\begin{enumerate}
\item a question by the teacher

\item the correct answer(s) to the question (separated by ``OR'')

\item a proposed answer in reply to the question from the student

\end{enumerate}

Consider the scenario where you are the teacher and have already asked the question, and received the reply from the student. Please compose a brief response giving feedback to the student about their answer. The correct answers are provided so that you know whether the student was correct or not.

For example, given 1) question: ``what is a color in the united states flag?''; 2) correct answer: ``white, blue, red''; 3) student reply: ``red'', your response could be something like ``that's right!''; for 3) reply: ``green'', you might say ``no that's not right'' or ``nope, a correct answer is actually white''.

Please vary responses and try to minimize spelling mistakes. If the same responses are copied/pasted or overused, we'll reject the HIT.

Avoid naming the student or addressing ``the class'' directly.

We will consider bonuses for higher quality responses during review. \\
\noindent\rule{16cm}{0.4pt}

\caption{Instructions Given to Turkers}
\label{insTurk}
\end{figure}

\subsection{Mechanical Turk Experiments} \label{sec:data-mturk}

We also extended WikiMovies using Mechanical
Turk so that real human teachers are  giving feedback rather than using a simulation.
As both the questions and feedback are templated in the simulation, they are now both
replaced with natural human utterances. Rather than having a set of simulated tasks, we have
only one task, and we gave instructions to the teachers that they could give feedback
as they see fit. The exact instructions given to the Turkers is given in Figure \ref{insTurk}.
In general, each independent response  contains feedback like
(i) positive or negative sentences; or (ii) a phrase containing the answer
or (iii) a hint,  which are similar to setups defined in the simulator. However,
 some human responses cannot be so easily categorized,
and the lexical variability is much larger in human responses.
Such a process also reflects the basic idea of curriculum learning \cite{bengio2009curriculum}, in which 
the model follows a specific curriculum, in which the model first learns from easy examples from the simulator and then 
from real dialogue interactions, which tend to be harder.
Some examples of the collected data are given in Figure~\ref{fig:mturk_data}.

\section{Methods}
\subsection{Reinforcement Learning} \label{sec:rl} 
In this section, we present the algorithms we used to train MemN2N in an online fashion.
Our learning setup can be cast as a particular form of Reinforcement Learning. 
The policy is implemented by the MemN2N model. 
The state is the dialogue history. 
The action space
corresponds to the set of answers the MemN2N has to choose from to answer the teacher's
 question. In our setting, the policy chooses only one action for each episode.
The reward is either $1$ (a reward from the teacher when the bot answers correctly) 
 or $0$ otherwise.
Note that in our experiments, a reward equal to $0$ might mean that the answer is incorrect or
that the positive reward is simply missing.
The overall setup is closest to standard contextual bandits, except that the reward is binary.

When working with real human dialogues, e.g., collecting data via Mechanical Turk,
it is easier to set up a task whereby a bot is deployed to respond to a large batch of utterances,
as opposed to a single one.
The latter would be more difficult to manage and scale up since it would require some
form of synchronization between the model replicas interacting with each human.

This is comparable to the real world situation where a teacher can either ask a student a single question and give feedback right away,
or set up a test that contains many questions and grade all of them at once. Only after the learner completes all questions, it can hear
feedback from the teacher.

We use {\it batch size} to refer to how many dialogue
episodes the current model is used to collect feedback before updating its parameters.
In the Reinforcement Learning literature, batch size is related to {\em off-policy}
learning since the MemN2N policy is trained
using episodes collected with a stale version of the model. Our experiments show
that our model and base algorithms are very robust to the choice
of batch size, alleviating the need for correction terms in the learning algorithm~\citep{bottou-13}.

We consider two strategies: (i) online batch size, whereby the target policy is
updated after doing a single pass over each batch (a batch size of 1 reverts to
the usual on-policy online learning); and (ii) dataset-sized batch, whereby
training is continued to convergence on the batch which is the size of the dataset,
and then the target policy is updated with the new model, and a new batch is drawn and the procedure iterates.
These strategies can be applied to all the methods we use, described below.
%

Next, we discuss the learning algorithms we considered in this work.

\subsubsection{Reward-Based Imitation (RBI)}
The simplest algorithm we first consider is the one employed in \newcite{weston2016dialog}.
RBI relies on positive rewards provided by the teacher.
It is trained to imitate the correct behavior of the learner, i.e.,
learning to predict the correct answers (with reward 1) at training time and disregarding the other ones.
This is implemented by using a {MemN2N} that maps a dialogue input to a prediction, i.e. 
using the cross entropy criterion on the positively rewarded subset of the data.

In order to make this work in the online setting which requires exploration to find the correct answer,
we employ an $\epsilon$-greedy strategy:
the learner makes a prediction using its own model (the answer assigned the highest probability)
 with probability $1-\epsilon$, otherwise it picks a random answer with probability $\epsilon$.
The teacher will then give a reward of $+1$ if the answer is correct, otherwise $0$.
The bot will then learn to imitate the correct answers:
predicting the correct answers while ignoring the incorrect ones.

\subsubsection{REINFORCE}
The second algorithm we use is the REINFORCE algorithm \citep{williams1992simple},
which maximizes the expected cumulative reward of the episode, in our case the expected reward provided by the teacher.
The expectation is approximated by sampling an answer from the model distribution.
Let $a$ denote the answer that the learner gives,
$p(a)$ denote the probability that current model assigns to $a$,
$r$ denote the teacher's reward, and $J(\theta)$ denote the expectation of the reward. We have:
\begin{equation}
\nabla J(\theta)\approx\nabla\log p(a) [r-b]
\end{equation}
where $b$ is the baseline value, which is estimated using a linear regression model
that takes as input the output of the memory network after the last hop,
and outputs a scalar $b$ denoting the estimation of the future reward.
The baseline model is trained by minimizing the mean squared loss between the estimated reward $b$ and actual reward $r$, $||r-b||^2$.
We refer the readers to \newcite{zaremba2015reinforcement} for more details.
The baseline estimator model is independent from the policy model, and its error is not backpropagated through the policy model.

The major difference between RBI and REINFORCE is that (i) the learner only tries to imitate correct behavior in RBI while in REINFORCE it also leverages the incorrect behavior,
and (ii) the learner explores using an $\epsilon$-greedy strategy in RBI while in REINFORCE it uses the distribution over actions produced by the model itself.

\subsubsection{Forward Prediction (FP)}
FP \citep{weston2016dialog} handles the situation where a numerical reward for a bot's answer is not available, meaning that there are no +1 or 0 labels available after a student's utterance.
Instead, the model assumes
 the teacher gives  textual feedback $t$ to the bot's answer, taking the form of a dialogue utterance,
and the model tries to predict this instead.
Suppose that $x$ denotes the teacher's question and $C$=$c_1$, $c_2$, ..., $c_N$ denotes the dialogue history as before.
In {\it FP}, the model first maps the teacher's initial question $x$ and dialogue history $C$ to
a vector representation $u$ using a memory network with multiple hops.
Then the model will perform another hop of attention over all possible student's answers  in $\mathbb{A}$, 
with an additional part that incorporates the information
of which candidate (i.e., $a$) was actually selected in the dialogue:
\begin{equation}
p_{\hat{a}}=\softmax(u^T y_{\hat{a}})~~~~~o=\sum_{\hat{a}\in \mathbb{A}} p_{\hat{a}} (y_{\hat{a}}+\beta\cdot {\bf 1}[\hat{a}=a] )
\end{equation}
where $y_{\hat{a}}$ denotes the vector representation for the student's  answer candidate $\hat{a}$.
$\beta$ is a (learned) d-dimensional vector to signify the actual action $a$ that the student chooses.
$o$ is then combined with $u$ to predict the teacher's feedback $t$ using a softmax:
\begin{equation}
u_1=o+u ~~~~
t=\softmax (u_1^T x_{r_1}, u_1^T x_{r_2}, ..., u_1^T x_{r_N})
\end{equation}
where $x_{r_{i}}$ denotes the embedding for the $i^{th}$ response.
In the online setting, the teacher will give textual feedback,
and the learner needs to update its model using the feedback.
It was shown in \newcite{weston2016dialog} that in an off-line setting this procedure can work either on its own, or in conjunction with a method that uses numerical rewards as well for improved performance.
In the online setting,  we consider two simple extensions:
\begin{itemize}
\item $\epsilon$-greedy exploration:  with probability $\epsilon$
the student will give a random answer, and with probability $1-\epsilon$ it
will give the answer that its model assigns the largest probability. This method enables the model to explore the space of actions and to potentially discover correct answers.
\item data balancing: cluster the set of teacher responses $t$ and then balance training across the clusters equally.\footnote{In the simulated data, because the responses are templates, this can be implemented by first randomly sampling the response, and then randomly sampling a story with that response; we keep the history of all stories seen from which we sample. For real data slightly more sophisticated clustering should be used.}
This is a type of experience replay \citep{mnih2013playing} but sampling with an evened distribution.
Balancing stops part of the distribution dominating the learning.
For example, if the  model is not exposed to sufficient positive and negative feedback,
and one class overly dominates, the learning process degenerates
to a model that always predicts the same output regardless of its input.

\end{itemize}

\section{Experiments}
Experiments are first conducted using our simulator, and then
using Amazon Mechanical Turk with real human subjects taking the role of the teacher.\footnote{
Code and data are available at
\tiny{\url{https://github.com/facebook/MemNN/tree/master/HITL}}.
}
\subsection{Simulator}
\paragraph{Online Experiments} \label{sec:online_exp}

\begin{figure}
\includegraphics[width=2.5in]{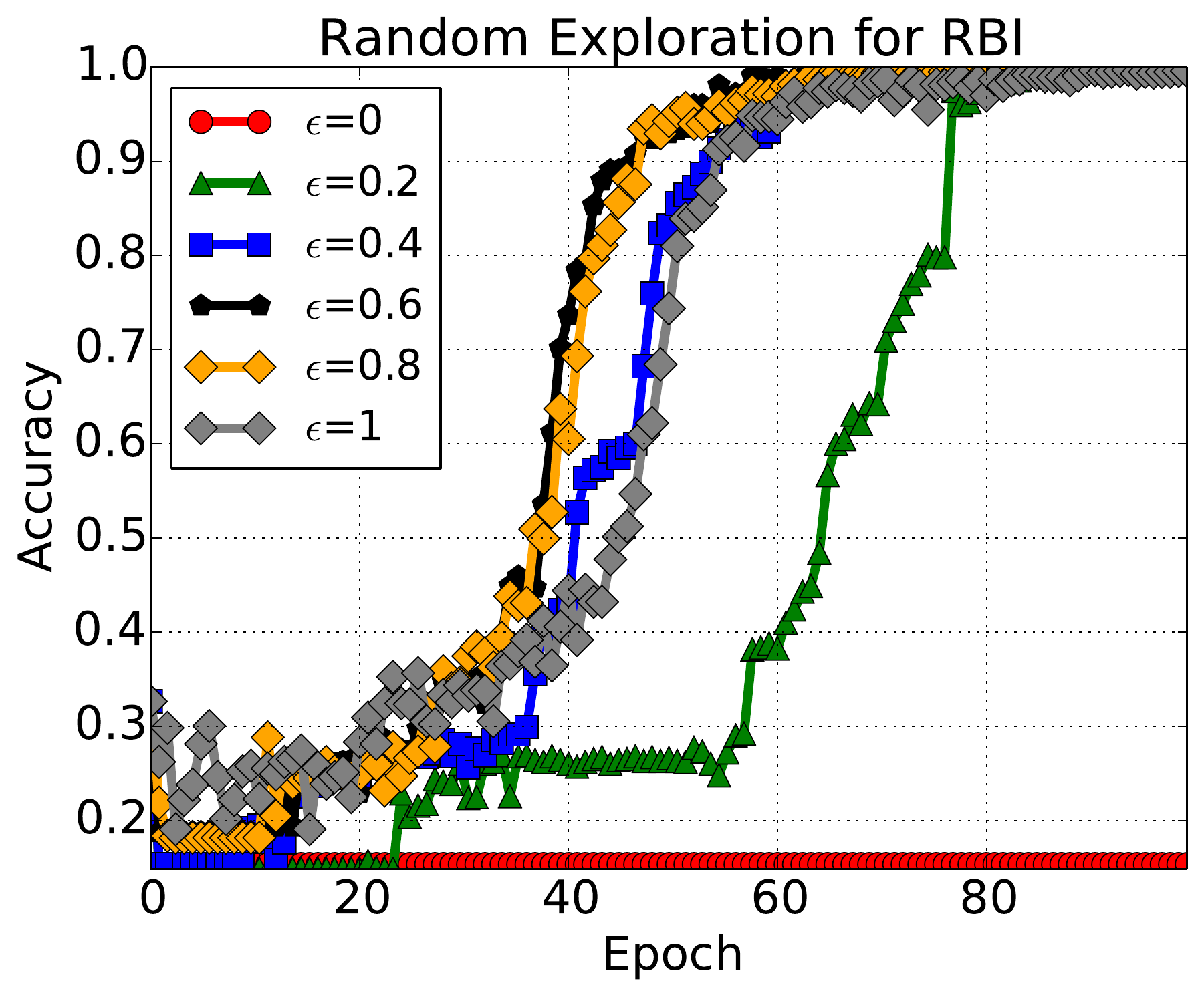}
\includegraphics[width=2.5in]{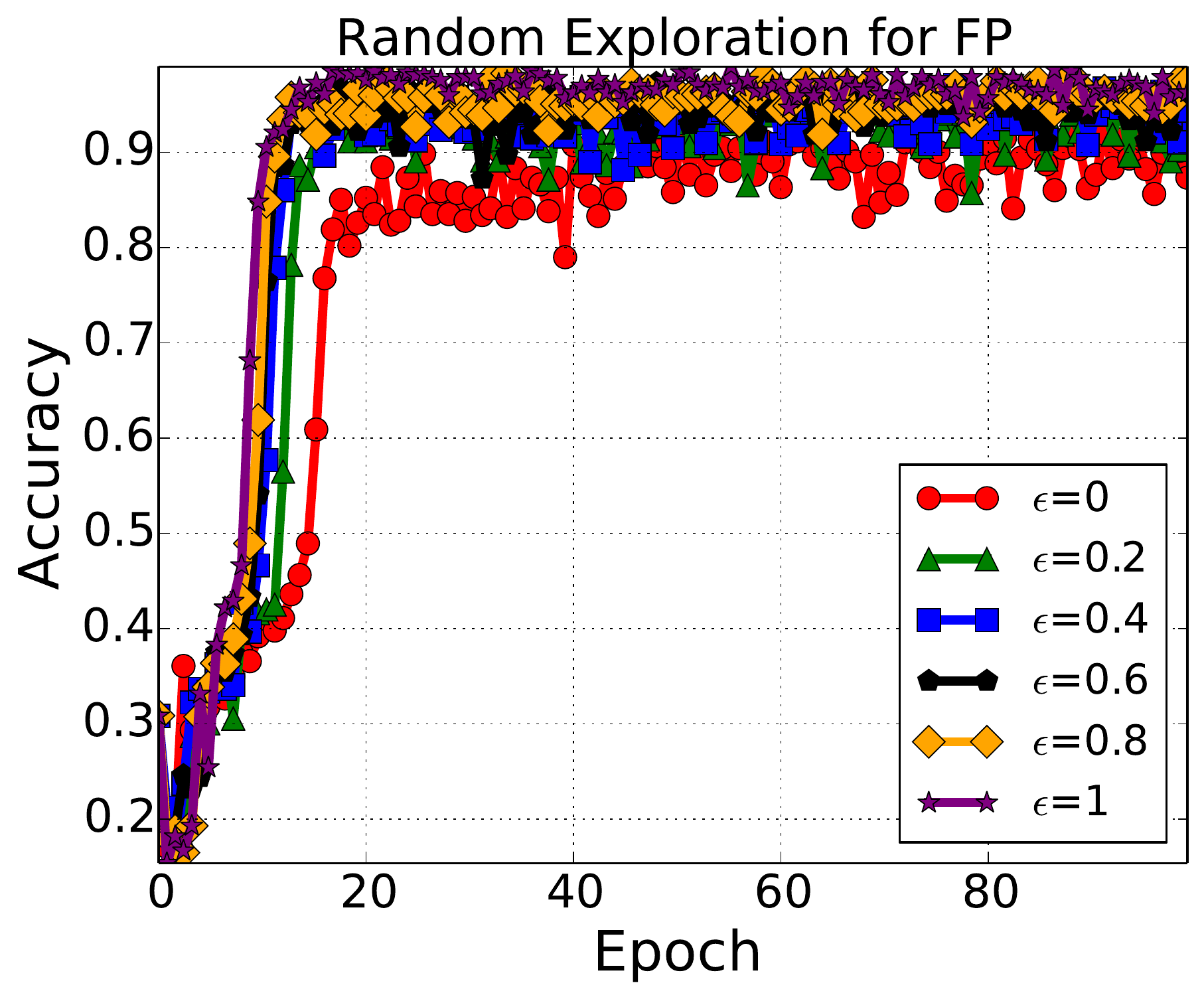}\\
\includegraphics[width=2.5in]{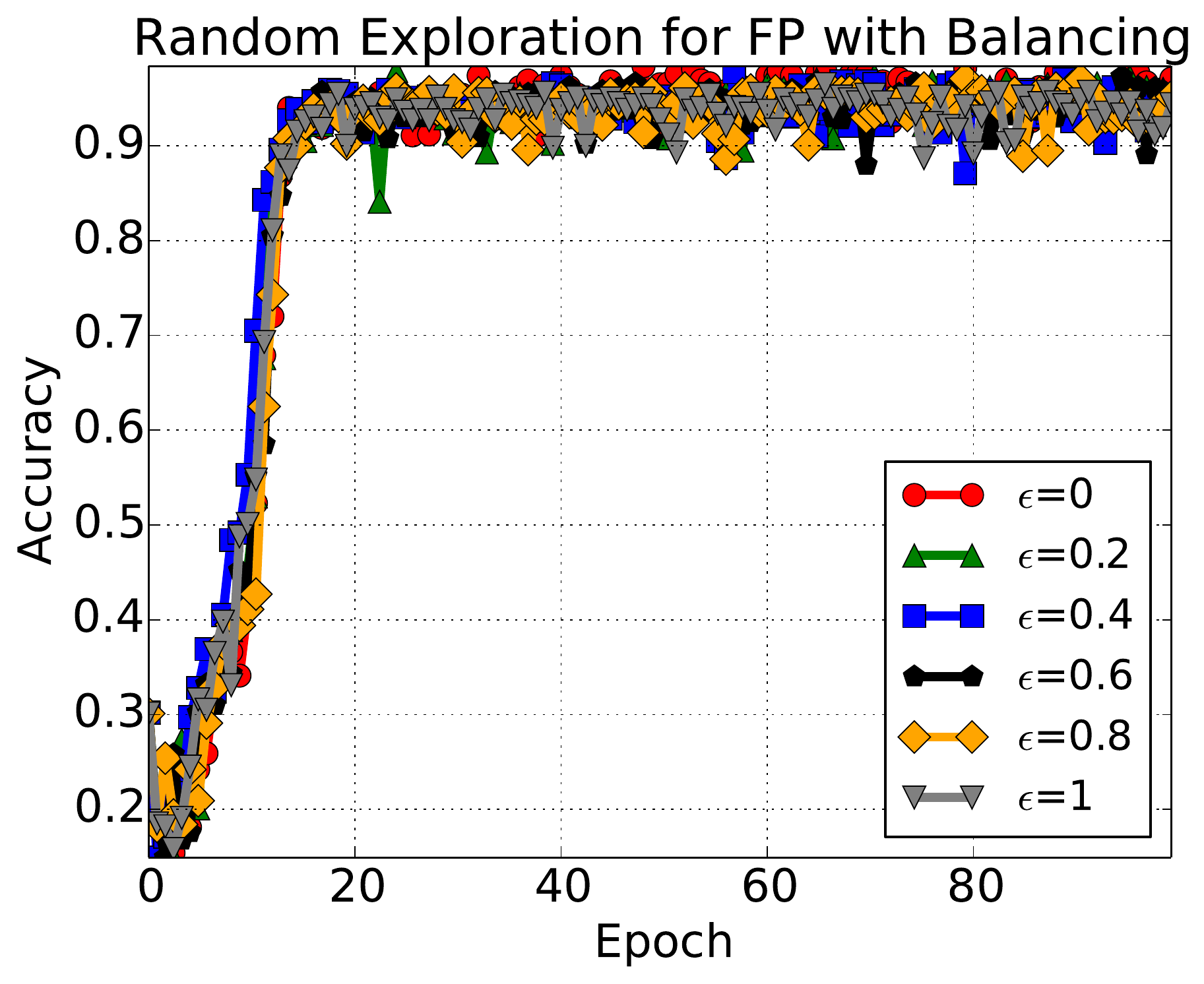}
\includegraphics[width=2.5in]{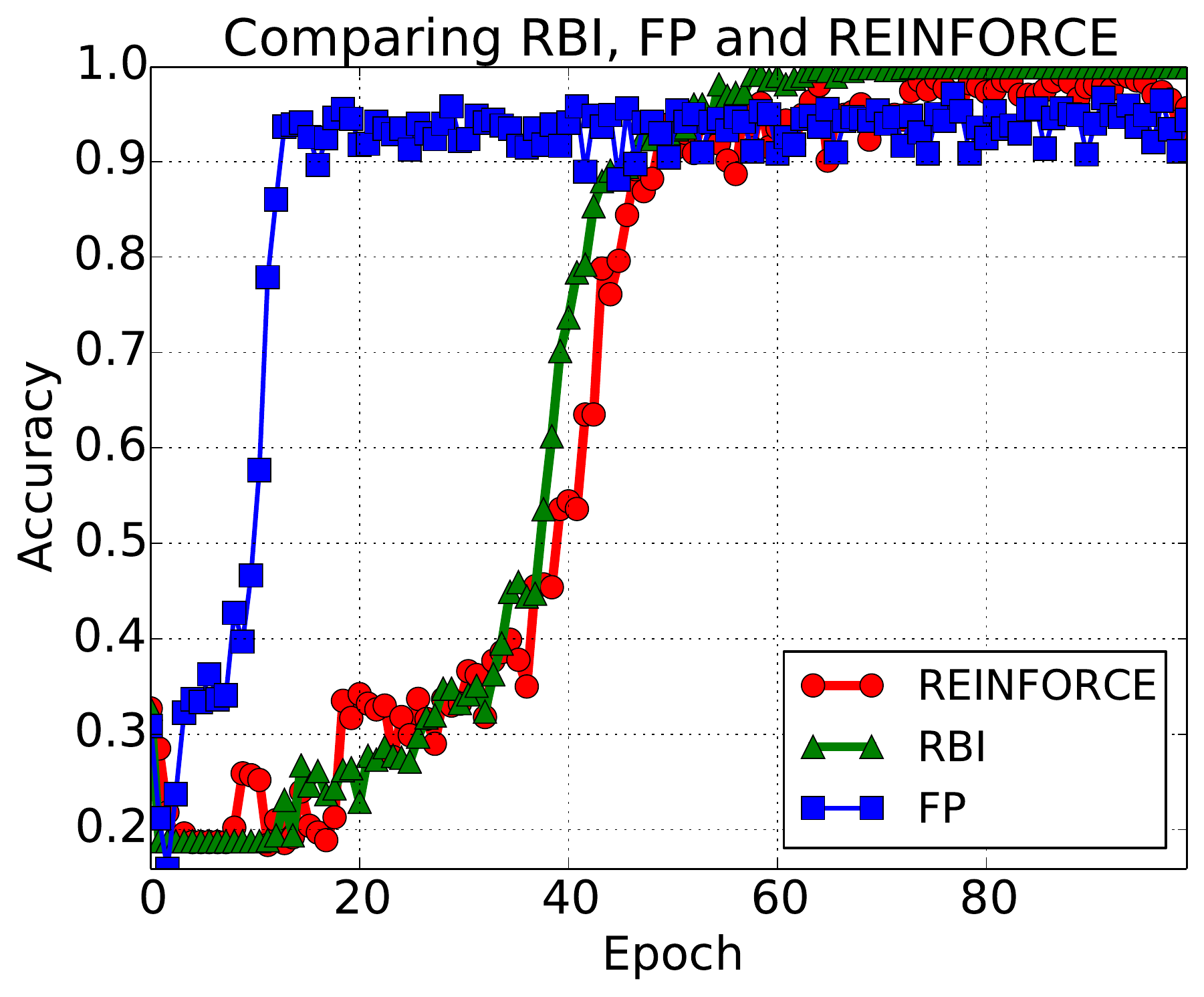}\\
\includegraphics[width=2.5in]{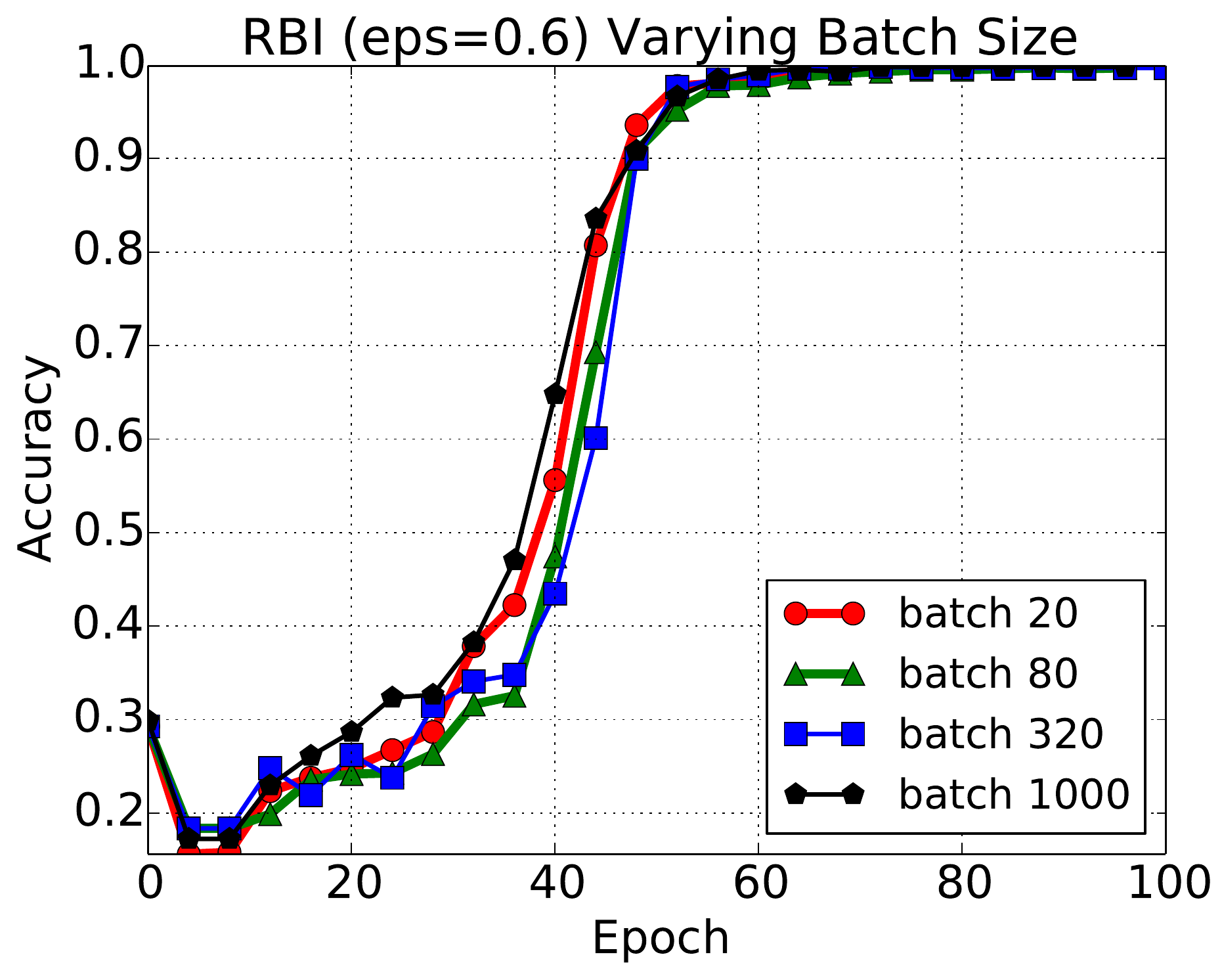}
\includegraphics[width=2.5in]{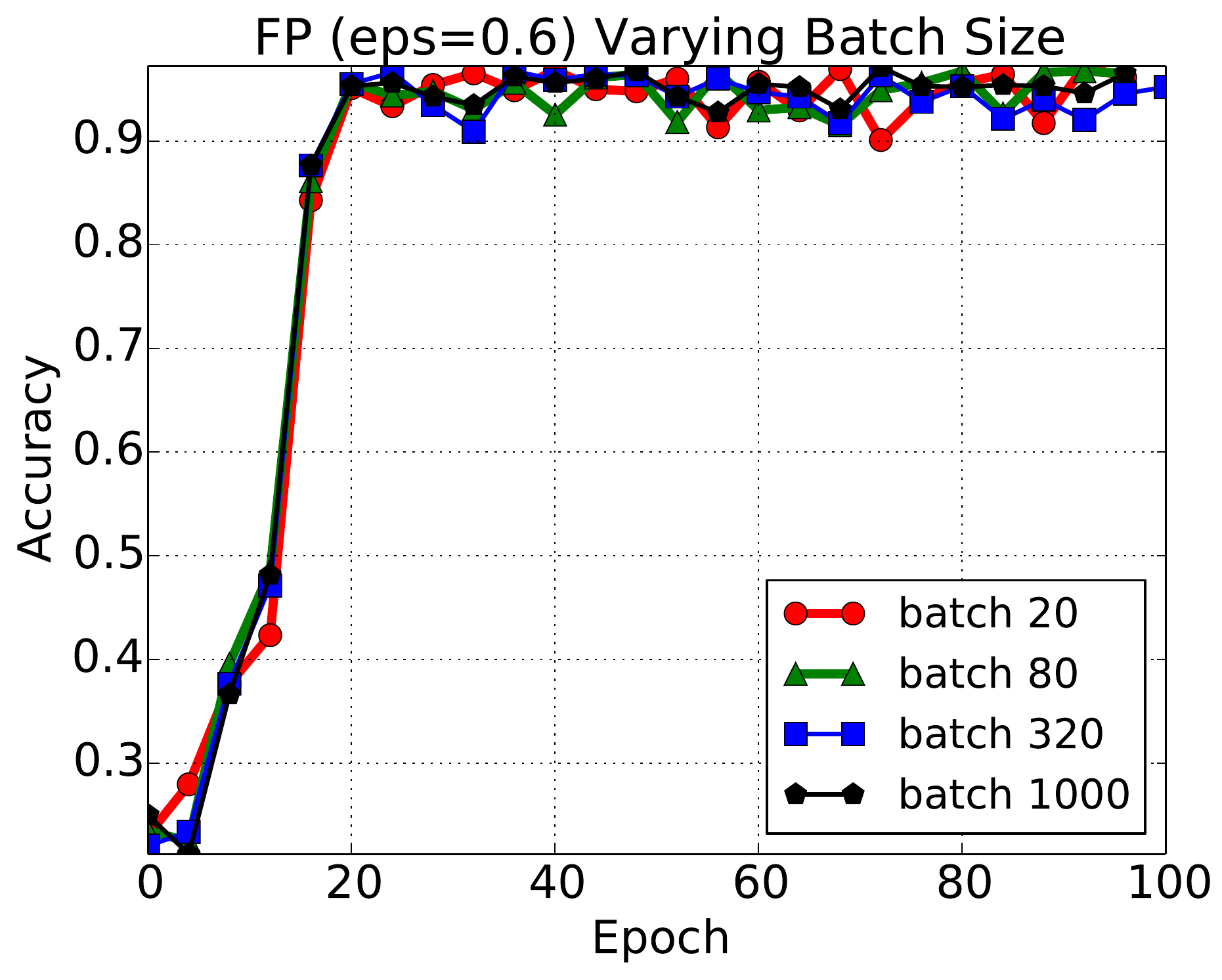}\\
\caption[Training epoch vs.\ test accuracy for bAbI (Task 6) varying exploration $\epsilon$ and batch size.]{  Training epoch vs.\ test accuracy for bAbI (Task 6) varying exploration $\epsilon$ and batch size.
Random exploration is important for both reward-based (RBI) and forward prediction (FP).
Performance is largely independent of batch size, and
RBI performs similarly to REINFORCE. 
Note that supervised, rather than reinforcement learning, with gold standard labels
achieves 100\% accuracy on this task.
\label{fig:online-babi-task6}
}
\end{figure}

\begin{figure*}
\center
\includegraphics[width=2.35in]{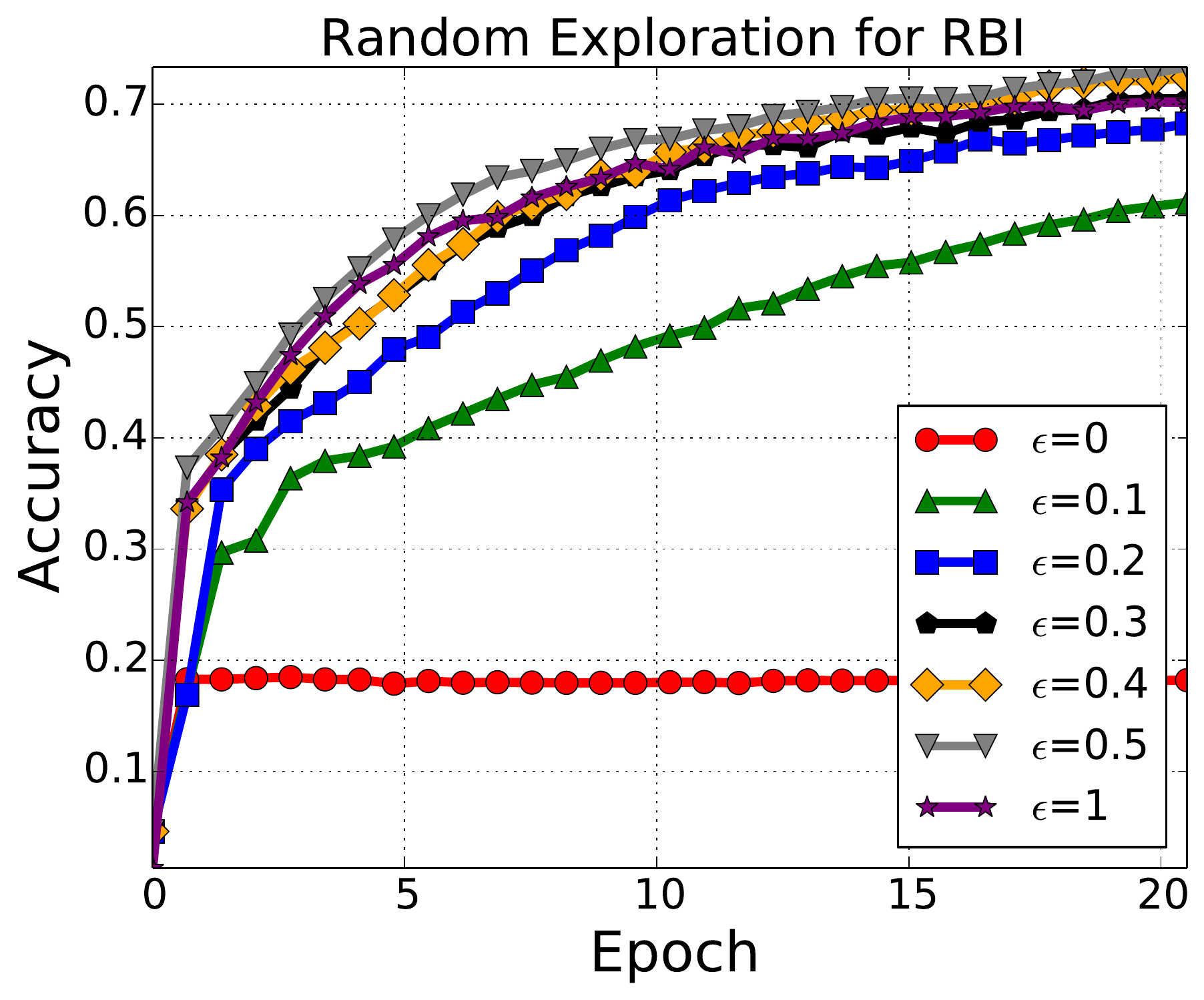}
\includegraphics[width=2.35in]{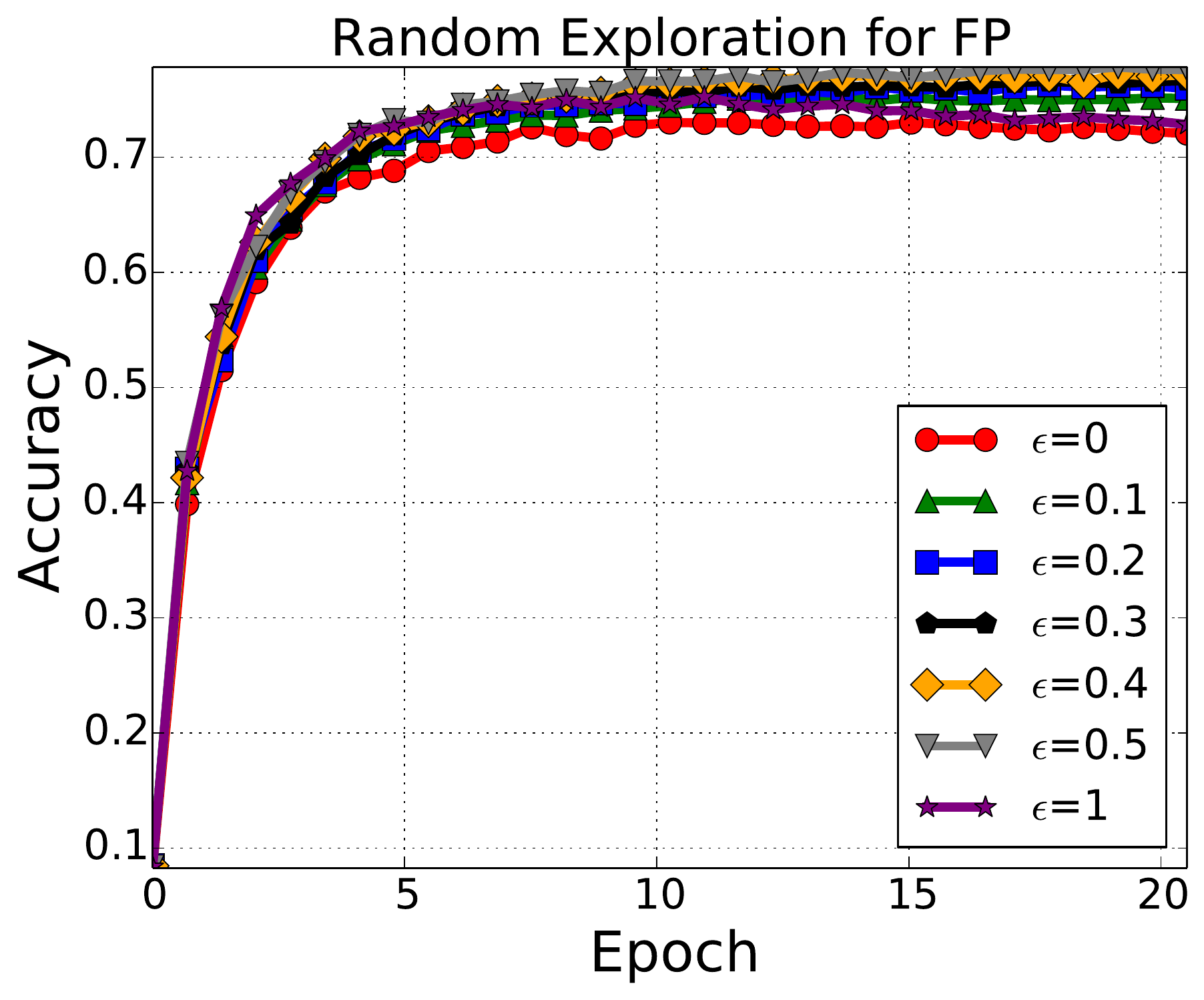}\\
\includegraphics[width=2.35in]{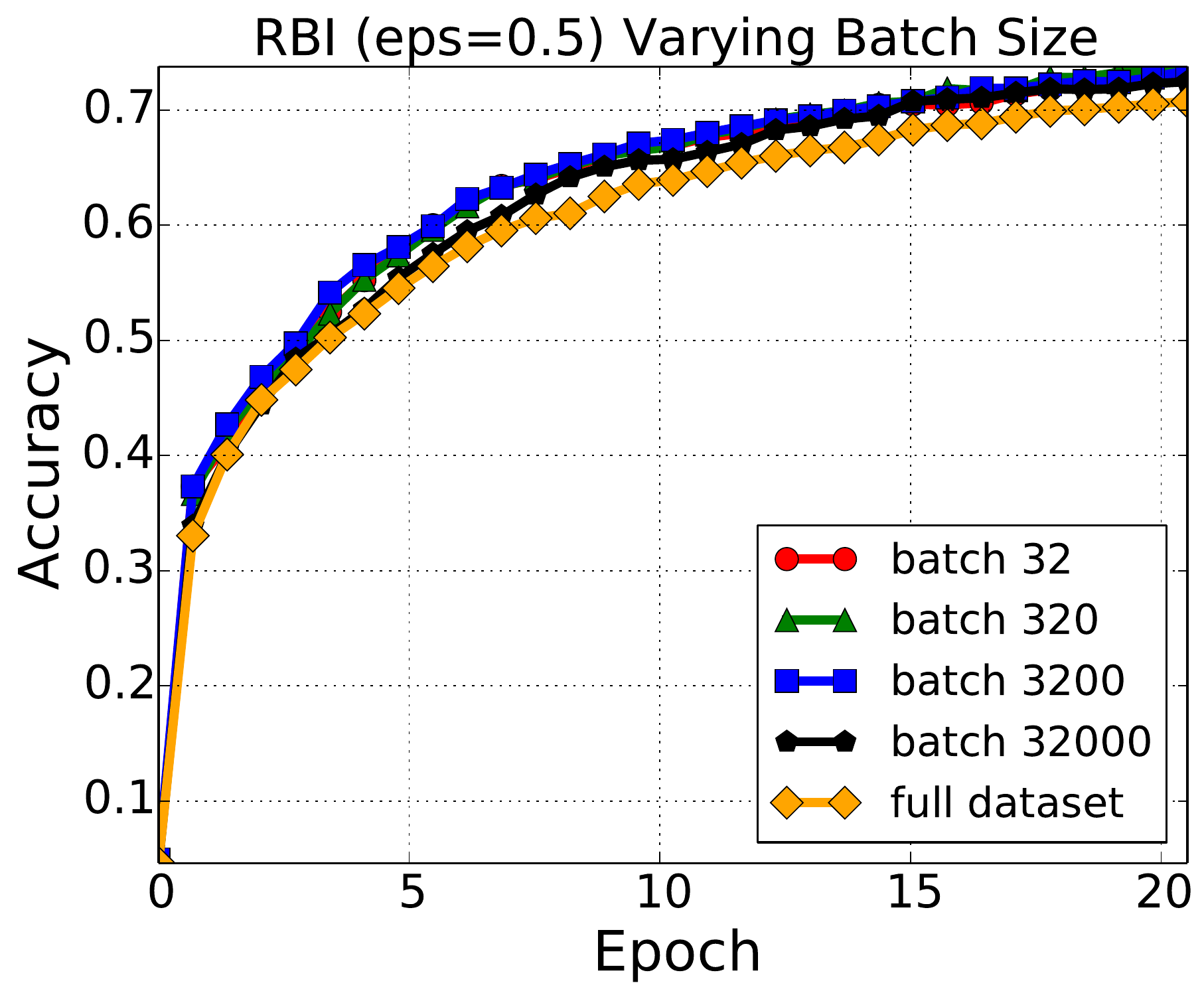}
\includegraphics[width=2.35in]{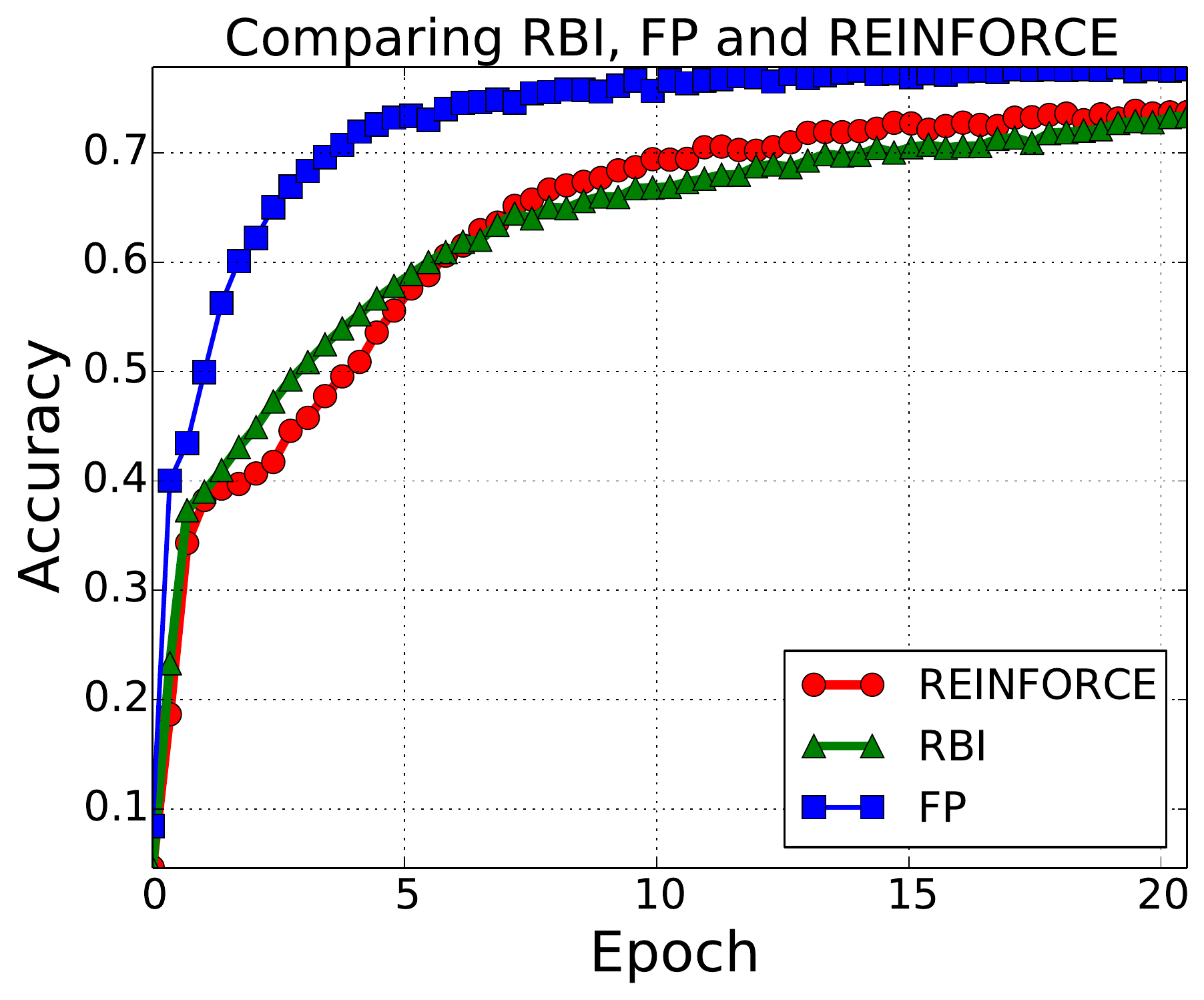}\\
\caption[Training epoch vs.\ test accuracy on Task $6$ varying exploration rate $\epsilon$]{ WikiMovies: Training epoch vs.\ test accuracy on Task $6$ varying (top left panel) exploration rate $\epsilon$ while setting batch size to $32$ for RBI,
(top right panel) for FP,
(bottom left) batch size for RBI, and (bottom right) comparing RBI, REINFORCE, and FP
  with $\epsilon=0.5$. 
The model is robust to the choice of batch size. RBI and REINFORCE perform comparably.
Note that supervised, rather than reinforcement learning, with gold standard labels
achieves 80\% accuracy on this task \citep{weston2016dialog}.
\label{fig:online-movieqa-task6}
}
\end{figure*}

In our first experiments, we considered both the bAbI and WikiMovies tasks
and varied batch size, random exploration rate $\epsilon$, and type of model.
Figure~\ref{fig:online-babi-task6} and Figure~\ref{fig:online-movieqa-task6}
shows (Task 6) results on bAbI and WikiMovies.

\newpage

Overall, we obtain the following conclusions:
\begin{itemize}
\item In general RBI and FP do work in a reinforcement learning setting, but can perform better with random exploration.
\item In particular RBI can fail without exploration. RBI needs random noise for exploring labels otherwise it can get stuck predicting a subset of labels and fail.
\item REINFORCE does not perform competitively. 
It obtains similar performance to RBI with optimal $\epsilon$. 
\item FP with balancing or with exploration via $\epsilon$ both outperform FP alone.
\item For both RBI and FP, performance is largely independent of online batch size.
\end{itemize}

\paragraph{Dataset Batch Size Experiments}
Given that larger online batch sizes appear to work well, and that this could be important in
a real-world data collection setup where the same model is deployed to gather a large amount
 of feedback from humans,
we conducted further experiments where the batch size is exactly equal to the dataset size
 and for each batch training is completed to convergence.
After the model has been trained on the dataset,
it is deployed to collect a new dataset of questions and answers, and the process is repeated.
Table~\ref{table:dataset-batch-babi} reports test error at each iteration of training,
using the bAbI Task $6$ as the case study.
The following conclusions can be made for this setting:
\begin{itemize}
\item RBI improves in performance as we iterate. Unlike in the online case,
RBI does not need random exploration.
We believe this is because the first batch, which is collected with a randomly initialized model,
contains enough variety of examples with positive rewards
 that the model does not get stuck predicting a subset of labels.
\item FP is not stable in this setting. 
This is because once the model gets very good
at making predictions (at the third iteration), it is not exposed to a sufficient number of
negative responses anymore.
From that point on, learning degenerates and performance drops as the model always predicts the same responses.
At the next iteration, it will recover again since it has a more balanced training set,
but then it will collapse again in an oscillating behavior.
\item FP does work if extended with balancing or random exploration with sufficiently large $\epsilon$.
\item RBI+FP also works well and helps with the instability of FP, alleviating the need for random exploration and data balancing.
\end{itemize}

Overall, our simulation
 results indicate that while a bot can be effectively trained fully online from bot-teacher interactions,
collecting real dialogue data in batches (which is easier to collect and iterate experiments over) is also a viable approach. We hence pursue the latter approach in our next set of experiments.


\begin{table*}[!tbh]
\center
\begin{tabular}{lccccccc}\toprule
Iteration  &     1     & 2     & 3     & 4     & 5     & 6 \\
\midrule
Imitation Learning               & 0.24  & 0.23  & 0.23 & 0.22 & 0.23 & 0.23 \\
Reward Based Imitation (RBI)     & 0.74  & 0.87  & 0.90 & {\bf 0.96} & {\bf 0.96} & {\bf 0.98}  \\
Forward Pred. (FP)               & {\bf 0.99}  & {\bf 0.96}  & {\bf 1.00} & 0.30 & {\bf 1.00} & 0.29  \\
RBI+FP                           & {\bf 0.99}  & {\bf 0.96}  &  {\bf 0.97} & {\bf 0.95} & 0.94 & {\bf 0.97} \\
FP (balanced)                    & {\bf 0.99} & {\bf 0.97} & {\bf 0.97} & {\bf 0.97} & {\bf 0.97} & {\bf 0.97} \\
FP (rand.\ exploration $\epsilon=0.25$)                & {\bf {\bf 0.96}} & 0.88 & 0.94 & 0.26 & 0.64 & {\bf 0.99} \\
FP (rand.\ exploration $\epsilon=0.5$)                 & {\bf 0.98} & {\bf 0.98} & {\bf 0.99} & {\bf 0.98} & {\bf 0.95} & {\bf 0.99} \\\bottomrule
\end{tabular}
\caption[Test accuracy of various models per iteration in the dataset batch size case]{Test accuracy of various models per iteration in the dataset batch size case
(using batch size equal to the size of the full training set) for bAbI, Task $6$.
 Results $>0.95$ are in bold.
\label{table:dataset-batch-babi}
}
\end{table*}

\subsection{Human Feedback} \label{sec:mturkexp}
We employed Turkers to both ask questions and then give textual feedback
on the bot's answers, as described in Section \ref{sec:data-mturk}.
Our experimental protocol was as follows.
We first trained a MemN2N using supervised (i.e., imitation) learning on a training 
set of 1000 questions produced by Turkers and using the known correct answers 
provided by the original dataset (and no textual feedback).
Next, using the trained policy, we collected textual feedback for the responses
of the bot for an additional 10,000 questions.
Examples from the collected dataset are given in Figure \ref{fig:mturk_data}.
Given this dataset, we compare various models: RBI, FP, and FP+RBI.
As we know the correct answers to the additional questions, we can  assign a positive reward
to questions the bot got correct. We hence measure the impact of the sparseness of this
reward signal, where a fraction $r$ of additional examples have rewards.
The models are tested on a test set of $\sim$8,000 questions (produced by Turkers),
and hyperparameters are tuned on a similarly sized validation set.
Note this is a harder task than the WikiMovies task in the simulator due to 
the use natural language from Turkers, hence lower test performance is expected.

Results are given in Table \ref{table:mturk-res}.
They indicate that both RBI and FP are useful.
When rewards are sparse, FP still works via the textual feedback 
while RBI can only use the initial 1000 examples when $r=0$.
As FP does not use numericalrewards at all, it is invariant to the parameter $r$.
The combination of FP and RBI outperforms either alone.

\begin{table*}[!tbh]
\begin{center}
\begin{tabular}{lcccc}\toprule
Model                         & $r=0$   &  $r=0.1$  &  $r=0.5$  & $r=1$ \\\midrule
Reward Based Imitation (RBI)  & 0.333    &  0.340   &  0.365   & 0.375 \\
Forward Prediction (FP)       & 0.358    &  0.358   &  0.358   & 0.358 \\
RBI+FP                        & 0.431    &  0.438   &  0.443   & 0.441 \\\bottomrule
\end{tabular}
\end{center}
\caption[Incorporating feedback from humans via Mechanical Turk.]{
Textual feedback is provided for 10,000 model predictions (from a model trained with 1k labeled
 training examples), and additional sparse
binary rewards (fraction $r$ of examples have rewards).
Forward Prediction
and Reward-based Imitation are both useful, with their combination performing best.
\label{table:mturk-res}
}
\end{table*}

\section{Conclusion}

We studied dialogue learning of end-to-end models using textual feedback and numerical rewards.
Both fully online and iterative batch settings are viable approaches to policy learning, as long as
possible instabilities in the learning algorithms are taken into account. Secondly, we showed 
for the first time that the  FP method can work
in both an online setting and on real human feedback.
Overall, our results indicate that it is feasible to build a practical
 pipeline that starts with a model trained on an initial fixed dataset,
which then learns from interactions with humans  in a (semi-)online fashion to improve itself.

For the results, we can see that there is still a gap between the model only trained based on the simulator and the one 
bootstrapped from human feedback. This suggests that further improvement could be achieved when more human-in-the-loop efforts are incorporated.

\chapter{Conclusion and Future Work}
\label{conclusion}
In this dissertation, we have presented methods for developing more human-like and interactive 
 conversational agents. We discussed the opportunities presented by 
current deep-learning models in chat-bot development, along with challenges that arise, and how to tackle them. 
In this section, we will conclude this thesis and discuss the challenges that are still faced by 
current conversational systems, and suggest future work to tackle them.

In Chapter 3, we discussed 
how to address 
the issue that current neural chit-chat systems tend to 
produce  dull, generic, and uninteresting responses such as ``{\it i don't know}" or 
``{\it i don't know what you are talking about}. We discovered that this is due to the maximum-likelihood objective function
adopted in standard sequence generation systems. We proposed using  mutual information 
as the objective function in  place of the MLE objective function, where we not only want the generated response to be specific 
to the input dialogue history, but  want the dialogue history to be specific to the generated response as well.
Using  MMI as an objective function, we observed that the model is able to generate more specific, coherent, and 
interesting responses.

In Chapter 4, we discussed how to give the model a consistent persona to preserve its speaker consistency. 
We proposed using user embeddings to encode the information of users, such as user attributes or word usage tendency. 
These embeddings are incorporated into the neural encoder-decoder models, pushing the model to generate more consistent responses. 
We also proposed a speaker-addressee model, in which the utterance that the conversational agent  generates not only
depends on the identify of the speaker, but also the identify of its dialogue parter. 
We observe a clear performance improvement introduced by the proposed persona model and the speaker-addressee model in automatic evaluation
metrics BLEU and perplexity. 
Human evaluation performed by Turkers also demonstrates that the proposed persona model is able to generate more consistent responses than
the vanilla \sts models.

In Chapter 5, we proposed a 
multi-turn dialogue
system, in which we are not only  concerned with the quality of single-turn conversations,  
but the overall success of multi-turn conversations. The MLE objective function used in standard \sts models is trained to predict the next turn utterance, and is thus
 not suited for a system optimized 
for multi-turn dialogue success, especially because the positive or negative outcome of a dialogue utterance in a multi-turn dialogue usually does not show up after a few turns later. 
We proposed to tackle this issue using reinforcement learning, in which we created a scenario for the two bots to talk with each other. We manually designed three types of rewards that we 
think are important for a successful conversation, namely, ease to follow, informational flow, and meaningfulness. The RL system is trained to push the system 
to generate responses that are highly rewarded regarding these three aspects. Human evaluation illustrates that the proposed RL system tends to generate more sustainable conversations. 

In Chapter 6, we proposed a more generic reward function for training a RL-based dialogue system: a
machine-generated
 response is highly rewarded it it is
indistinguishable human-generated responses. 
We proposed  using adversarial learning, in which we jointly train two models, 
a generator that defines the probability of generating a dialogue sequence given input dialogue history, and 
a discriminator
that labels dialogues as human-generated or machine-generated. 
We cast the task as a reinforcement learning problem, in which the quality of machine-generated utterances is measured by its ability to fool the discriminator into believing that it is a human-generated one. The output from the discriminator is used as a reward to the generator, pushing it to generate 
utterances indistinguishable from human-generated dialogues. 
Human evaluation reveals that that the proposed adversarial reinforcement learning is able to generate better-quality responses. 

In Chapter 7, we focused on the task of developing interactive QA dialogue agents, where we give  an agent the ability to ask its dialogue partner questions. 
We define three scenarios where we think the bot can benefit from asking questions:
(1) question clarification, in which the bot asks its dialogue partner for text clarification; (2) knowledge operation, in which the bot asks the teacher for hints on relevant knowledge-base fact; and (3)
knowledge acquisition, in which the bot asks about new knowledge. 
We validated that the bot always benefits from asking questions in these three scenarios. 
We additionally explored how a bot can learn when it should ask questions by designing a setting where the action of question asking comes at a cost. 
This has real-world implications since it would be  user-unfriendly if a dialogue agent asks questions all the time. 
We designed a reinforcement learning system, within which the bot is able to automatically figure out when it should ask questions that will lead to highest future rewards.

In Chapter 8, we designed a RL system that gives the bot the ability to learn from online feedback: adapting its model when making mistakes and reinforcing the model when users' feedback is positive.  
The bot takes advantage of both explicit numerical rewards, and textual feedback which is more natural in human dialogue. 
We explore important issues such as how a bot can be most efficiently trained using a minimal amount of user feedback, how to avoid pitfalls such as instability during online learning with different types of feedback via data balancing and exploration, and how to make learning with real humans feasible via data batching.
We show that it is feasible to build a pipeline that starts from a model trained with fixed data and then learns from interactions with humans to improve itself. 

Together, these contributions have created a system that is able to generate more interactive, smart, user-consistent,  
long-term successful chit-chat system. 
~\\~\\
We will conclude this dissertation with a discussion of
 the challenges that current chit-chat systems still face, and
suggest some avenues for
future research.

\paragraph{Context}
In this thesis, we use a hierarchical LSTM models with attention to capture contexts, where a word-level of LSTM is used to obtain the representation for each context sentence, and
another level  LSTM combines sentence-level representations into a context vector to represent the entire dialogue history. 
But
it is not
straightforwardly
 clear
how much context information this context vector is able to capture, and how well 
this hierarchical attention model is able to 
  separate informational 
wheat from chaff. This incapability might stem from (1) the lack of capacity of current neural network models, where 
one single context information 
doesn't  have enough capacity to 
 encode all context information or (2) the model's incapability of figuring out 
which context sentence provides more evidence 
than others
on what the bot should say next. 
Tackling these issues is extremely important in real-world applications such as  customer service chatbot development. Think about  the domain of mailing package tracking, 
in which
the bot needs to memorize some important information such as a tracking number throughout the conversation. 
Combining  information extraction methods
or a slot filling strategy 
 that extracts 
important entities in the dialogue history with representation-based neural models has the potential to tackle this problem.
Intuitively, only a very small fraction of key words in the dialogue history  have significant guidance on what the bot should talk about. Using a key-word based information extraction
model  that first extracts these keywords and then incorporates them into the context neural model provides more flexibility in leveraging information from larger history context.

Furthermore, textual context only consists one specific type of evidence when we humans converse. 
Our conversations are also grounded on visual context, e.g., what we see and the environment (e.g., in a car) in which the conversation takes place.
Being able to leveraging visual information (and also other types of information) will significantly narrow down the 
topics to discuss, and will consequentially lead to responses of better quality.

\paragraph{Logics and Pragmatics}
Think about the following two contexts of an ongoing conversation: {\it A: are you going to the party?} {\it B: Sorry, I have an  exam tomorrow}. 
 From this context, we know that the speaker $B$ is not going to the party because he has to prepare for the upcoming exam, on which 
 the conversation that follows should be based. This requires a sequence of reasoning steps, namely,
 {\it having an exam tomorrow} $\rightarrow$ 
  {\it having to prepare for the exam} $\rightarrow$ {\it being occupied } $\rightarrow$ {\it not able to attend the party}. Straightforward as it seems to humans, 
  such a line of reasoning is 
  extremely hard for current machine learning systems, especially in an open-domain: manually labeling all the reasoning chains in an open domain is unrealistically work-demaning. 
  We thus need a logic deduction model, which automatically learns these implicit reasoning chains from a massive amount of training data and then incorporate them to conversational generation. 
     
 \paragraph{Background and Prior Knowledge}
 Human conversations always take place in a specific context or background. The granularity could be as small/specific as the location that a conversation takes place in (e.g., a cafeteria or a theater), or as large as during  war time vs peaceful time. 
 This background has
   a significant impact on how the conversation should go. 
 The context also includes user information, personal attributes, or even the speaker's general feeling for his dialogue partner. For example whether the addressee is responsible or honest. 
 Challenges for handing the background issue are two fold: (1) at the training data level, it is hard to collect comprehensive information about the background in which a dialogue takes place. As we discussed in the previous sections, most large-scale available datasets  come from social media like Twitter, online forums like reddit or movie scripts, which usually lack a detailed description of 
 the background, for example, it is impossible to collect thorough information about the persona of a speaker participating in the Twitter conversation. 
 One can think of the persona model described in Section \ref{persona} as building up speaker information/profile based on its previous generated conversations. 
 But only using hundreds or thousands of dialogue turns that a user previously published on Twitter is far from fully knowing them. 
  (2) the implication that a specific context has for a conversation happening in that context requires a huge amount of prior common-sense knowledge. When we humans converse, these common senses are rarely explicitly mentioned or described since a dialogue participant just takes them for granted. This means that even if we have concrete context information for a conversation, 
 why the conversation that takes place in this  context as it is is not in the least clear, since a significant amount of common sense information is omitted by the speakers.
 This poses significant challenges for imitation based machine learning systems (e.g., the \sts model), since without knowing why, purely imitating the way humans talk using the training set is not an optimal way to understand human communications. 
 
I hope that this dissertation can inspire researchers in the direction of dialogue understanding and generation, and encourage ongoing research to 
tackle the issues described just above.

\bibliographystyle{acl}
\bibliography{ref}
\end{document}